\documentclass[runningheads]{llncs}

\usepackage{algorithm}      % 
\usepackage{algorithmic}    % 
\usepackage{multicol}       % table formatting
\usepackage{multirow}       % table formatting
\usepackage{subcaption,lipsum,booktabs} % 20240709 - table formatting, subtable
\usepackage[square,numbers]{natbib}     % 20240709 - citation from multiple sources
\usepackage{tabularx}   % 20240711 - Adaptive width table

\makeatletter               % table formatting (setting horizontal line width)
\def\hlinewd#1{%
\noalign{\ifnum0=`}\fi\hrule \@height #1 \futurelet
\reserved@a\@xhline}
\makeatother
\usepackage{color}  
\usepackage[dvipsnames]{xcolor}         % diverse colors

\newcommand{\mywidth}{1.0}              % figure/table formatting

\usepackage{rotating}

\usepackage{eccvabbrv}

\usepackage{graphicx}
\usepackage{booktabs}

\usepackage[accsupp]{axessibility}  % Improves PDF readability for those with disabilities.

\usepackage{orcidlink}

\begin{document}

\title{Efficient Neural Video Representation with Temporally Coherent Modulation} 


\author{Seungjun Shin$^*$\orcidlink{0000-0003-0891-0861} \and
Suji Kim$^*$\orcidlink{0009-0002-9286-0552} \and
Dokwan Oh\orcidlink{0000-0001-9139-8699}}

\authorrunning{S. Shin, S. Kim, D. Oh}

\institute{Samsung Advanced Institute of Technology \\
\email{\{sj0216.shin, sujii.kim, dokwan.oh\}@samsung.com} }

\maketitle
\def\thefootnote{*}\footnotetext{Equally contributed.}
\def\thefootnote{\arabic{footnote}}

\begin{abstract}
 Implicit neural representations (INR) has found successful applications across diverse domains. To employ INR in real-life, it is important to speed up training. In the field of INR for video applications, the state-of-the-art approach \cite{kim2022scalable} employs grid-type parametric encoding and successfully achieves a faster encoding speed in comparison to its predecessors \cite{chen2021nerv}. However, the grid usage, which does not consider the video's dynamic nature, leads to redundant use of trainable parameters. As a result, it has significantly lower parameter efficiency and higher bitrate compared to NeRV-style methods \cite{chen2021nerv, li2022nerv, chen2023hnerv} that do not use a parametric encoding. To address the problem, we propose \textit{Neural Video representation with Temporally coherent Modulation} (NVTM), a novel framework that can capture dynamic characteristics of video. By decomposing the spatio-temporal 3D video data into a set of 2D grids with flow information, NVTM enables learning video representation rapidly and uses parameter efficiently. Our framework enables to process temporally corresponding pixels at once, resulting in the fastest encoding speed for a reasonable video quality, especially when compared to the NeRV-style method, with a speed increase of over 3 times. 
 Also, it remarks an average of 1.54dB/0.019 improvements in PSNR/LPIPS on UVG (Dynamic)  (even with 10\% fewer parameters) and an average of 1.84dB/0.013 improvements in PSNR/LPIPS on MCL-JCV (Dynamic), compared to previous grid-type works. By expanding this to compression tasks, we demonstrate comparable performance to video compression standards (H.264, HEVC) and recent INR approaches for video compression. Additionally, we perform extensive experiments demonstrating the superior performance of our algorithm across diverse tasks, encompassing super resolution, frame interpolation and video inpainting.
\end{abstract}

\section{Introduction}
\label{sec:intro}

Implicit neural representation (INR) is a technique that represents a signal as a continuous function of its corresponding coordinates. Because it is effective to handle complex signals, INR has gained considerable attention across various domains such as images \cite{sitzmann2020implicit, chen2021learning}, sounds \cite{szatkowski2022hypersound, su2022inras}, 3D objects and scenes \cite{fang2023one, takikawa2021neural, dong2022pina, mildenhall2020nerf, genova2020local}, and compression \cite{dupont2021coin, dupont2022coin}. Following this trend, video applications of INR are now being explored in many studies \cite{kim2022scalable, chen2022videoinr}. They have unique advantages such as the ability to play videos at arbitrary resolutions and frame rates, as well as the capability for video inpainting. Furthermore, leveraging INR in video compression leads to remarkable breakthroughs \cite{li2022nerv,chen2023hnerv,maiya2023nirvana, zhao2023dnerv, he2023towards, gomes2023video}.

\begin{figure}[!t]
\renewcommand{\mywidth}{0.5}
\centering
    \begin{subtable}[t]{0.41\textwidth}
        \centering
        \begin{tabular}{c}
        \includegraphics[width=\linewidth]{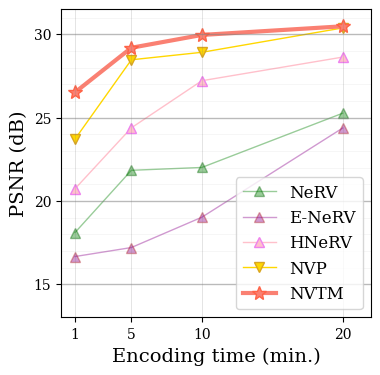}
        \end{tabular}
    \end{subtable}
    \begin{subtable}[t]{0.55\textwidth}
        \centering
        \begin{tabular}{cc}
        \includegraphics[width=\mywidth\linewidth]{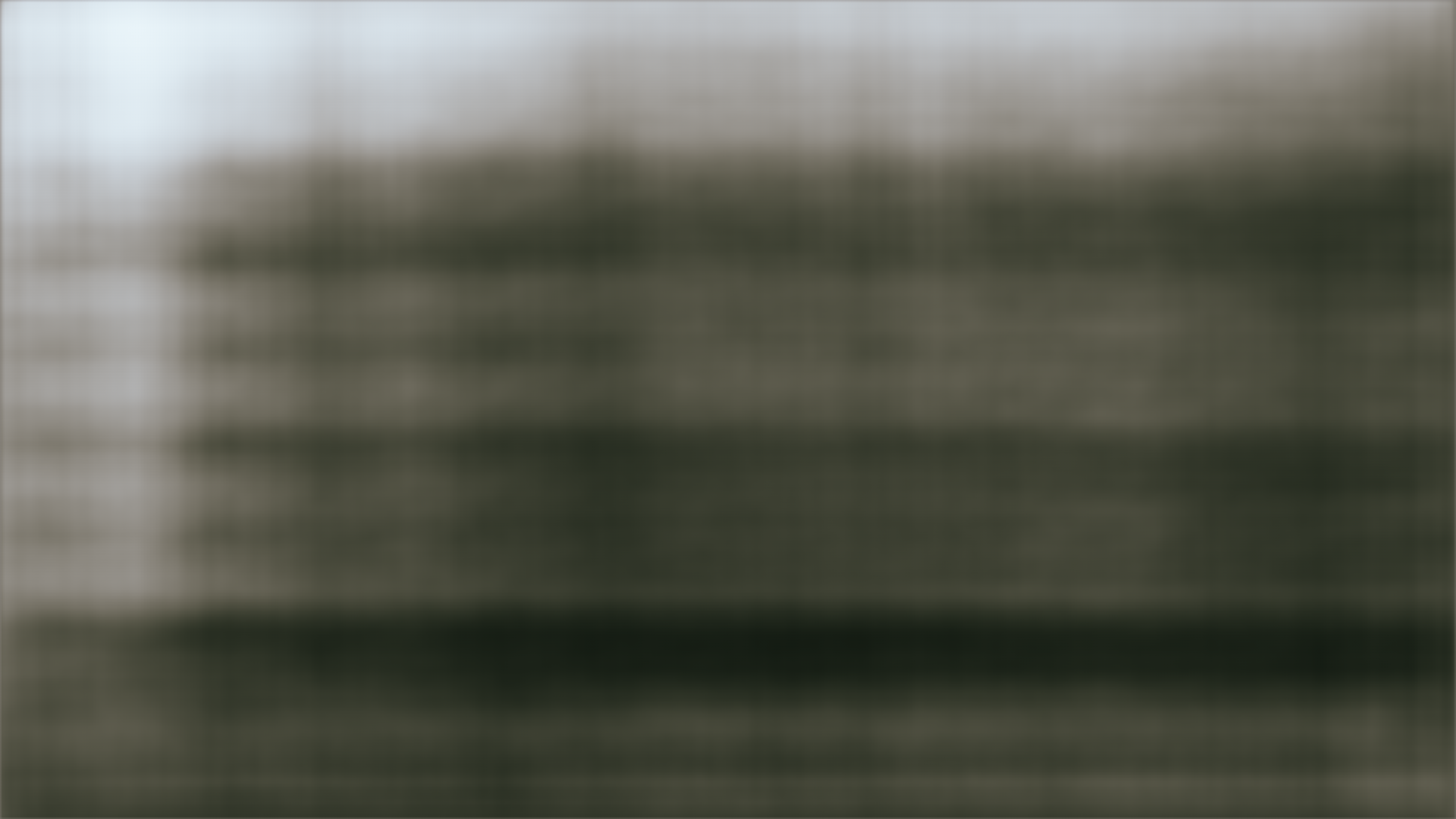} &
        \includegraphics[width=\mywidth\linewidth]{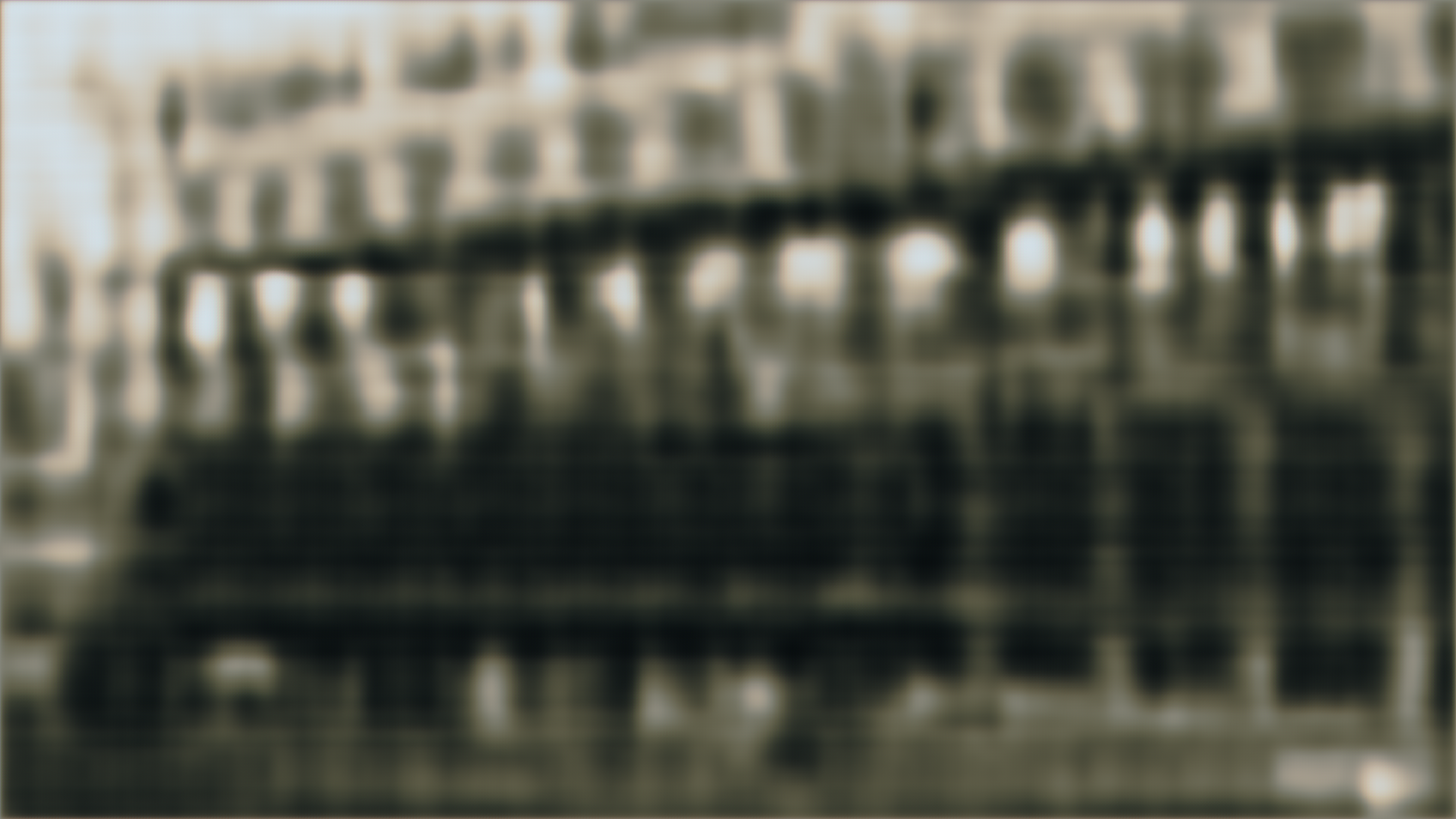} \\
        E-NeRV~\cite{li2022nerv} & HNeRV~\cite{chen2023hnerv} \\
        \includegraphics[width=\mywidth\linewidth]{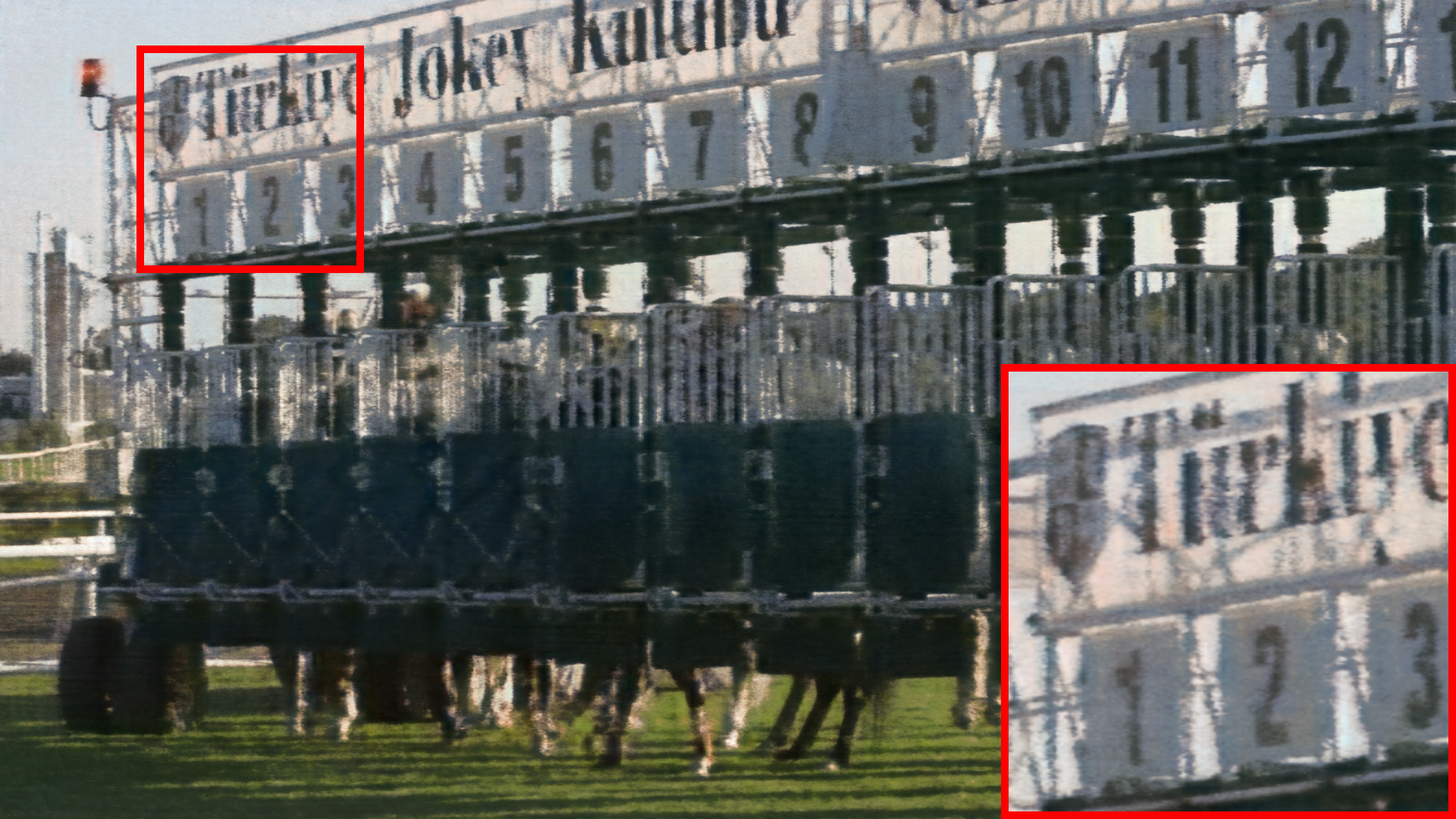} & 
        \includegraphics[width=\mywidth\linewidth]{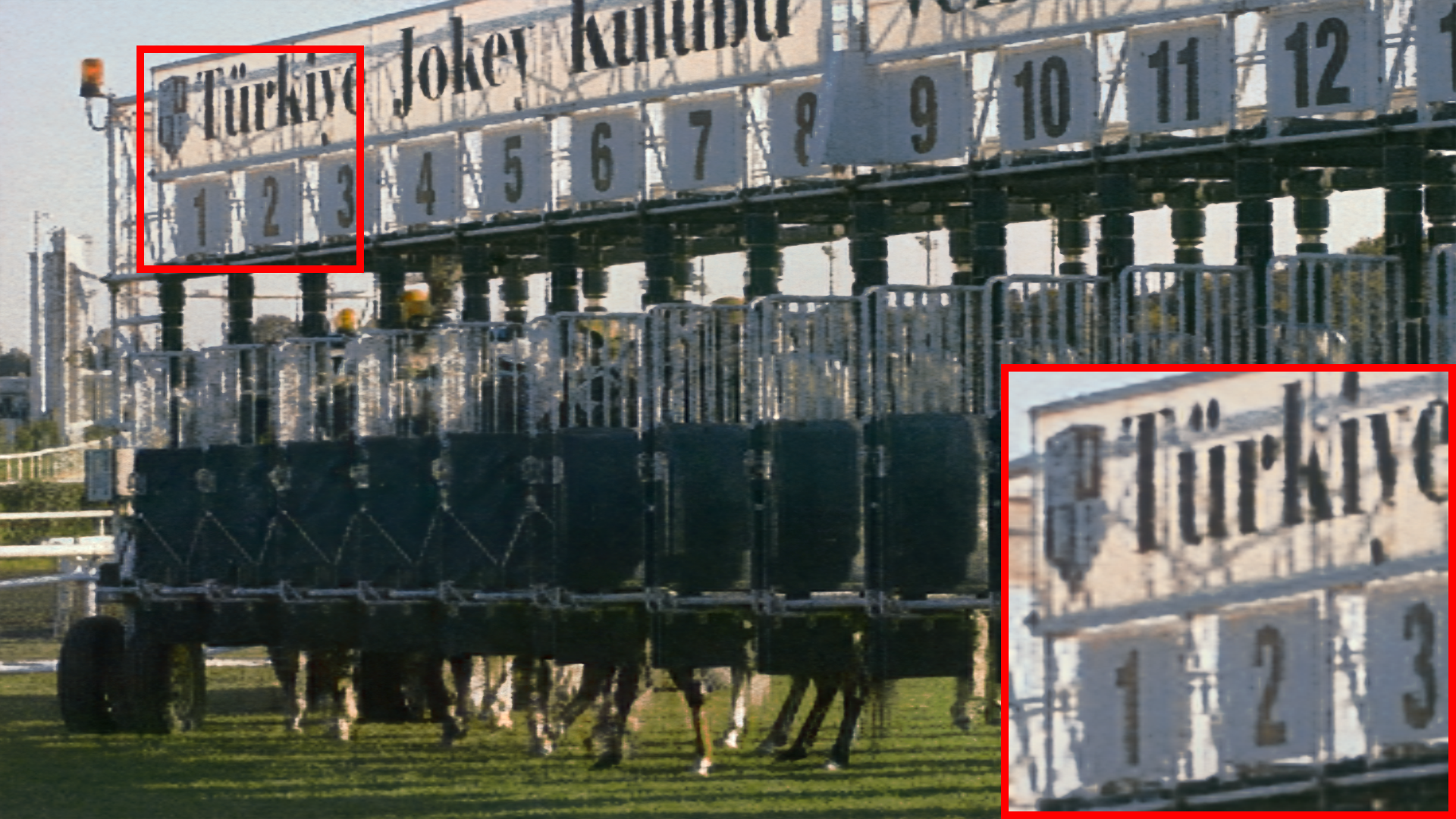} \\
        NVP \cite{kim2022scalable} & NVTM (Ours) \\
        \end{tabular}
    \end{subtable}
\vspace{-0.2cm}
\caption{\small \textbf{Fast encoding speed with high image quality.} (Left) The encoding speed in UVG, where all models are configured at \textbf{0.1bpp} and evaluated on the same resource conditions.
NVTM learns quickly and achieves 30dB \textbf{3$\times$faster than the NeRV-series}. 
(Right) Video reconstruction results on ReadySetGo sequence \textbf{after training for 1 minutes}.
While E-NeRV and HNeRV exhibit blurry outputs, NVP and NVTM, based on parametric encoding, quickly capture complex representations.
Further, NVTM excels at representing fine details such as text and numbers.
}
\vspace{-0.5cm}
\label{fig:encoding_time}
\end{figure}

NeRV \cite{chen2021nerv}, a framework that iteratively combines convolution and pixel-shuffle operators, was proposed for the application of INR in video reconstruction and compression. Numerous follow-up studies are conducted \cite{li2022nerv,chen2023hnerv,maiya2023nirvana, zhao2023dnerv, he2023towards, gomes2023video} and they emphasize the applicability in video compression by diminishing input dimensions or replacing input coordinates with image features.
However, they lost one of INR's major advantages, which is the capacity to produce outputs at various resolutions using a single learned model. In addition, the slow encoding speed still remains as the main challenge in those architectures. Since all parameters of network must be updated for every pixel, serious computational inefficiency occurs and the encoding time increases. To overcome this challenge, parametric encoding (e.g., grid) is being widely adopted \cite{chabra2020deep,fridovich2022plenoxels, liu2020neural, chibane2020implicit, jiang2020local}. They achieve a faster training speed via their strong locality (computing-efficient), however, they have the drawback that the model size needs to increase according to the input dimension (parameter-inefficient). Also, the attempts to directly embedding videos into a 3D grid \cite {girish2023shacira, muller2022instant} or decomposing into three 2D grids \cite{kim2022scalable} did not sufficiently consider the dynamic nature of videos. This results in the duplication of parameters, and large parameter size being required for reasonable performance. 
% This results in a large parameter size being required for reasonable performance. 

In the field of video processing, it is important to deal with temporal redundancy between adjacent frames \cite{chu2020learning, yang2020spatial}. Video codecs \cite {wiegand2003overview, sullivan2012overview} also efficiently encode videos by dealing such temporal redundancy with motion compensation similar to the philosophy, leaving only residual information in each frame. Several studies \cite{lee2023ffnerv, rho2022neural} in the INR fields have adopted a warping-residual structure to eliminate temporal redundancy. However, no studies have been conducted to consider removing temporal redundancy while using parametric encoding. Therefore, we propose a computing-efficient (fast encoding speed, shown in Figure~\ref{fig:encoding_time}) and parameter-efficient (high reconstruction quality, shown in Table~\ref{tb:mainresult_final}). INR framework that takes into account the dynamic characteristics of videos. The key idea is utilizing a series of 2D grids to represent videos by employing the same modulation to the corresponding pixels. Overall, we make the following contributions:

\begin{itemize}
\setlength\itemsep{0.05em}
    \item We propose a novel framework \textit{Neural Video representation with Temporally coherent Modulation} (NVTM), which applies consistent modulation equally to corresponding along the time axis.
    \item Our framework achieves a fast training speed and high parameter efficiency on video representation.
    \item We validate the performance on extensive experiments with various datasets and various tasks including video reconstruction, video compression, video super resolution, video frame interpolation, and video inpainting compared to state-of-the-art methods.
\end{itemize}

\section{Related Works}
\subsection{Implicit Neural Representation (INR)}
INR, also known as neural fields or coordinate-based neural representation (CNR), has emerged as a new paradigm for representing complex and continuous signals. It interprets data as a continuous signal and proposes a methodology where data is encoded into a neural network using coordinate inputs. 
\subsection{INR for videos}
\label{related:inr_for_videos}
Video data is composed of consecutive frames, and many studies have attempted to find a better framework to apply INR to video.
Since pixel-wise INR, which output the $(r,g,b)$ for 3D coordinate input $(x,y,t) \in \mathbb{R}^{3}$, has slow encoding speed and low parameter efficiency, {NeRV \cite{chen2021nerv} proposed to frame-wise INR with 1D coordinate input $t \in \mathbb{R}^{1}$.
Although this frame-wise INR on does not consider spatial input $(x,y) \in \mathbb{R}^{2}$, it could efficiently represent videos with comparable performance in video compression. Subsequent studies also have provided notable improvements. \cite{li2022nerv} improved performance by eliminating redundancy in model parameters, and \cite{gomes2023video} leveraged coding efficiency by imposing constraints on weight entropy. Furthermore, \cite{maiya2023nirvana, bai2023ps} have extended the frame-wise INR to the patch-wise INR, enabling an improved representation of videos. However, they still have a limitation in that they cannot be expanded spatially, then can only be decoded at a fixed resolution size.

On the other hand, some approaches try to encode the difference between frames, instead of directly encoding the frames themselves. \cite{he2023towards} generates the entire video using warping and upsampling from given compressed keyframes, and \cite{lee2023ffnerv} reconstructs the final frame by generating a flow map and independent frames between adjacent frames and aggregating them. These methods effectively reduce the temporal redundancy which is inherent in video data, and demostrate outstanding performance in video compression. In addition, \cite{chen2023hnerv, zhao2023dnerv} suggested that 1D coordinate input $t \in \mathbb{R}^{1}$ in frame-wise INR was insufficient for accurately modeling the video's context feature. Based on this finding, \cite{chen2023hnerv} proposed a structure that integrates context features with a video-specific decoder, whereas \cite{zhao2023dnerv} established a structure that combines context features with difference features. These studies exhibited superior performance compared to traditional frame-wise INRs. 

\begin{table}[!t]
    \caption{\small 
    The performance on NVP~\cite{kim2022scalable} drops when the parameter size of temporal axis ($T$) is decreased while the overall size of parameters ($X{\times}Y{\times}T$) is maintained. These results are based on 600-frame HD videos of UVG, and demonstrate a degradation in performance as the $T$ becomes smaller than the video length.
    }
    \label{tb:nvp_degradation}
    \centering
    \begin{small}
    \setlength{\tabcolsep}{6pt}
    \begin{tabular}{c|ccccc}
        \toprule
        $T$ & 600 & 300 & 200 & 100 & 60 \\
        \midrule
        PSNR & 36.34 & 35.59 {\scriptsize (-0.75)} & 33.99 {\scriptsize (-2.35)} & 31.65 {\scriptsize (-4.69)} & 29.47 {\scriptsize (-6.87)} \\
        \bottomrule
    \end{tabular}
    \end{small}
\vspace{-0.5cm}
\end{table}

Despite implementing several structural improvements, subsequent studies on NeRV still exhibit a very slow learning speed. As shown in Figure~\ref{fig:encoding_time}, unlike other models that successfully capture numbers and text in just one minute of training, HNeRV \cite{chen2023hnerv} and E-NeRV \cite{li2022nerv} fail to do the same and exhibits a blurry artifact. 

To address these problems, NVP \cite{kim2022scalable} indicates a new direction of pixel-wise INR for video, while achieving a fast encoding speed. It effectively learns the video representation by using parametric encoding (e.g., grid) that are used to improve the learning speed in INR's field \cite{muller2022instant, deng2023compressing}. Specifically, by decomposing the 3D coordinates into three 2D coordinates ${(x,y), (y,t), (t,x)}$ and employing a 3D sparse grid, they successfully trained an pixel-wise implicit video representation framework. However, this approach has a definite limitation as it simply treats videos as 3D data, without considering their dynamic nature at all. 

As shown in Table~\ref{tb:nvp_degradation}, performance degradation occurs if the grid parameters of the sparse 3D grid are not sufficiently secured along the time axis when the parameter size of temporal axis is lower than video length.
Particularly, despite the overall parameters remaining the same, as the grid parameters decrease along the time axis, the performance degradation becomes more severe. This implies that it does not properly remove temporal redundancy. In this paper, we propose a fast and parameter-efficient video representation using a grid-type parameter encoding that considers the dynamics of the video. while having 3D coordinate input $(x,y,t) \in \mathbb{R}^{3}$.

\subsection{Modulation for INR}
Although INR can represent each specific data instance successfully, it lacks generalization and requires re-training from scratch whenever different data instances are applied. Therefore, 
unlike conventional paradigm which inputs coordinates and outputs data values, several studies \cite{mehta2021modulated, dupont2022data, spatialfuncta23} have researched to further modulate network operations. \cite{mehta2021modulated} introduced an auxiliary modulator in parallel to the base network, controlling the frequency and phase of the base network to increase its representational power. \cite{dupont2022data, spatialfuncta23} proposed to learn the instance-specific shift modulation latent, allowing the base network to represent the entire dataset while each shift modulation latent represent each instance. However, alternative approaches \cite{szatkowski2022hypersound, figueiredo2023frame} utilize a hyper-network, which determines the weights of the base network according to each data, completely altering the operation of the base network. While primary studies focus on an instance-wise modulation, in this paper, we introduce the concept of pixel-wise modulation to represent video data more efficiently.

\section{Methodology}
\label{sec:Methodology}
\begin{figure}[!t]
\centering
\includegraphics[width=1.00\linewidth]{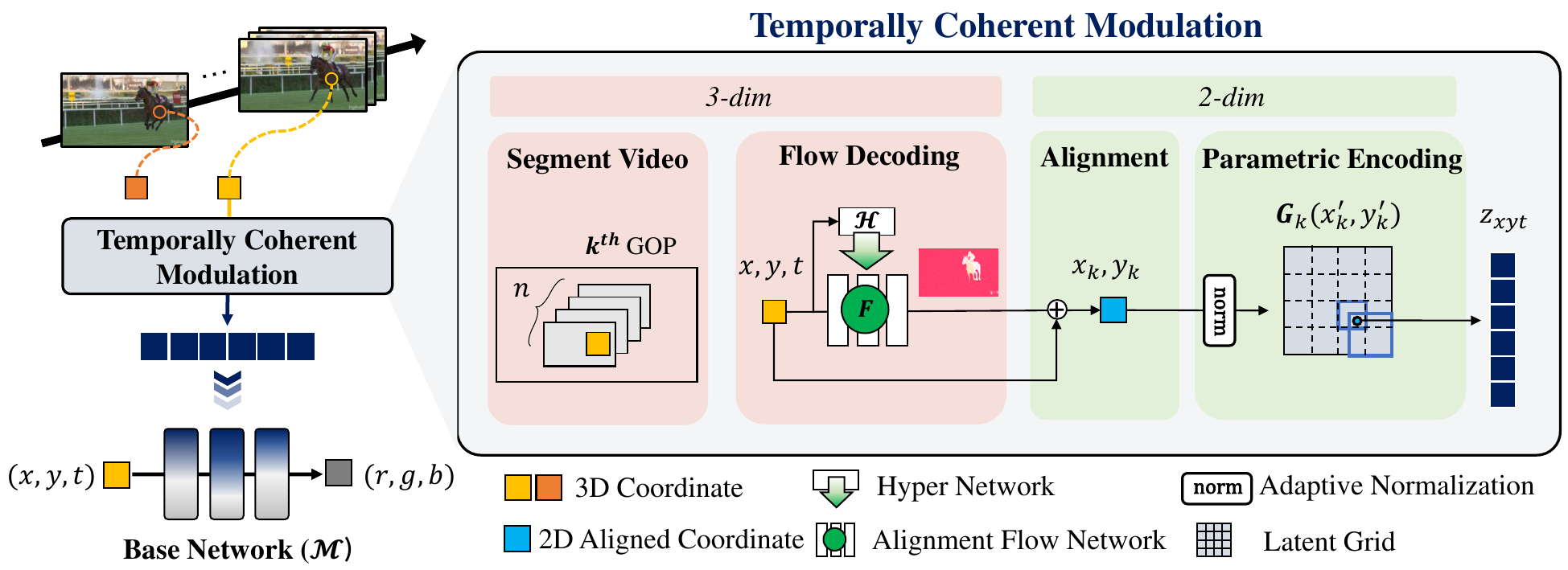}
\caption{
\small
\textbf{Overview of NVTM.}
NVTM generates the same modulation latent for temporally correlated pixels between consecutive frames, and the latent is used to modulate the base network. To obtain this latent, 1) input video is split into GOP units, 2) network $F$ generates an alignment flow to transform 3D coordinate $(x,y,t)$ to specific time $t_k$ in $k$-th GOP unit, 3) 2D aligned coordinated $(x_{k}, y_{k})$ is obtained by adding $(x,y)$ and the alignment flow. 4) The temporally coherent latent $(z_{xyt})$ is extracted from the latent grid $G_{k}$ using normalized $(x'_{k}, y'_{k})$.
Following the process, the temporally correlated 3D coordinates (yellow square and orange square) are mapped to the same 2D coordinate, thereby ensuring they share the same modulation latent representation. This shared modulation helps in the fast and efficient learning of video representation.
}
\label{fig:main_archi}
\vspace{-0.2cm}
\end{figure}

\subsection{Overall Framework}
The overview is described on Figure~\ref{fig:main_archi}. 
%In INR for videos, the 
The base network $\mathcal{M}$ takes the 3D coordinate $(x, y, t)$ as the input and produces $(r, g, b)$ as the output, and this can be expressed as $(r,g,b) = \mathcal{M}(x,y,t)$. The base network $\mathcal{M}$ also takes the latent $z_{xyt}$, obtained from each pixel, as a modulation value which affects the network's behavior. 
The process of computing the latent is detailed in following subsection.
\begin{equation}
    \centering
    \begin{aligned}
    \label{eq:modulation}
     (r,g,b) = \mathcal{M}_{\theta(z_{xyt})}(x,y,t).
    \end{aligned}
\end{equation} 
Utilizing the latent $z_{xyt}$ to modulate the base network enables temporally coherent modulation $\mathcal{M}_{\theta(z_{xyt})}$. We adopt the modulated-SIREN \cite{mehta2021modulated} as the modulation scheme for the base network. 

\subsection{Temporally Coherent Modulation}
NVTM is composed of an alignment flow network and multiple 2D latent grids. As mentioned previously, our key idea is to group similar pixels together and apply the same modulation, allowing the model to learn the pixel values quickly and sufficiently, even with fewer model parameters. 

\paragraph{\textbf{Segment Video}}
As mentioned before, it is needed to group similar pixels within the video. However, grouping all the pixels in the video is challenging and not effective. Therefore we aim to segment the input video into group-of-pictures (GOPs) of size $n$ and match corresponding pixels along time within each GOP. The number of GOPs, $m$, is calculated by dividing the total number of video frames with $n$. $k$-th GOP is composed of $\{t|(k-1)n \leq t < kn\}$-th frames, and the grouping alignment will executed within each GOP unit. 

\paragraph{\textbf{Flow Decoding}}
To match corresponding pixels, we align 3D coordinates into 2D coordinates at a specific time $t_k$, which is defined each GOP unit. We refer to this time as the keyframe time, and used the first frame of the each GOP unit as the keyframe in this work. For this process, the network $F$ generates a flow from input time $t$ to keyframe time $t_k$ for each 3D coordinate input $(x,y,t)$.

\begin{equation}
    \centering
    \begin{aligned}
    \label{eq:flow_naive}
    \text{Flow}_{t\xrightarrow{}t_k}(x,y) = F(x,y,t)
    \end{aligned}
\end{equation}
Since this alignment flow $F(x,y,t)$ is likely to be similar the optical flow from time $t$ towards keyframe time $t_k$, we utilize the optical flow as a guidance to train $F$, and we add an auxiliary loss at the beginning steps of training. In the decoding phase, the optical flow is not necessary as the output of $F$ is used directly. 

Meanwhile, the optical flow appears to move spatially over time, reflecting the movements of objects in video sequences \cite{figueiredo2023frame}.
Based on this flow observation, we make network $F$ be influenced from $t$ for easily learning flows with fewer parameters. For this, we adopt a hyper-SIREN \cite{figueiredo2023frame} as $F_{\mathcal{H}(t)}(x,y)$, which uses a hyper-network $\mathcal{H}(t)$ to generate SIREN $F$'s weights over time $t$. 
Further, to offset the difference in flow scale caused by the interval to $t_k$ at each $t$, we incorporated the log scale factor into the output scaling of the model.

\begin{equation}
    %\smallT
    \centering
    \begin{aligned}
    \label{eq:flow_hyper}
    \text{Flow}_{t\xrightarrow{}t_k}(x,y) = \log{(t-t_{k}})F_{\mathcal{H}(t)}(x,y)
    \end{aligned}
\end{equation}

\paragraph{\textbf{Alignment}}
Using the alignment flow, each 3D coordinate $(x,y,t)$ is warped into the 2D aligned coordinate $(x_{k},y_{k})$ of keyframe time $t_k$.
% the generated flow
\begin{equation}
    %\smallT
    \centering
    \begin{aligned}
    \label{eq:dimension_reduction_naive}
    (x_{k},y_{k}) = (x,y) + \log{(t-t_{k}})F_{\mathcal{H}(t)}(x,y)
    \end{aligned}
\end{equation}
\paragraph{\textbf{Parametric Encoding}}
Parametric encoding takes the form of extracting values from a parameter group corresponding to each input.
We utilize a parameter group structured as a 2D-grid type for each GOP and designate as the latent grid $G_k$ for the $k$-th GOP.
The 2D aligned coordinate $(x_{k},y_{k})$ serves as the input to $G_k$.
The input of the latent grid must satisfy $\in [0,1]$.
However, given that both alignment flow and the initial 3D coordinate $(x,y,t)$ range in  $[0,1]$, their sum, which results in $(x_k,y_k)$, may not satisfy this condition.
To adjust them, some naive approaches such as clipping or simple re-normalization $(x_k-\text{min}(x_k))/(\text{max}(x_k)-\text{min}(x_k))$ can be considered, but they have unacceptable side effects. Clipping occurs information loss and re-normalization decreases grid parameter efficiency. 

Therefore, we propose an adaptive normalization method, which can optimize the spatial utilization of the grid while containing the maximum amount of information. We search the largest area with a higher pixel density than the predefined threshold $r_{th}$ and define the area as \{$x^{min}_{k}$, $y^{min}_{k}$, $x^{max}_{k}$, $y^{max}_{k}$\}. Finally, $(x_k, y_k)$ are normalized into $(x'_k,y'_k)$ using the calculated min and max values.
\begin{equation}
    \centering
    \begin{aligned}
    x'_k = \text{Clip}\{(x_k-x^{min}_{k})/(x^{max}_{k}-x^{min}_{k}), (0,1)\} \\
    y'_k = \text{Clip}\{(y_k-y^{min}_{k})/(y^{max}_{k}-y^{min}_{k}), (0,1)\} \\
    \end{aligned}
\end{equation}
This adaptive normalization ensures that areas with a high pixel occupancy are properly normalized, whereas sparse regions are effectively handled by clipping the coordinates.

Modulation latent is obtained from the normalized coordinate, as $z_{xyt} = G_{k}(x'_{k},y'_{k})$.
Additionally, we extend modulation latent to utilizing two or more latent grids of neighboring GOPs. 
We define a neighbor index set $P$, and the final latent is obtained by concatenating all latents computed from the latent grids $\{G_{k},G_{k+1},..,G_{k+p}\}$ of the neighboring GOPs belonging to $P=\{0,1,..,p\}$.
\begin{equation}
    \centering
    \begin{aligned}
    \label{eq:2d_grid_multi}
     z_{xyt} = \text{concat}\{G_{k+p}(x'_{k+p},y'_{k+p}) |p \in P \}        
    \end{aligned}
\end{equation}

\paragraph{\textbf{Loss}} 
Total loss is a combination of the reconstruction loss and the auxiliary loss with the weight factor $w_{aux}$. The reconstruction loss $L_{recon}$ is the Mean Squared Error (MSE) between the original and reconstructed pixels, while auxiliary loss $L_{aux}$ is the MSE between the alignment flow and the optical flow. Then the total loss is calculated as $L_{total} = L_{recon} + w_{aux} \cdot L_{aux}$.

\section{Experimental Results}
\label{sec:exp}
\subsection{Implementation Details}
\label{sec:exp_implementation_details}
\paragraph{\textbf{Dataset.}}
We conduct experiments on UVG \cite{mercat2020uvg} and MCL-JCV \cite{wang2016mcl} datasets, which are widely used in various video tasks such as compression and quality assessment. Since our proposed approach is designed for videos which contain temporal dynamic information, we target on dynamic sequences among them.
Hence, we select 4 sequences from UVG and 5 sequences from MCL-JCV, which have large motion and sufficient spatial/temporal information. We convert those from raw YUV videos into RGB format and use the complete set of 600 frames for each sequence in UVG HD and initial 100 frames for each sequence in MCL-JCV HD. Details of statistics and data procedures are described in supplementary.

\paragraph{\textbf{Model configuration and training details.}}
These are our default experiment setting and more exploration are experimented on Section~\ref{sec:ablation}.
We configure NVTM as a default setting with a GOP size $n$ as 10. And we configure the index set P as \{0, 1\}.
To capture some static characteristics of video (e.g., still images), we additionally add a single 2D grid as static feature, similar to NVP \cite{kim2022scalable}. 
We utilize RAFT \cite{teed2020raft} to generate the optical flows, and set $w_{aux}$ as 0.5 for auxiliary loss of the alignment flow network. We use threshold value $r_{th}$ as 0.5 for adaptive normalization to ensure that at least half of the area is considered effective. All experiments are conducted on a single NVIDIA A100 GPU. More details are described in supplementary materials.

\begin{table}[!t]
\caption{\small  Encoding speed on video reconstruction. All models are configured as 0.1bpp and we compare their reconstruction performance (PSNR) based on the encoding time (i.e., the training time). 
\textbf{Bold} values represent the best value for each encoding time, and evaluated epoch or step (e/s) of each model is denoted.
} 
\label{tb:encodingtime_psnr}
    \setlength{\tabcolsep}{6pt}
    \centering
    \begin{small}
    \begin{tabular}{c|ccccc}
        \multicolumn{6}{c}{UVG (Dynamic)} \\
        \toprule
        \multirow{2}{*}{Models} & \multicolumn{5}{c}{Encoding time} \\
        ~ & $\sim$~1min. & $\sim$~5min. & $\sim$~10min. & $\sim$~20min. & $\sim$~60min.\\ 
        \midrule
        NeRV \cite{chen2021nerv} & 18.10$_\text{/1e}$ & 21.84$_\text{/9e}$& 22.01$_\text{/18e}$& 25.27 & 30.88  \\ %\cline{2-7}
        E-NeRV \cite{li2022nerv}& 16.66$_\text{/1e}$ & 17.20$_\text{/5e}$ & 19.02$_\text{/10e}$ & 24.38 & 30.00   \\ %\cline{2-7}
        HNeRV \cite{chen2023hnerv}& 20.73$_\text{/2e}$ & 24.39$_\text{/13e}$ & 27.22$_\text{/26e}$ & 28.64 & \textbf{32.51} \\ 
        \midrule
        NVP \cite{kim2022scalable}& 23.37$_\text{/250s}$ & 28.48$_\text{/1250s}$ & 28.93$_\text{/2500s}$ & 30.39 & 31.40   \\ %\cline{2-7}
        NVTM (Ours) & \textbf{26.5}2$_\text{/111s}$ & \textbf{29.20}$_\text{/556s}$ & \textbf{29.97}$_\text{/1111s}$ & \textbf{30.49} & 31.85 \\ 
        \bottomrule
    \end{tabular}
    \end{small}
    \begin{small}
    \vskip 0.5em
    \begin{tabular}{c|ccccc}
        \multicolumn{6}{c}{ MCL-JCV (Dynamic)}\\
        \toprule
        \multirow{2}{*}{Models} & \multicolumn{5}{c}{Encoding time} \\
        ~ & $\sim$~1min. & $\sim$~5min. & $\sim$~10min. & $\sim$~20min. & $\sim$~60min.\\ 
        \midrule
        NeRV \cite{chen2021nerv}& 19.09$_\text{/10e}$ & 21.93$_\text{/50e}$ & 23.10$_\text{/100e}$ & 24.67 & 28.33   \\ 
        E-NeRV \cite{li2022nerv}& 16.26$_\text{/7e}$ & 16.67$_\text{/36e}$ & 17.80$_\text{/72e}$ & 23.76 & 28.11  \\ 
        HNeRV \cite{chen2023hnerv}& 20.66$_\text{/12e}$ & 24.36$_\text{/63e}$ & 26.30$_\text{/126e}$ & 29.73 & 32.27  \\ 
        \midrule
        NVP \cite{kim2022scalable}& 25.34$_\text{/294s}$ & 29.26$_\text{/1471s}$ & 29.41$_\text{/2941s}$ & 31.13& 32.53  \\ 
        NVTM (Ours) & \textbf{27.71}$_\text{/250s}$ & \textbf{30.85}$_\text{/1250s}$ & \textbf{31.65}$_\text{/2500s}$ & \textbf{32.09} & \textbf{33.57}  \\ 
        \bottomrule
    \end{tabular}
    \end{small}
\end{table}

\begin{table}[!t]
\caption{\small  Video reconstruction performance with grid-type models. 
Each value represents the average on UVG and MCL-JCV respectively. 
\textbf{Bold} is the best value and $^\star$ indicates that ours uses 10\% fewer parameters compared to other methods.
We display the video reconstruction visualizations on videoSRC05 sequence as the pairs of decoded image and crop-zoomed images.
The below images are visualization of FLIP \cite{andersson2020flip} calculated from the original frame, the bright regions represent errors, while the darker colors indicate better performance.
} 
\label{tb:mainresult_final}
\centering
\setlength{\tabcolsep}{5pt}
\begin{small}
    \begin{tabular}{c|ccc|ccc}
    \toprule
    \multirow{2}{*}{Model} & \multicolumn{3}{c|}{UVG (Dynamic)} & \multicolumn{3}{c}{MCL-JCV (Dynamic)}\\ 
    ~ & {\small Params.} & {\small PSNR↑} & {\small LPIPS↓} & {\small Params.} & {\small PSNR↑} & {\small LPIPS↓} \\
    \midrule
    Instant-NGP \cite{muller2022instant} & 145M & 37.08 & 0.126 & 29M & 39.32 & 0.093\\ %\hline
    3D ModSIREN \cite{mehta2021modulated} & 134M & 37.24 & 0.095 & 29M & 36.96 & 0.134 \\ %\hline
    NVP \cite{kim2022scalable} & 136M & 39.00 & 0.090 & 29M & 39.55 & 0.093 \\ %\hline
    NVTM (Ours) & 122M$^{\star}$ & \textbf{40.54} & \textbf{0.071} & 28M & \textbf{41.39} & \textbf{0.080} \\ 
    \bottomrule
    \end{tabular}
    \setlength{\tabcolsep}{0.7pt}
    \begin{tabular}{cccc}
    \includegraphics[width=0.245\textwidth]{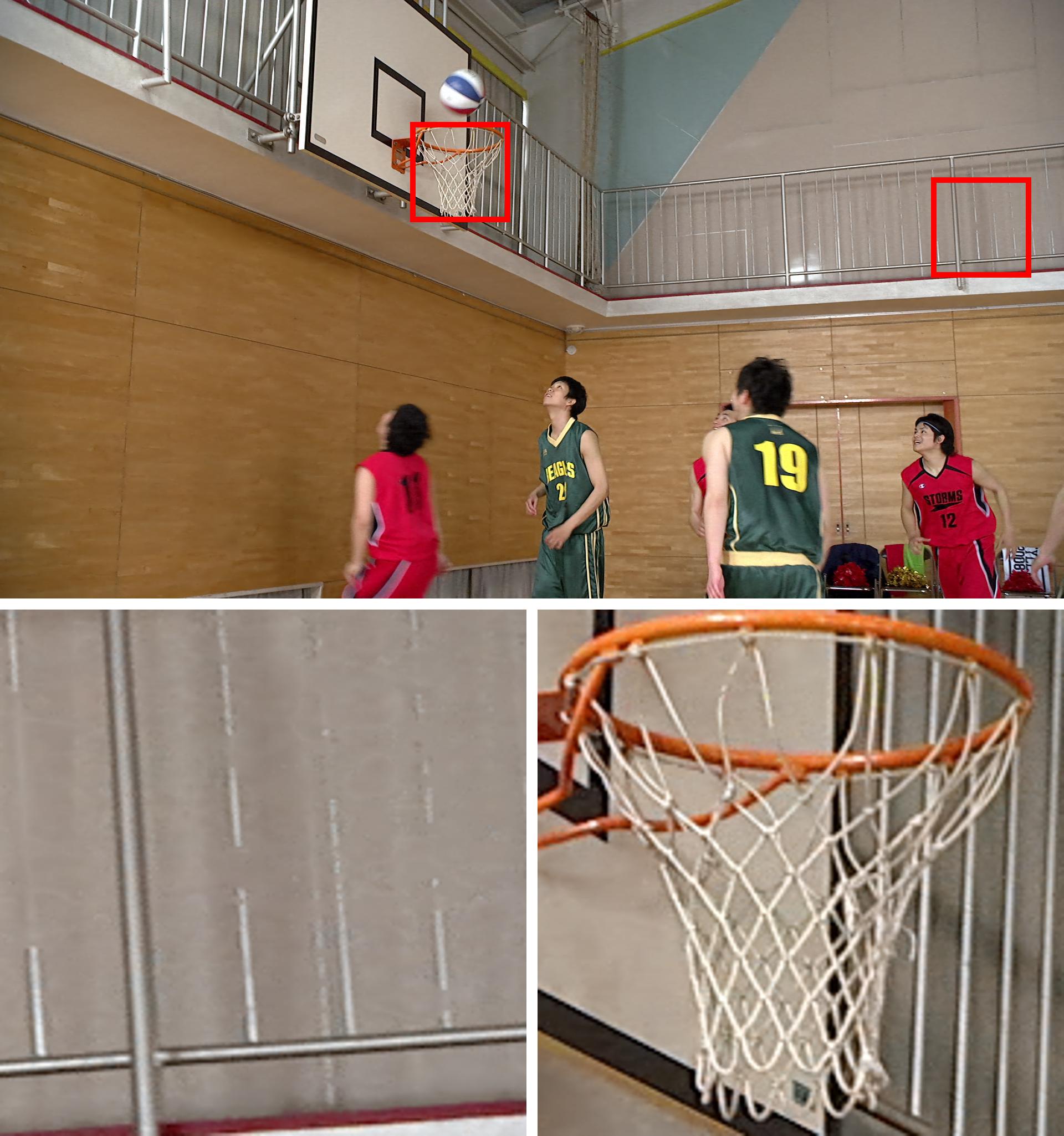} 
    & \includegraphics[width=0.245\textwidth]{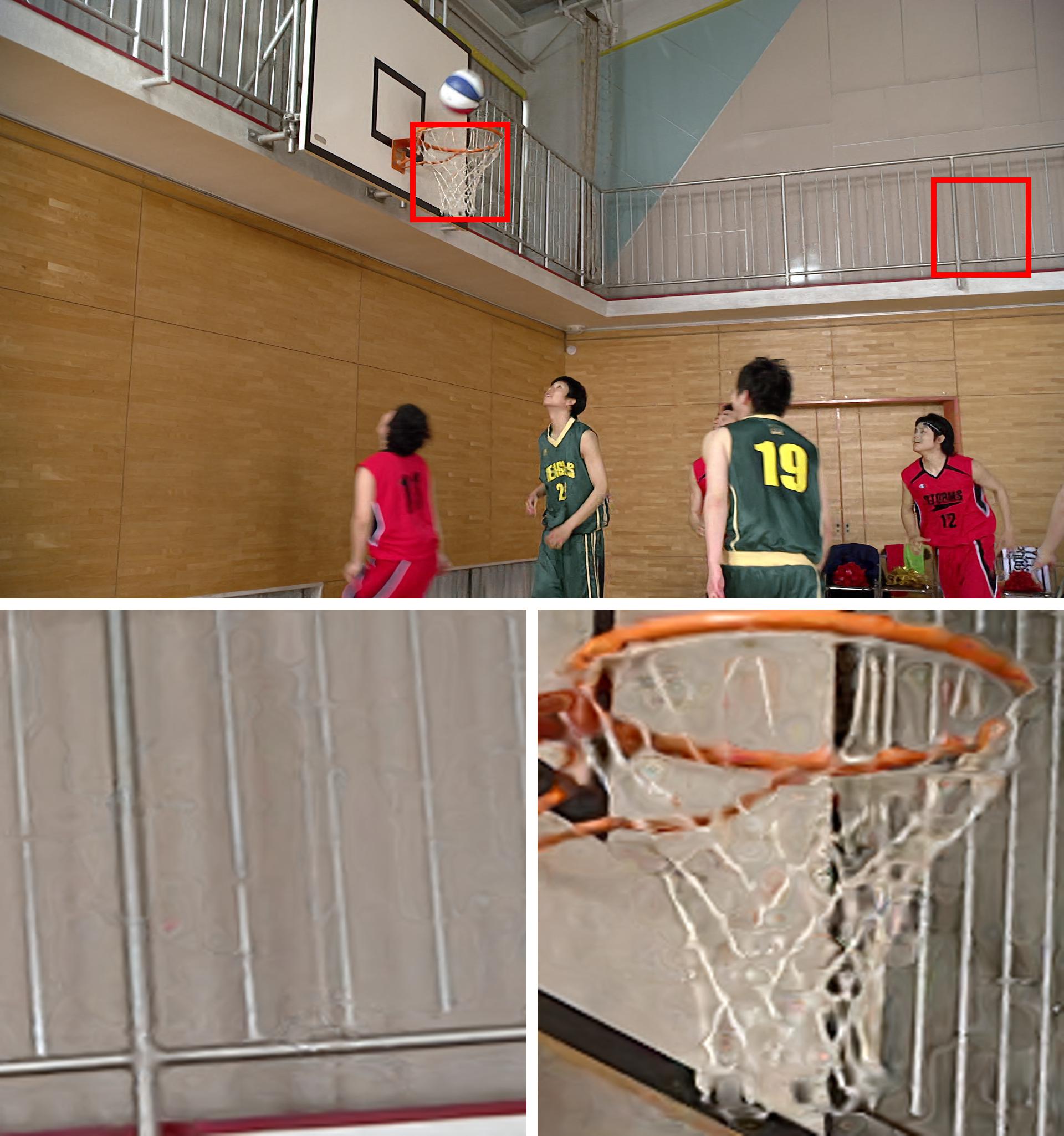} 
    & \includegraphics[width=0.245\textwidth]{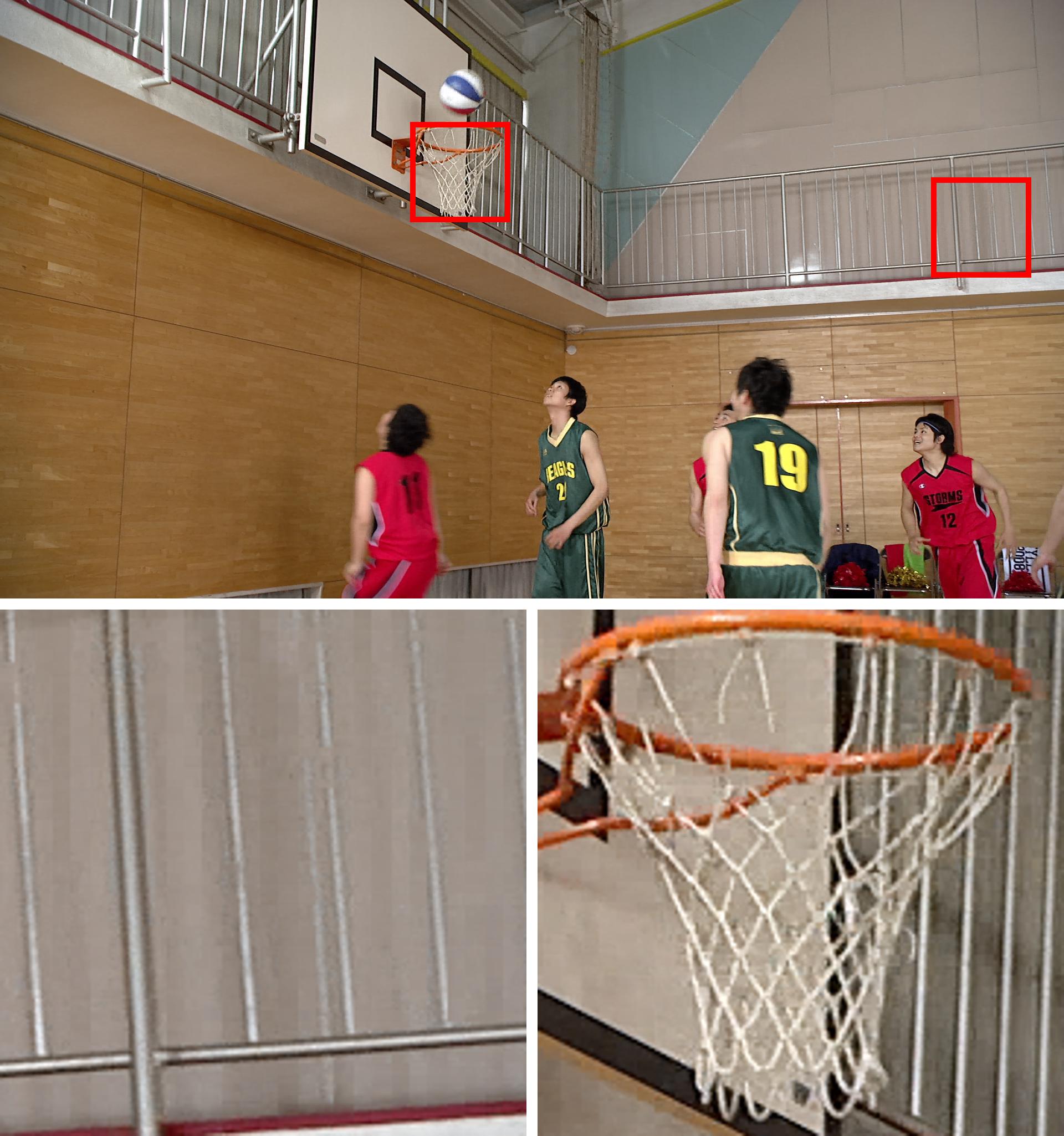} 
    & \includegraphics[width=0.245\textwidth]{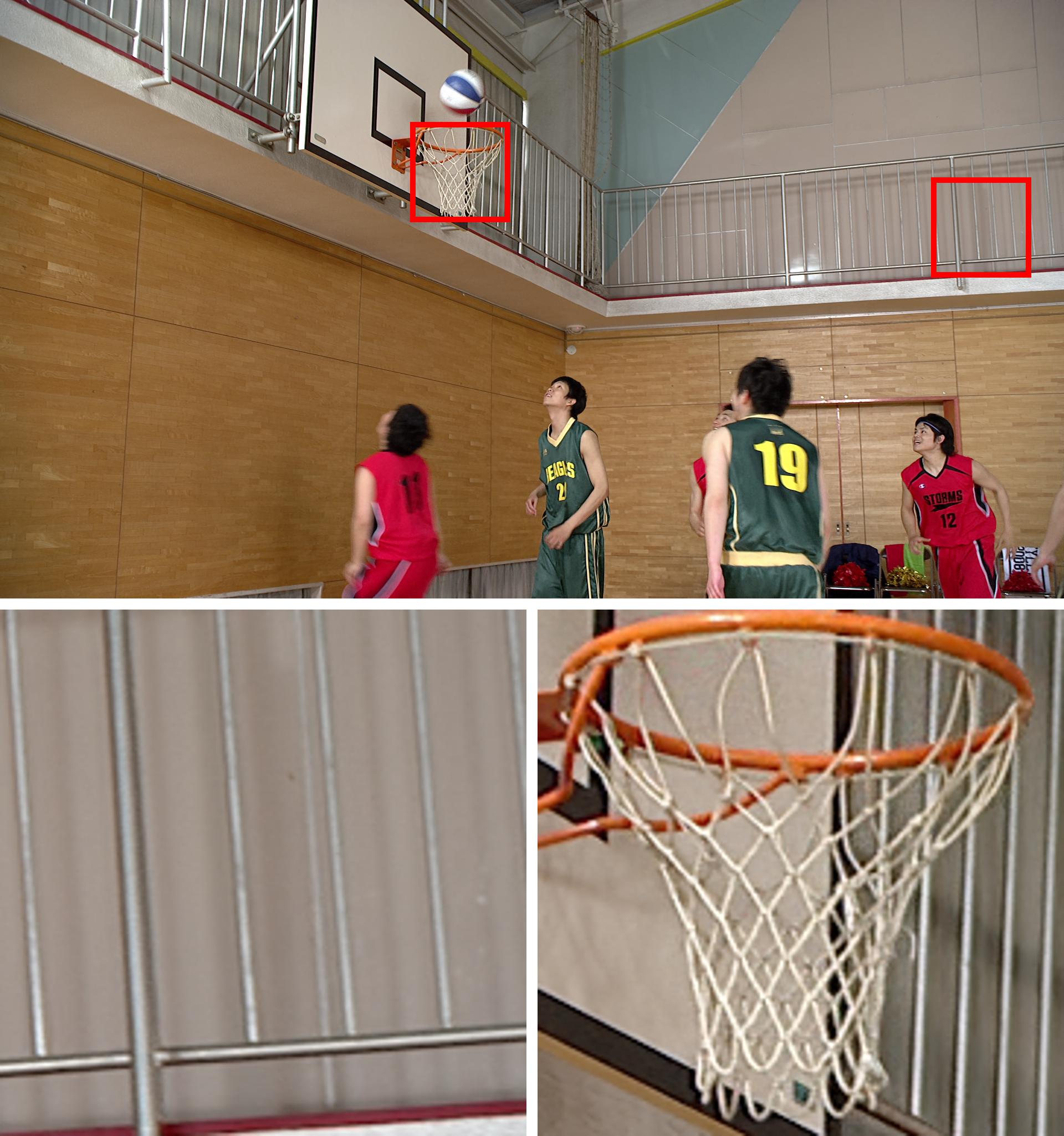} \\
    \includegraphics[width=0.245\textwidth]{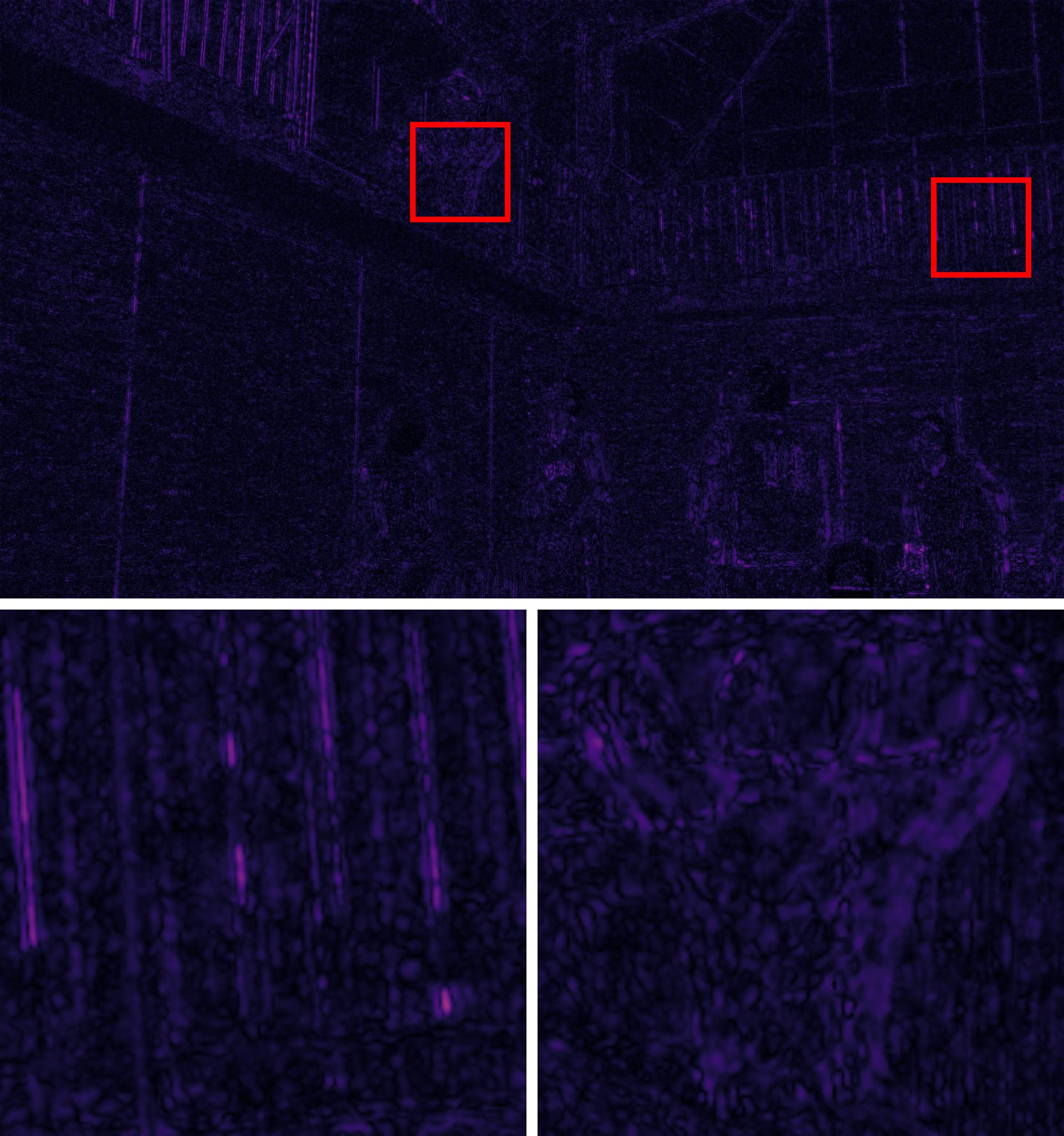} 
    & \includegraphics[width=0.245\textwidth]{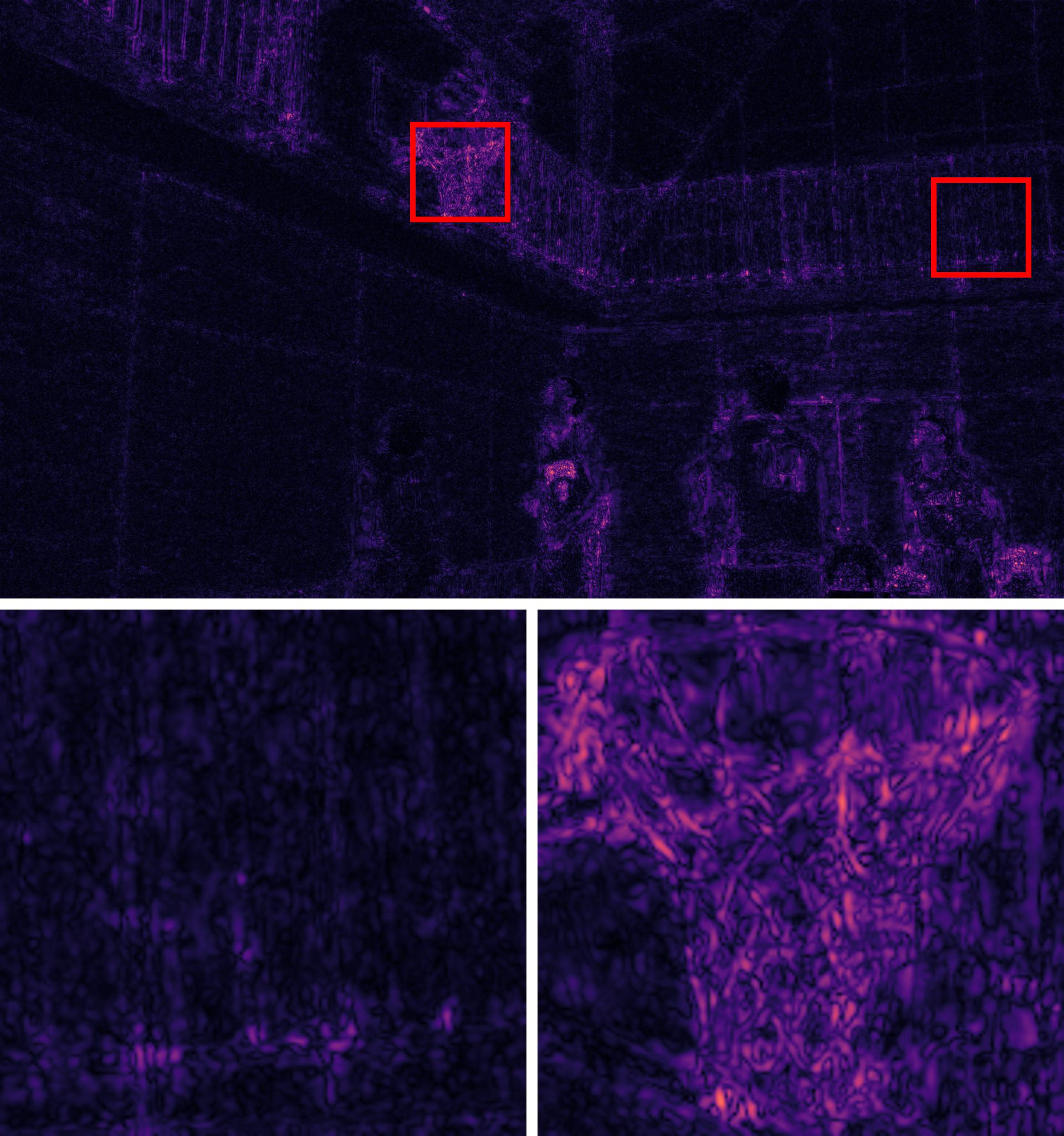} 
    & \includegraphics[width=0.245\textwidth]{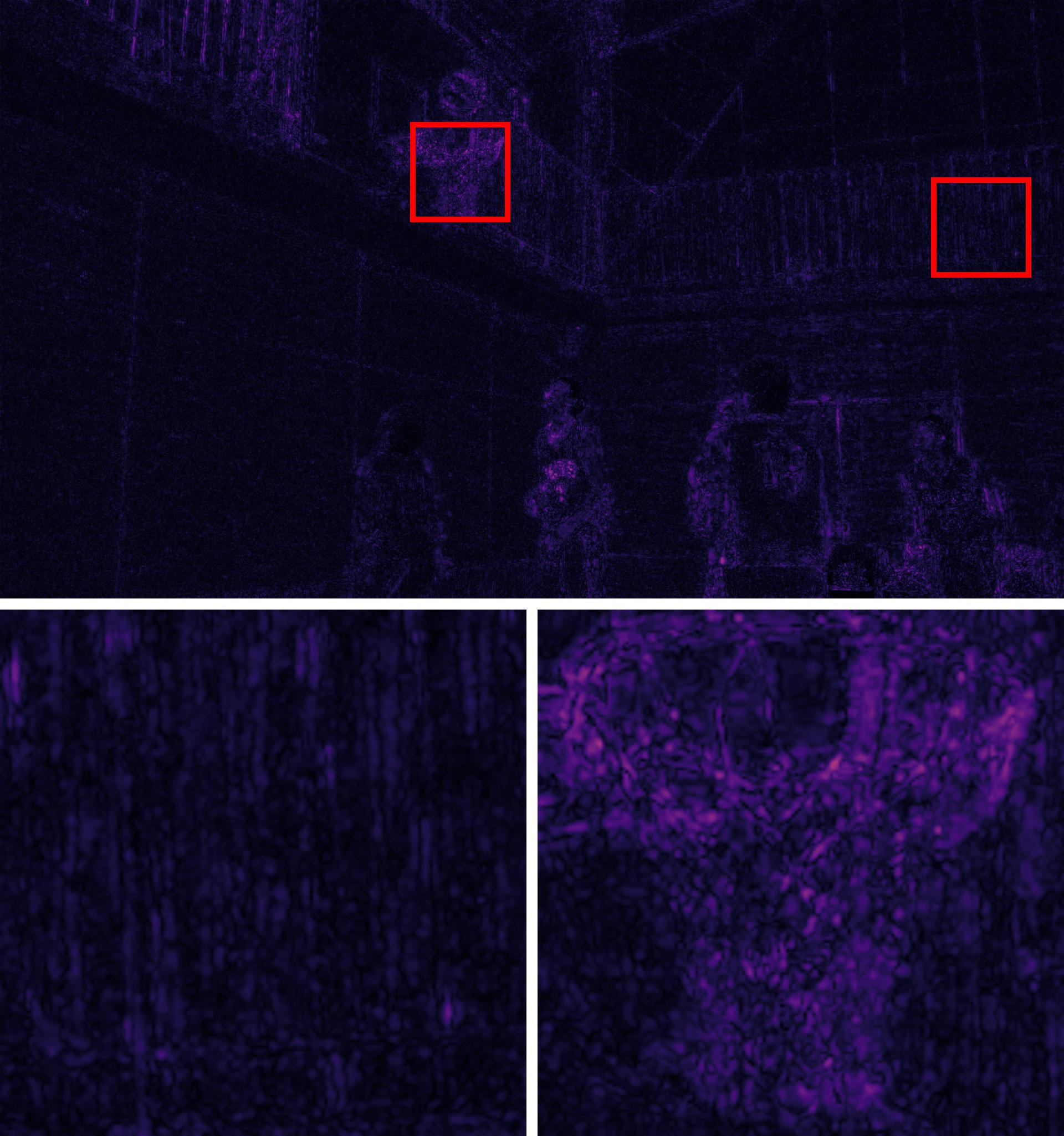} 
    & \includegraphics[width=0.245\textwidth]{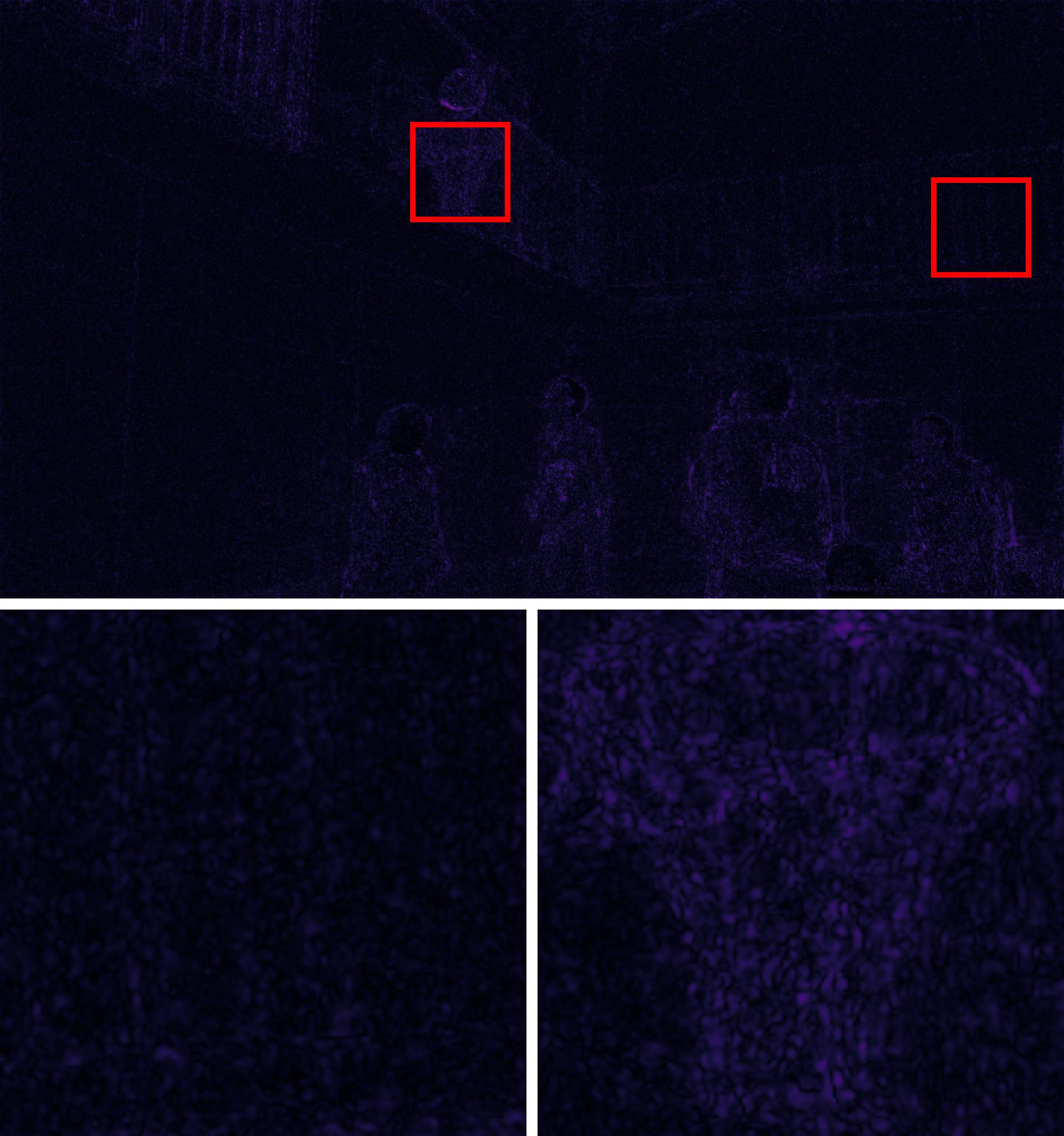} \\
    Instant-NGP \cite{muller2022instant} & 3D ModSIREN \cite{mehta2021modulated} & NVP \cite{kim2022scalable} & NVTM (Ours) \\
    \end{tabular} 
\end{small}
\end{table}

\paragraph{\textbf{Evaluation.}}
We evaluate with Peak Signal-to-Noise Ratio (PSNR) and Learned Perceptual Image Patch Similarity (LPIPS) \cite{zhang2018unreasonable}.
We compare our model both with grid-type models (fast encoding time) and NeRV-style models (efficient parameter size).
NeRV-style models, including NeRV~\cite{chen2021nerv}, E-NeRV~\cite{li2022nerv} and HNeRV~\cite{chen2023hnerv}, are reproduced by author's implementation. Since they have low bpp-levels, we also evaluate ours with 0.1bpp and compare with them. 
This experiment demonstrate real-world video settings, as Netflix recommends 5Mbps\footnotemark[1] as the minimum speed for Full HD video streaming, and 0.1bpp corresponds to a closely aligned bitrate of 4.97Mbps at 24 fps.\footnotetext[1]{\scriptsize \url{https://help.netflix.com/en/node/306}}
In addition, we experiment with grid-type models, including 3D ModSIREN, Instant-NGP~\cite{muller2022instant} and NVP~\cite{kim2022scalable}. 3D ModSIREN refers to the use of 3D grid as modulation latents in modulated-SIREN \cite{mehta2021modulated} without any dimension reduction, Instant-NGP is implemented by adjusting network size and NVP is reproduced according to author setting. Since the performance is dependent on target video scale, we additionally design them as a smaller parameter size for smaller resolution or short video length settings.

\subsection{Video Reconstruction: Encoding Speed}
\label{subsec:encoding_speed}
To apply on practical service, INR method must be quickly encoded. Then we first report the performance of the models when trained for 1/5/10/20/60 minutes under the same resource conditions on Figure~\ref{fig:encoding_time} and Table~\ref{tb:encodingtime_psnr}. All models are configured in 0.1bpp, as following their bit range. Our model prominently exhibits fast encoding and quickly reaches over 30dB compared to the NeRV series. NVP, which also uses grid-based parametric encoding, encodes quickly but its performance is inferior to ours.

\subsection{Video Reconstruction: Parameter Efficiency}
\label{subsec:param_efficency}
We also compared the performance of NVTM with other parametric encoding methods to explore how parameter efficient it is. Table~\ref{tb:mainresult_final} shows that NVTM outperforms on various video sequences. NVTM has 1.54dB/0.019 improvements of PSNR/LPIPS even with 10\% fewer parameters on UVG (Dynamic), and 1.84dB/0.013 improvements on MCL-JCV (Dynamic). %We also qualitatively compare on decoded image as visualized on Figure~\ref{fig:mainresult_vis}. 
From qualitative comparison on decoded images, we can observe how well our model preserves the fine details such as the thin iron bar and the basketball hoop. These results can be interpreted as NVTM has better parameter efficiency by dealing the coherent information of consecutive frames.

\begin{table}[!t]
\caption{\small  Video super resolution and frame interpolation on UVG (Dynamic). 
We expand spatial coordinates ($\times$2) and temporal coordinates ($\times$2) respectively on decoding time.
Below figures are cropped images and FLIP visualizations from super resolution results on Bosphorus sequence (left) and frame interpolation results on Jockey sequence (right).
The bright regions in FLIP figures represent errors, while the darker colors indicate better performance.
}
\label{tb:sr_fruc}
\centering
\begin{small}
\setlength{\tabcolsep}{10pt}
\begin{tabular}{c|cc|cc}
    \toprule
    \multirow{2}{*}{Model} & \multicolumn{2}{c|}{Super Resolution}& \multicolumn{2}{c}{Frame Interpolation} \\
    ~  & PSNR↑ & LPIPS↓ & PSNR↑ & LPIPS↓  \\ 
    \midrule
    3D ModSIREN \cite{mehta2021modulated}& 34.72 & 0.265 & 25.41 & 0.221  \\ 
    NVP \cite{kim2022scalable}& 31.87 & 0.396 & 23.88 & 0.394  \\ 
    NVTM (Ours) & \textbf{35.82} & \textbf{0.240} & \textbf{30.49} & \textbf{0.134}  \\ 
    \bottomrule
\end{tabular}
\newcommand{\myheight}{1.62cm}
\newcommand{\myoffset}{1.0cm}
\newcommand{\myscale}{0.9}
\setlength{\tabcolsep}{0.5pt}
\begin{tabular}{ccccccc}
    \centering
    ~ & ~ & \multicolumn{2}{c}{Super Resolution}& \multicolumn{2}{c}{Frame Interpolation} \\
    \raisebox{0.7cm}{\scalebox{\myscale}[\myscale]{\begin{sideways}3D\end{sideways}}}
    \scalebox{\myscale}[\myscale]{\begin{sideways}ModSIREN\end{sideways}}
    & ~
    & \includegraphics[height=\myheight]{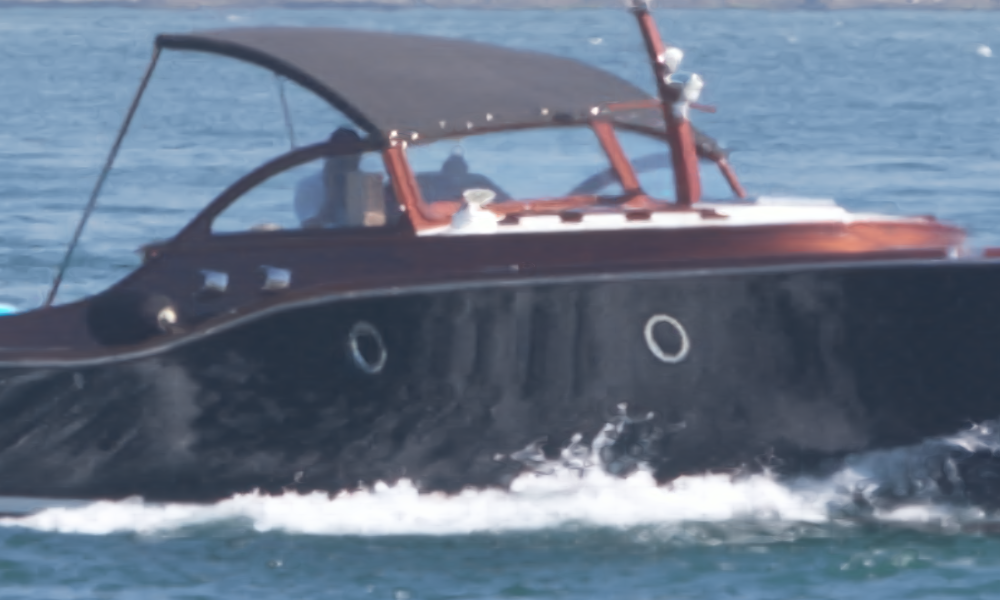} 
    & \includegraphics[height=\myheight]{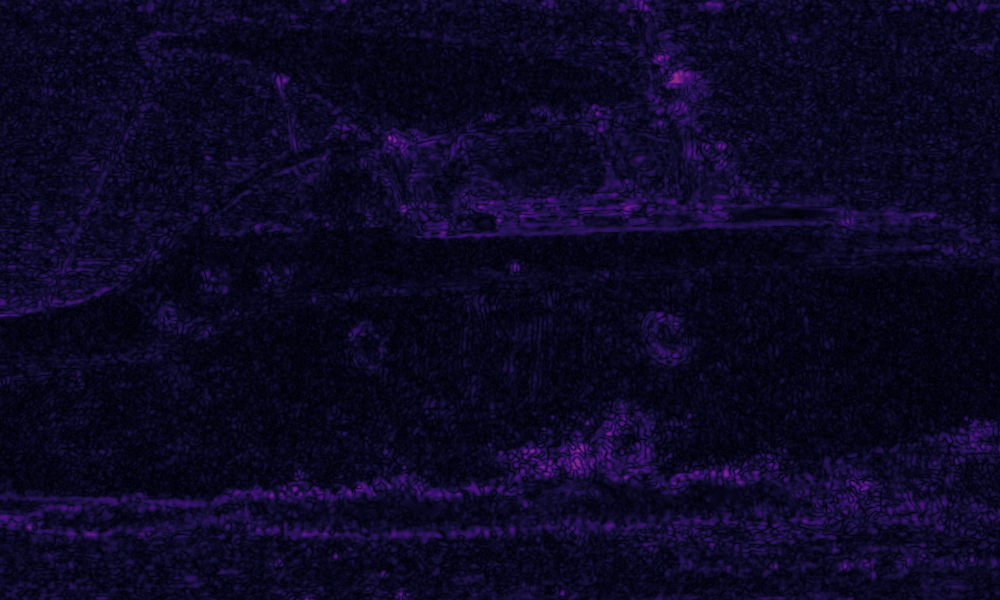}
    & \includegraphics[height=\myheight]{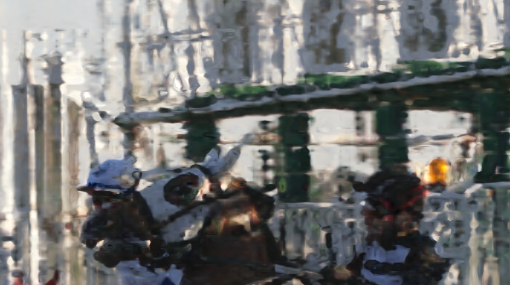} 
    & \includegraphics[height=\myheight]{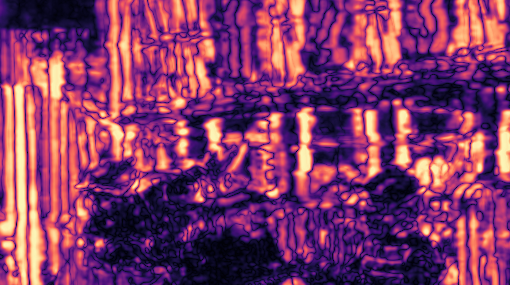}  \\
    % ~
    % & 
    \raisebox{\myoffset}{\multirow{2}{*}{\scalebox{\myscale}[\myscale]{\begin{sideways}NVP\end{sideways}}}}
    & ~
    & \includegraphics[height=\myheight]{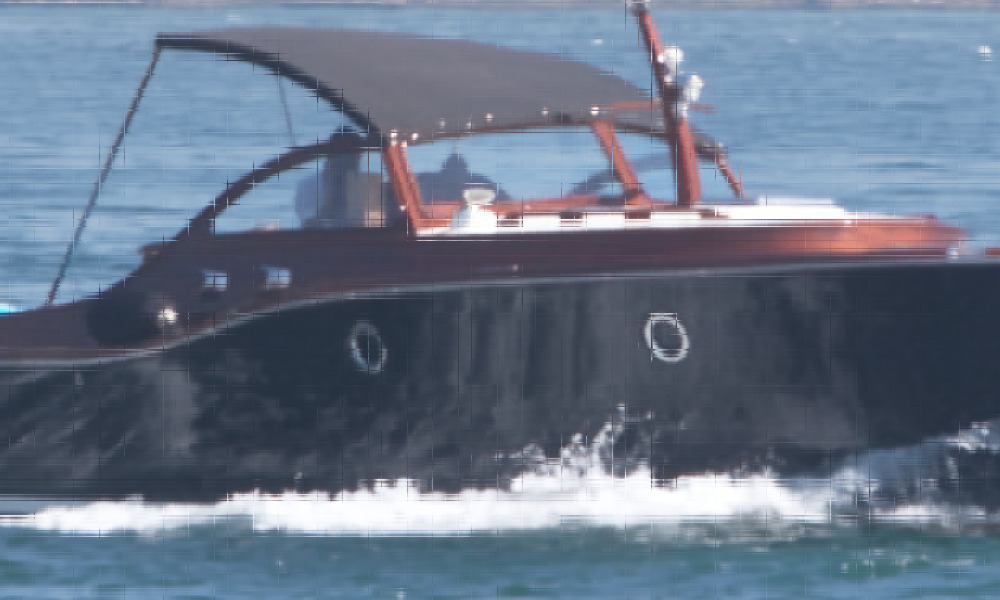} 
    & \includegraphics[height=\myheight]{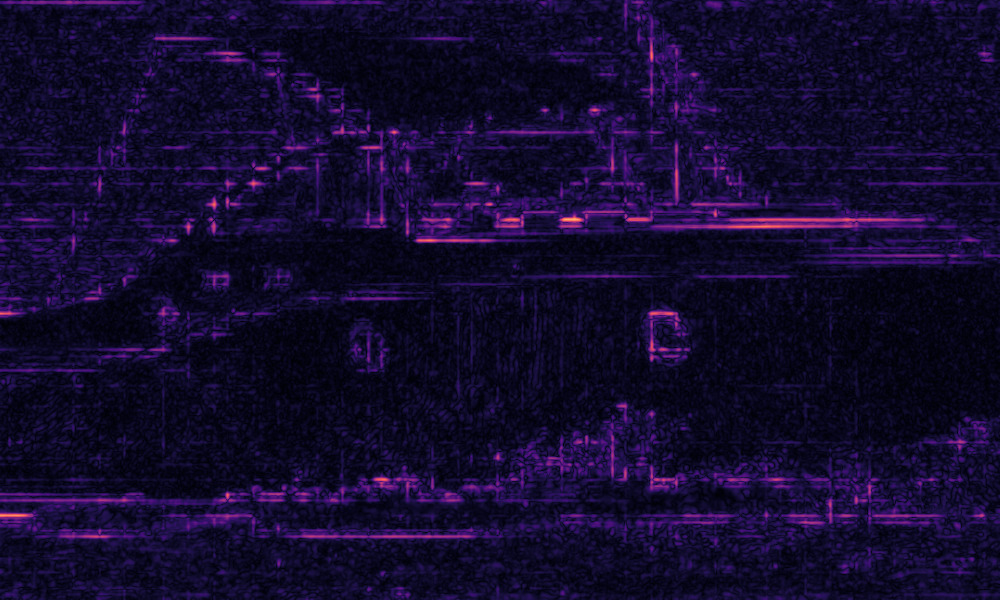}
    & \includegraphics[height=\myheight]{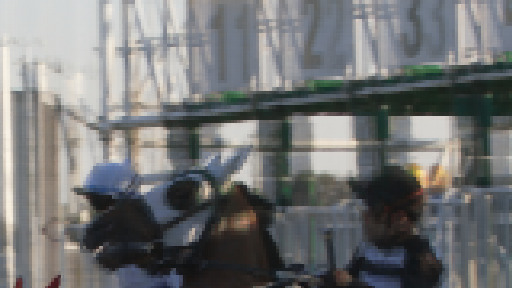} 
    & \includegraphics[height=\myheight]{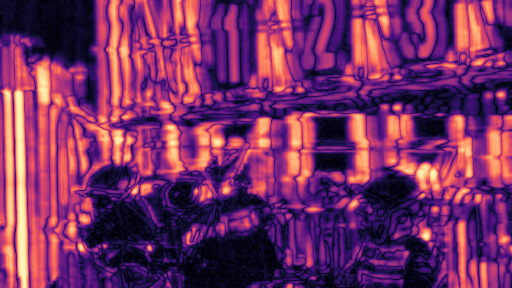} \\
    % ~
    % & 
    \raisebox{\myoffset}{\multirow{2}{*}{\scalebox{\myscale}[\myscale]{\begin{sideways}NVTM\end{sideways}}}}
    & ~
    & \includegraphics[height=\myheight]{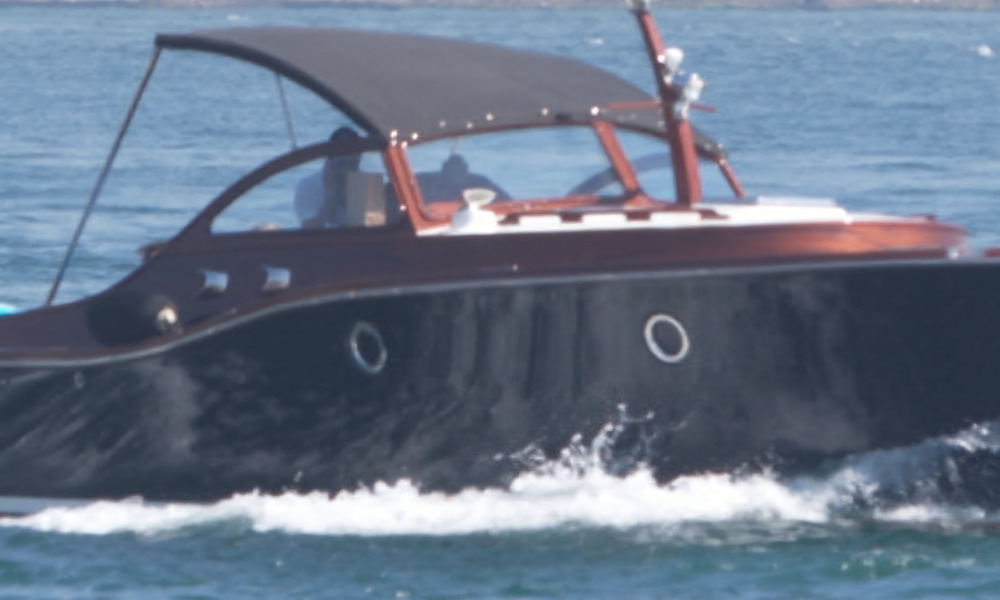} 
    & \includegraphics[height=\myheight]{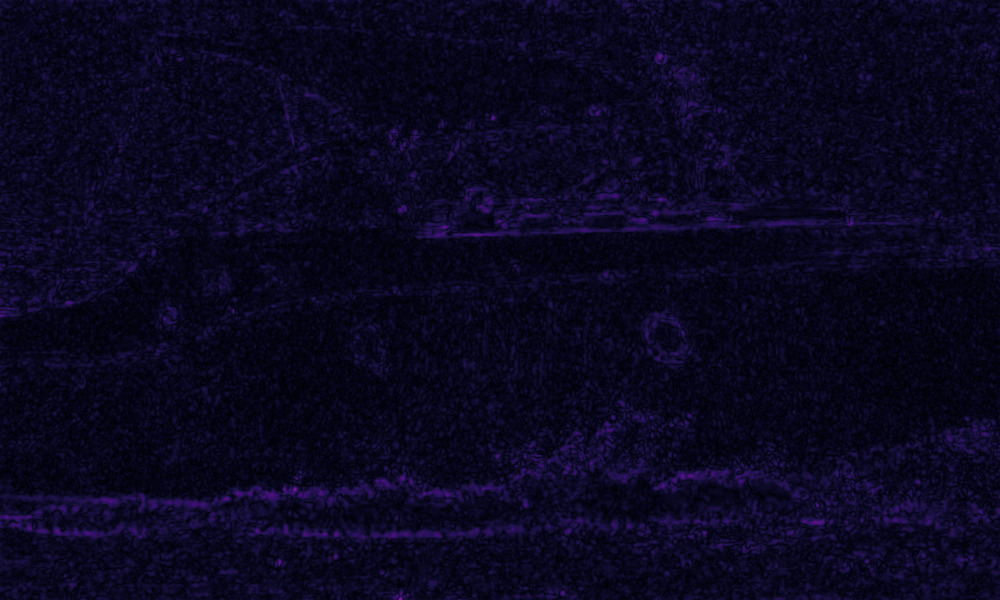}
    & \includegraphics[height=\myheight]{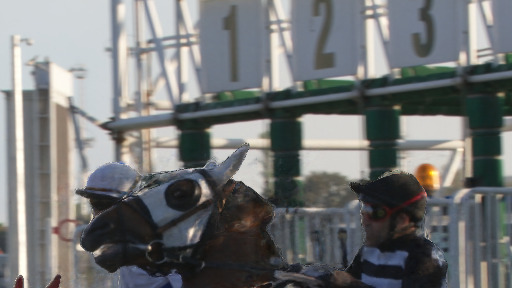}
    & \includegraphics[height=\myheight]{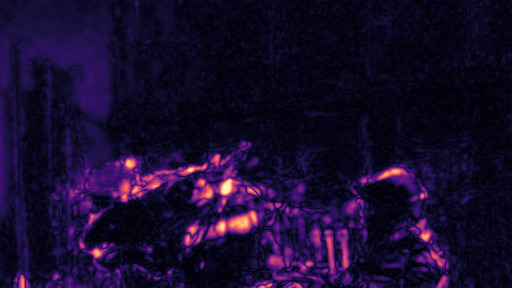} \\
\end{tabular}
\end{small}
\vspace{-0.2cm}
\end{table}

\subsection{Downstream Tasks}
\label{subsec:downstream}
\paragraph{\textbf{Video super resolution and frame interpolation.}}
One of the major advantages of INR is its capability to capture intermediary points in both temporal and spatial dimensions.
For video super resolution, we decode all models with doubled spatial coordinate and evaluate with early-defined 4K resolution videos (T,H,W→T,2H,2W). Similarly, for video frame interpolation, we first train models with odd number images and decode them with doubled temporal coordinate and evaluate with original video sequence (T/2,H,W→T,H,W).
The evaluated results are in Table~\ref{tb:sr_fruc}, NVTM shows much fewer errors for both intermediate spatial and temporal values than others. 
Meanwhile, since 3D ModSIREN can densely encode pixels utilizing 3D coordinates directly, it might be slightly advantageous in generating intermediate values and outperforms NVP. However, our approach, despite not using 3D-shaped grid parameters, demonstrates impressive results, indicating its successful decomposition of 3D video data.

\begin{figure}[!hbpt]
\centering
    \begin{subtable}[t]{0.56\textwidth}
        \centering
        \caption{Video Inpainting}
        \label{fig:inpainting}
        \begin{tabular}{cc}
            \includegraphics[width=0.43\linewidth]{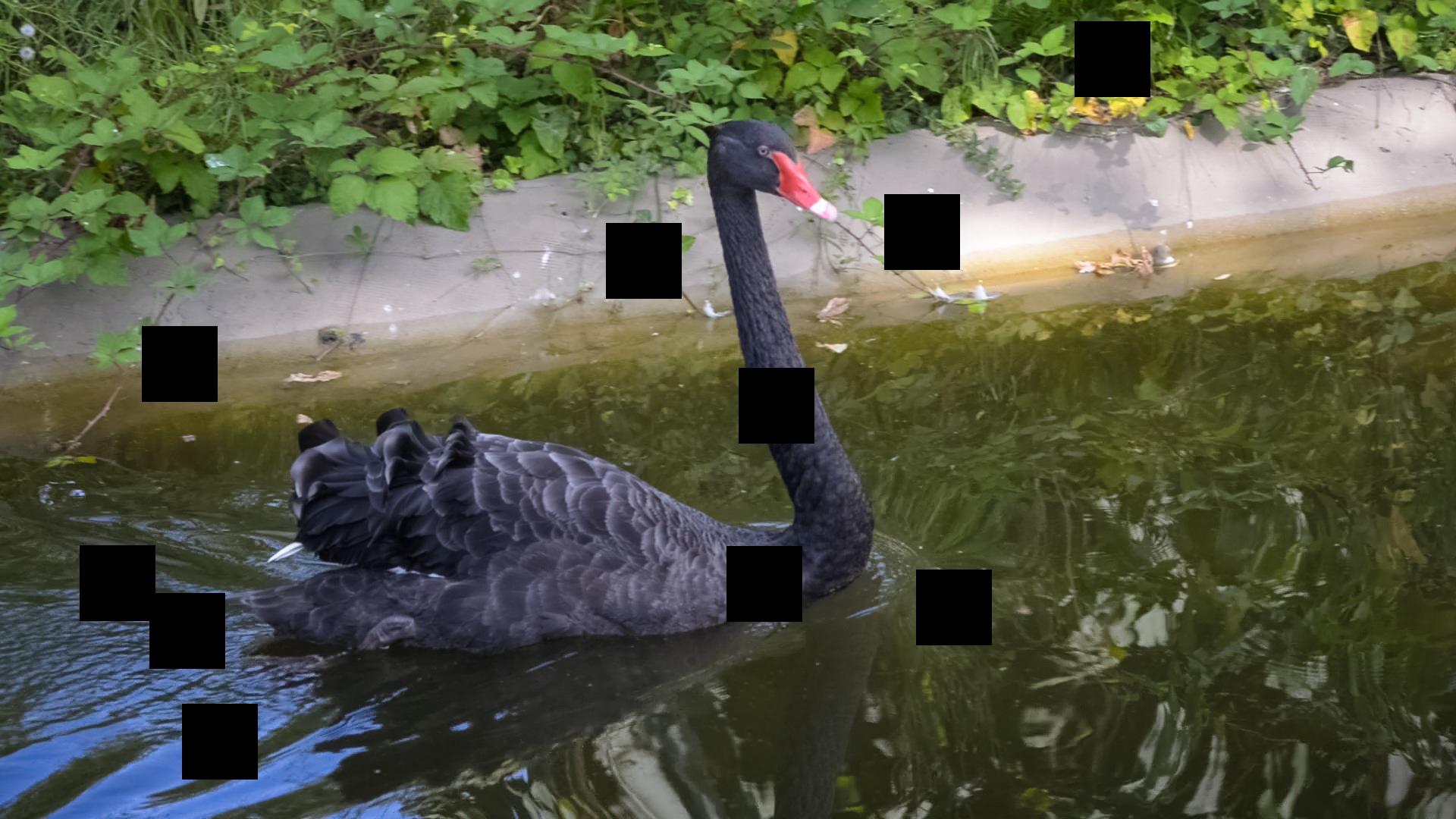} 
            &\includegraphics[width=0.56\linewidth]{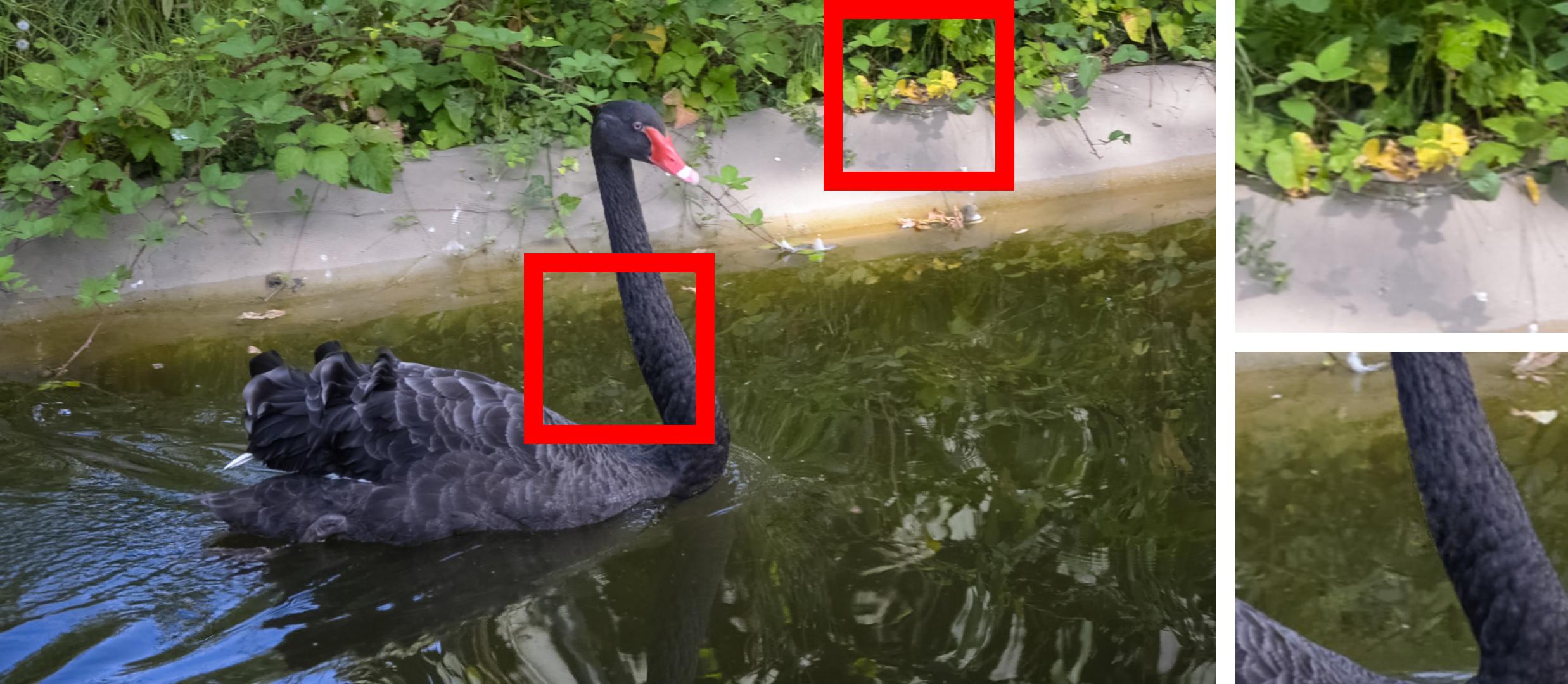} \\
            \includegraphics[width=0.43\linewidth]{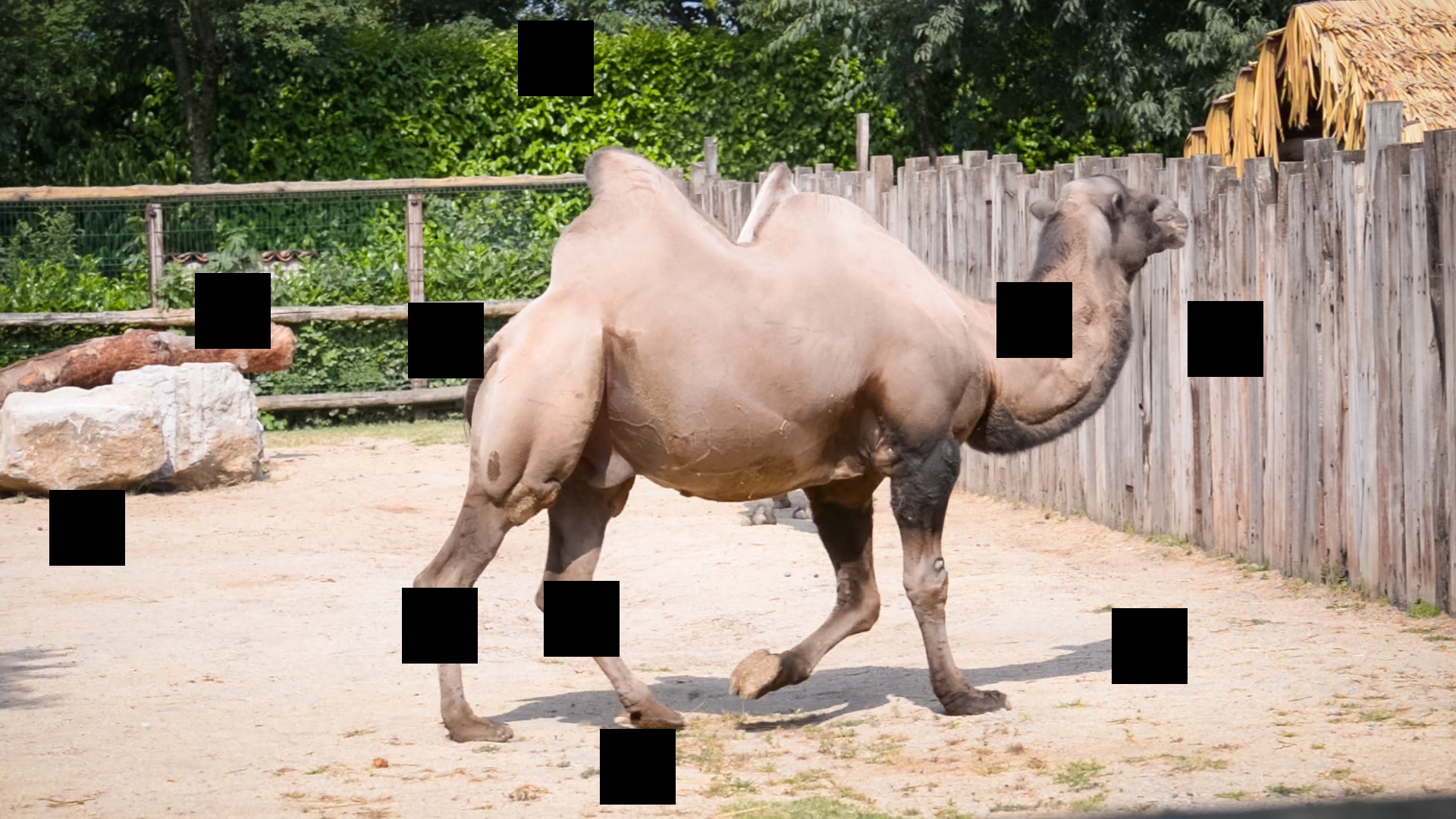} 
            &\includegraphics[width=0.56\linewidth]{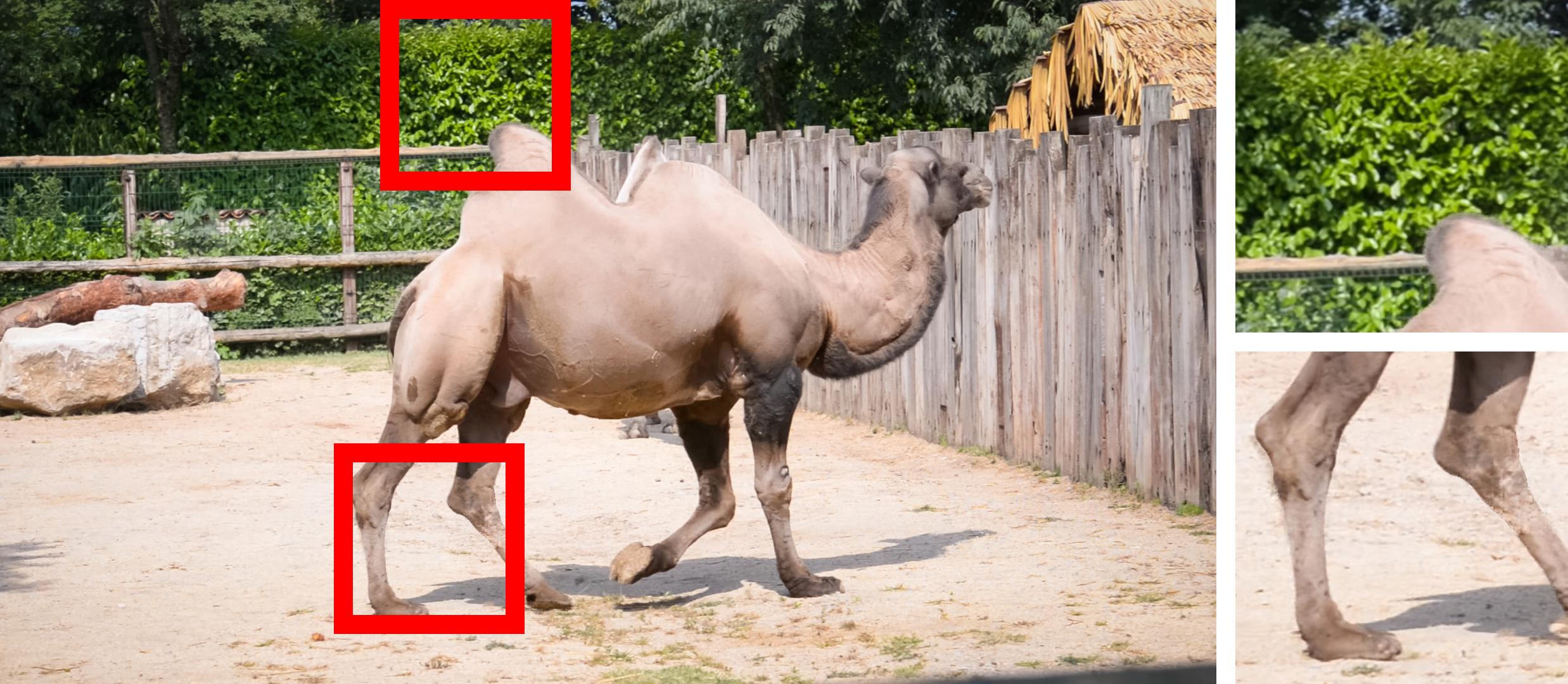} \\
         \end{tabular} 
    \end{subtable}
    \begin{subtable}[t]{0.43\textwidth}
        \centering
        \caption{Video Compression}
        \label{fig:compression_vis}
        \begin{tabular}{c}
        \includegraphics[width=\linewidth]{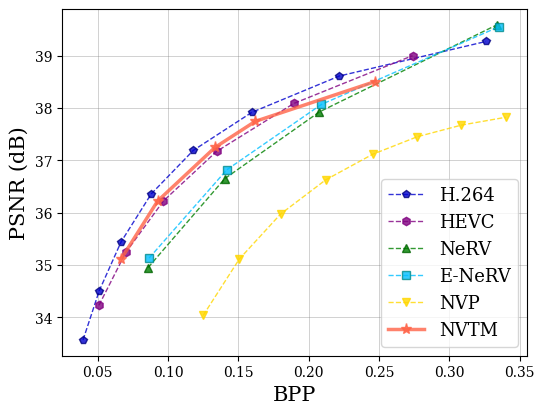}
        \end{tabular}
    \end{subtable}
\vspace{-0.5cm}
\caption{\small Video inpainting and compression performance.
(a) Visualization of video inpainting on Blackswan and Camel sequences in DAVIS2017. Although the masked regions are excluded during encoding, the NVTM successfully decodes them by utilizing temporally coherent modulation latent from adjacent frames.
(b) BPP-PSNR plot of video compression on UVG (Dynamic). We encode all models with each video sequence and evaluate as following authors guided.}
\vspace{-0.2cm}
\end{figure}

\paragraph{\textbf{Video inpainting.}}
We further explore the potential of NVTM in the video inpainting task. We use DAVIS2017 \cite{DAVIS2017} HD dataset. We conduct a random box experiment, training with random box masked images and targeting to reconstruct the complete frames, as previous works \cite{kim2022scalable}. We generate masked images with 10 random boxes masking with $100 \times 100$ sized on every frame.
As seen in Figure~\ref{fig:inpainting}, our method exhibits a remarkable restoration performance on the masked regions. This highlights that our representation is learned as aggregated by reference on similar pixels along the temporal axis.

\paragraph{\textbf{Video compression.}}
Video compression is one of main applications in INR for video \cite{chen2021nerv, li2022nerv, kim2022scalable, chen2023hnerv, zhao2023dnerv, he2023towards, gomes2023video}, and they attempt to prune, quantize, or compress the model parameters after training.
We also compress model parameter by applying existing codecs, as grid-type INR approaches tried \cite{kim2022scalable, muller2022instant}. Especially, since our model is decomposed with a series of 2D grids notated as $\text{G} := (G_{1},...,G_{m})$, we applied HEVC video compression on the grid parameters and further compress effectively.
In Figure~\ref{fig:compression_vis}, we compare our model with standard video codecs (H.264 \cite{wiegand2003overview}, HEVC \cite{sullivan2012overview}) and state-of-the-art methods \cite{chen2021nerv, li2022nerv, kim2022scalable, he2023towards}. 
The details of the evaluation are described in supplementary. %Section~\ref{sec:sup_compression}.
NVTM demonstrates compression performance similar or slightly better than \cite{chen2021nerv, li2022nerv} which have lower training speeds, and notably outperforms \cite{kim2022scalable} which has faster training speed as ours. Here, we can confirm the superiority of NVTM when considering both parameter-efficiency and computing-efficiency.

\begin{table}[t]
\caption{\small Ablation study on framework design in UVG (Dynamic).}
\vspace{-0.1cm}
\label{tb:ablation}
\setlength{\tabcolsep}{4pt}
    \begin{subtable}[t]{0.5\textwidth}
        \centering
        \caption{Framework design}
        \label{subtable:modules}
        \begin{tabular}{ccc}
        \toprule
            Adaptive norm. & Static feature & PSNR  \\ 
            \midrule
            x & x & 40.24  \\ %\hline
            $\checkmark$ & x & 40.36  \\ %\hline
            $\checkmark$ & $\checkmark$ & \textbf{40.54}  \\ 
            \bottomrule
        \end{tabular}
    \end{subtable}
    \begin{subtable}[t]{0.25\textwidth}
        \centering
        \caption{Latent process}
        \label{subtable:modulation}
        \begin{tabular}{cc}
            \toprule
            Modulation & PSNR  \\ 
            \midrule
            x & 38.90  \\ %\hline
            $\checkmark$ & \textbf{40.54}  \\ 
            \bottomrule
        \end{tabular}
    \end{subtable}
    \begin{subtable}[t]{0.23\textwidth}
        \centering
        \caption{Index set}
        \label{subtable:reference}
        \begin{tabular}{cc}
            \toprule
            $P$ & PSNR  \\ 
            \midrule
            $\{0\}$ & 39.45  \\ %\hline
            $\{0, 1\}$ & \textbf{40.54}  \\ 
            \bottomrule
        \end{tabular}
    \end{subtable}
\vspace{-0.3cm}
\end{table}

\begin{table}[!t]
\caption{ \small Ablation study on GOP size. 
The performance for each video sequence vary depending on the GOP size.
As the smaller size of GOP results in dividing video sequences into more segments for coordinate alignment, we modify the model configuration to ensure similar overall model parameters in each experiment. 
}
\label{tb:abl_gop}
\centering
\setlength{\tabcolsep}{4pt}
\begin{small}
    \begin{tabular}{c|cccc}
    \toprule
        GOP & Bosphorus & Jockey & ReadySetGo & YachtRide  \\ 
        \midrule
        5 & 43.27 & \textbf{40.41} & \textbf{39.80} & \textbf{39.57}  \\ %\hline
        10 & \textbf{43.30} & 40.17 & 39.70 & 38.88  \\ %\hline
        20 & 43.00 & 39.29 & 38.54 & 35.65 \\ %\hline 
        60 & 40.34 & 35.39 & 33.50 & 32.53 \\
    \bottomrule
    \end{tabular}
\end{small}
\vspace{-0.3cm}
\end{table}

\begin{table}[!t]
    \setlength{\tabcolsep}{4pt}
    \centering
    \begin{small}
    \caption{\small  Temporal scalability on video length. 
    Experiments on video lengths exceeding 600 frames are conducted on videos composed of concatenated sequences, each labeled according to the initial letter of sequence names.}
    \label{tb:temporal_scalability}
    \centering
    \begin{tabular}{c|cccccc}
        \toprule
        Model & 100 & 200 & 300 & 600 & 1200 {\tiny (B+J)} & 2400 {\tiny (B+J+R+Y)}\\ 
        \midrule
        3D ModSIREN \cite{mehta2021modulated} & 38.48 & 39.37 & 40.06 & 40.92 & 37.37 &  38.82 \\ 
        NVP \cite{kim2022scalable} & 41.23 & 41.08 & 41.12 & 41.47 & 40.23 & 40.21\\ 
        NVTM (Ours) & \textbf{43.70} & \textbf{43.54} & \textbf{43.50} & \textbf{43.41} & \textbf{41.54} & \textbf{41.05} \\ 
        \bottomrule
    \end{tabular}
    \end{small}
\vspace{-0.3cm}
\end{table}

\subsection{Ablations Studies}
\label{sec:ablation}
\paragraph{\textbf{Framework design}}
We conduct ablation studies on our framework modules.
Table~\ref{subtable:modules} demonstrates the positive impacts of adaptive normalization and static features, with improvements of 0.12dB and 0.19dB respectively.
We compare the effectiveness of using latent as a modulation to the base network versus using it as a direct input. Table~\ref{subtable:modulation} indicates that using it as a modulation is more effective for representing video.
Also we experiment on effect of neighbor index set $P$ on Table~\ref{subtable:reference}.

\paragraph{\textbf{GOP size}}
Our framework uses a fixed GOP size to divide the video into segments.
We experiment with different GOP sizes on Table~\ref{tb:abl_gop}. 
% We found that using a GOP size of 5 has better performance compared to 10 in some video sequences.
From our analysis about the degree of motion for each sequence (described on Section~\ref{subsection:dynamic_selection}), we find that some sequences with relatively large motion energy exhibit improved performance when the video is divided more finely with a GOP value of 5.
Conversely, for sequence which has relatively small motion, a GOP value of 10 yielded better performance.
From this tendency, we believe that NVTM can achieve better performance if it uses variable size of GOP.

\paragraph{\textbf{Video duration}}
We verify temporal scalability that NVTM consistently achieves the standout performance across various video lengths on Table~\ref{tb:temporal_scalability}.

\begin{figure}[t]
\centering
    \begin{subfigure}{0.50\linewidth}
    \includegraphics[width=\linewidth]{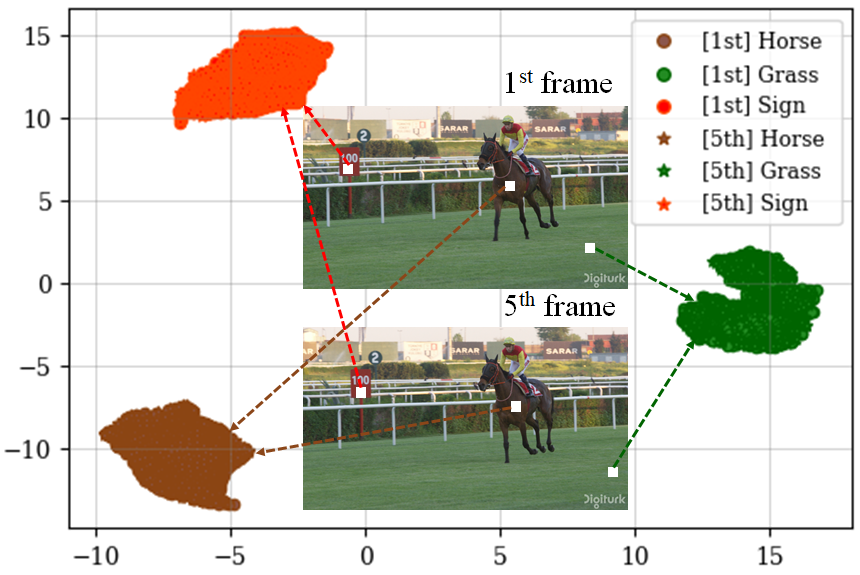} 
    \caption{Temporal consistency analysis}
    \label{fig:tempconsistent}
    \end{subfigure}
    \begin{subfigure}{0.48\linewidth}
    \includegraphics[width=\linewidth]{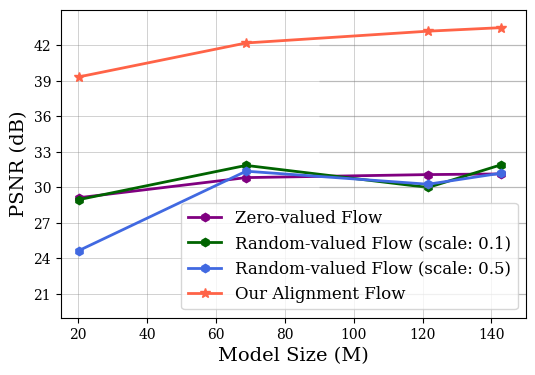}
    \caption{Alignment flow analysis}
    \label{fig:tca_working}
    \end{subfigure}
\caption{\small 
(a) t-SNE visualization of modulation latent $z_{xyz}$ from our alignment module on corresponding pixels (1${st}$ and 5${th}$ frame). We select areas with similar pixel information, i.e. RGB values, and for ease of verification, these are denoted as \{Horse, Grass, Sign\}. The latent derived from the 1${st}$ frame and 5$th$ are marked with circle and star respectively. The analysis is based on segments, each consisting of 400 pixels. 
(b) Effects of alignment flow. 
Each line represents the performance with replacing our alignment method on Bosphorus sequence.
\textcolor{Purple}{\textbf{Purple}} indicates aligning with zero-valued flow (i.e., its spatial coordinate).
\textcolor{OliveGreen}{\textbf{Green}} and \textcolor{RoyalBlue}{\textbf{blue}} indicates aligning with random-valued flow in a notated scale of source video resolution.}
\label{fig:tempconsistent}
\end{figure}

\subsection{Analysis}
\paragraph{\textbf{Temporal Consistency Modulation}}
We propose that by assigning the same modulation latent ($z_{xyz}$) to similar pixels across consecutive frames, the network could learn more rapidly and achieve higher performance. To confirm this, we analysis on $z_{xyz}$ corresponding to pixels that appeared to be similar in Figure~\ref{fig:tempconsistent}.
We can observe that the latent values derived from similar pixel areas across different frames are represented as similar embeddings.
These findings validate our intention that our network produces identical modulation latents from similar pixels in consecutive frames.

\paragraph{\textbf{Alignment Flow}}
We propose to align the 3-dimensional $(x,y,t)$ to the 2-dimensional $(x,y)$, using an alignment flow derived from Equation~\ref{eq:flow_hyper}. To validate the effectiveness and usefulness of this method, we compare it with other alignment methods in Figure~\ref{fig:tca_working}. Both zero-valued flow and the random-valued flow, unlike our method, does not consider motion or pixel similarity, and simply map the video into 2D. 
From the results, we can verify that the proposed method demonstrated sufficient performance (much over 30dB) even with fewer parameters, whereas other methods were significantly deficient in performance.

\section{Conclusion}
In this study, we proposed a novel approach for implicit neural video representation, which involves temporal coordinate alignment and modulation latent encoding to effectively capture video dynamics at the pixel level. Extensive experiments verified that the NVTM outperforms existing methods of implicit neural video representation on various video related tasks. We anticipate that our framework will provide inspiration for the follower on INR for videos.
% \subsection*{Acknowledgements}

\clearpage
%%%%%%%%%%%%%%%%%%%%%%%%%%%%%%%%%%%%%%%%%%%%%%%%%%%%%%%%%%%%%%%%%%%%%%%%%%%%%%%%%%%%%%%%%%%%%%%%%%%%%%
%%%%%%%%%%%%%%%%%%%%%%%%%%%%%%%%%%%%%%%%%%%%%%%%%%%%%%%%%%%%%%%%%%%%%%%%%%%%%%%%%%%%%%%%%%%%%%%%%%%%%%

\bibliographystyle{splncs04}
\bibliography{main}

%%%%%%%%%%%%%%%%%%%%%%%%%%%%%%%%%%%%%%%%%%%%%%%%%%%%%%%%%%%%%%%%%%%%%%%%%%%%%%%%%%%%%%%%%%%%%%%%%%%%%%
%%%%%%%%%%%%%%%%%%%%%%%%%%%%%%%%%%%%%%%%%%%%%%%%%%%%%%%%%%%%%%%%%%%%%%%%%%%%%%%%%%%%%%%%%%%%%%%%%%%%%%

% ---- Supplementary ----
%
% Supplementary

\clearpage  
\appendix
\section{Dataset Description}
\label{sec:sup_dataset}
\subsection{Dataset}
% \paragraph{\textbf{UVG Dataset.}} 
\paragraph{\textbf{UVG (Dynamic).}} 
The commonly used UVG dataset \cite{mercat2020uvg} on video reconstruction is composed of six sequences with 600 frames and one sequence with 300 frames, all with a resolution of 1920$\times$1080. 
As a default dataset setting, we conduct our experiments on 600-frame Bosphorus, Jockey, ReadySetGo, and YachtRide sequences, which are selected following the procedures described in Section~\ref{subsection:dynamic_selection}. Our results also include experiments conducted with various settings, such as changes in resolution and frame count.
% \paragraph{\textbf{MCL-JCV.}}
\paragraph{\textbf{MCL-JCV (Dynamic).}}
The MCL-JCV dataset \cite{wang2016mcl} consists of 24 videos with resolution 1920$\times$1080.
Each sequence in MCL-JCV consists of 120 - 150 frames, and we use the first 100 frames from each sequence.
We conduct our experiments on 04, 05, 11, 20, and  21 sequences, which are selected following the procedures described in Section~\ref{subsection:dynamic_selection}.

\subsection{Dynamic sequence selection}
\label{subsection:dynamic_selection}
Since our method is designed to capture temporal dynamic nature of video, we aim to demonstrate its effectiveness through a variety of videos that contain at least minimal spatial and temporal information. 
To address this, we refer to spatial perceptual information (SI) and temporal perceptual information (TI) scores based on ITU P.910 \cite{installations1999subjective} and motion energy (ME) calculated using optical flow as criteria.
SI indicates the maximum amount of spatial details over times, which is calculated by standard deviation over the pixels in each Sobel filtered frame. TI indicates the maximum amount of temporal variation between successive frames based on motion difference feature. SI and TI are calculated as 
\begin{equation}
    \centering
    \begin{aligned}
    \text{SI} = \text{max}_{time}\{\text{std}_{space}[\text{Sobel}(F_n)]\}, \\
    \text{TI} = \text{max}_{time}\{\text{std}_{space}[F_n(i,j)-F_{n-1}(i,j)]\},
    \end{aligned}
\end{equation}
where $F_{n}$ represents $n$-th frame.
ME is an average of optical flow scales (OF) which are calculated by optical flow network on successive frames, which standing for motion scales in a video. 
ME is calculated as 
\begin{equation}
    \centering
    \begin{aligned}
    \text{ME} = \text{avg}_{time}\{\text{\small{OF}}(F_n, F_{k})\}, \\
    \end{aligned}
\end{equation}
where $F_{k}$ represents a predefined certain frame in each GOP.
We select target sequences with significant ME score in the UVG and MCL-JCV datasets after excluding those with SI and TI scores less than 30 and 20, respectively. Table~\ref{tb:dataset_stats} shows the selected sequences we use, and their statistics.

\begin{table}[h]
\centering
\setlength{\tabcolsep}{3.5pt}
\begin{footnotesize}
    \caption{\small Statistics on video sequences in UVG and MCL-JCV.
    Upper visualization indicates SI, TI and ME statistics for every video sequence, X-axis and Y-axis indicate SI and YI respectively, and a circle size indicates ME scale. We select target sequences which have high ME among where SI and TI exceeds gray-dashed borderline, and they are colored with darker colors.
    }
    \label{tb:dataset_stats}
    \begin{tabular}{c|c|c|rrr}
    \multicolumn{6}{c}{\includegraphics[width=0.60\linewidth]{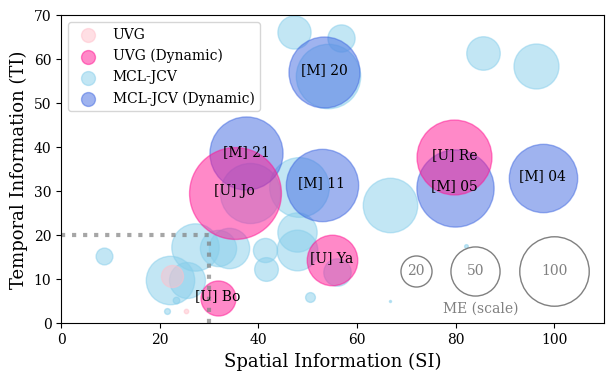}} \\
    \hline
    \multirow{2}{*}{Dataset} & \multirow{2}{*}{Sequence} & \multirow{2}{*}{Frames} & \multicolumn{3}{c}{Statistics} \\
    & & & ME & SI & TI \\ \hline
    % Dataset & Sequence & Frames & ME & SI & TI \\ \hline
    \multirow{4}{*}{UVG (Dynamic)} & Bosphorus & 1-600 & 25.43 & 31.71 & 5.84 \\
    ~ & Jockey & 1-600 & 175.01 & 35.17 & 29.65 \\
    ~ & ReadySetGo & 1-600 & 116.26 & 79.71 & 37.77 \\
    ~ & YachtRide & 1-600 & 53.00 & 54.93 & 14.42 \\ \hline
    \multirow{5}{*}{MCL-JCV (Dynamic)} & videoSRC04 & 1-100 & 96.90 & 97.61 & 33.05 \\
    ~ & videoSRC05 & 1-100 & 123.54 & 79.78 & 30.86 \\
    ~ & videoSRC11 & 1-100 & 108.47 & 52.85 & 31.43 \\
    ~ & videoSRC20 & 1-100 & 104.56 & 53.36 & 57.07 \\
    ~ & videoSRC21 & 1-100 & 110.78 & 37.50 & 38.64 \\
    \hline
    \end{tabular}%
\end{footnotesize}
\end{table}

\subsection{Data preparation}
Following prior works \cite{chen2021nerv, kim2022scalable}, we use FFmpeg \cite{tomar2006converting} to extract RGB frames from the raw YUV videos in both the UVG and MCL-JCV datasets. 
\begin{footnotesize}
\begin{verbatim}
ffmpeg -f rawvideo -vcodec rawvideo -s 1920x1080 -r 120 \
-pix_fmt yuv420p -i INPUT.yuv OUTPUT/%05d.png
\end{verbatim}
\end{footnotesize}

\section{Implementation Details of NVTM}
% \textcolor{\ssjwritecolor}{Our framework consists of the alignment flow network, latent grids, and a base network as described on Section~\ref{sec:Methodology}.}
Our framework consists of the \texttt{alignment flow network}, 
\texttt{latent grids}, \\ a \texttt{static module}, and a \texttt{base network} as described on Section~\ref{sec:Methodology}. 
In this section, we describe model configuration and compression scheme in detail. 
the \texttt{latent grids} and the \texttt{static module} are implemented as 2D grids using DenseGrid in the tiny-cuda-nn \cite{tinycudann}.

\subsection{Model configuration} 
The specific implementation configuration for each module is as follows.
Each GOP unit of the \texttt{alignment flow network} is configured with Hyper-SIREN \cite{figueiredo2023frame}, which is composed of a single-layer hyper-network and a 5-layer network, with each layer consisting of 8 neurons.
We set the \texttt{static module} as a 2D grid configured with a base resolution of (16$\times$16), 16 levels, a scale of 1.35, and 2 features per level.
Each GOP unit of \texttt{latent grid} is composed of a 2D grid configured with a base resolution of (16$\times$16), 7 levels, a scale of 1.8, and 4 features per level.
The \texttt{base network} is composed of three layers with 185 neurons.
\newline \indent Additionally, we provide a small-sized configuration to target low bitrate (0.1bpp): \texttt{latent grid} as a 2D grid configured with a base resolution of (16$\times$16), 5 levels, a scale of 1.8, and 2 features per level, and the \texttt{base network} as three layers with 165 neurons.
% \newline\noindent We provide two sizes of configuration for \texttt{latent grids} and \texttt{base network}. 
% \newline \noindent \textbf{small}: Each GOP unit of \texttt{latent grid} is composed of a 2D grid configured with a base resolution of (16$\times$16), 5 levels, a scale of 1.8, and 2 features per level.
% The \texttt{base network} is composed of three layers with 165 neurons.
% \newline \noindent \textbf{base}: 

\subsection{Compression scheme}
After the training process, all parameter weights of NVTM are quantized into 8-bit, and additional compression was performed on the \texttt{latent grids}.
% After training, 8-bit quantization is proceed on all parameter weights of NVTM. 
% We added additional compression to the model parameter of grids in TCA.
% We added additional compression to the parameter latent grids.
We regard the series of 2D grids as a video which contains temporally related information. Therefore, we employ HEVC \cite{sullivan2012overview} compression to further reduce bpp.
% To apply the compression on the multi-resolution grid series, compression was performed separately for each resolution, as the following command.
For this, we first save each model parameters of the same resolution (i.e., same level of multi-resolution grids) into individual folders (e.g., folder\_$i$). Then they  are compressed as following command:
\begin{footnotesize}
\begin{verbatim}
for (level=1; level<=${n_levels}; level++)
do
   ffmpeg -framerate 25 -i folder_${level} -c:v hevc \
    -x265-params "bframes=0" -crf {crf} ${level}.mp4
done
\end{verbatim}
\end{footnotesize}
We can reduce the model's bitrate to match the target bpp by adjusting the Constant Rate Factor (CRF) of HEVC. We emphasize that this process operates significantly faster than others, such as pruning.
For small-sized model, compression with CRF 5~$\sim$~7 yields approximately 0.1 bpp.

\subsection{Training settings}
We train models for 100K iterations, with each batch randomly sampled to include 0.1\% of the total coordinates per iteration as \cite{kim2022scalable}.
We use AdamW optimizer \cite{loshchilov2017decoupled} implemented with a learning rate of $10 \time 10^{-2}$ and weight decay of $10 \time 10^{-3}$.
We apply a cosine annealing scheduler with a minimum learning rate of $10 \time 10^{-5}$ and a diminishing duration of the entire iterations.
All experiments are conducted on a single NVIDIA A100 GPU.

\section{Implementation Details of Baselines}
To emulate real-world application scenarios, we conduct the comparison at 0.1 bpp, over a time span of 1 hour on Section~\ref{subsec:encoding_speed}. We compare a NVTM with NeRV \cite{chen2021nerv}, E-NeRV \cite{li2022nerv}, HNeRV \cite{chen2023hnerv} and NVP \cite{kim2022scalable}.
\vspace{0.2cm} \newline \textbf{NeRV.}
We use (48, 384) as the (C1, C2) parameters for the UVG dataset (600 frames) and (17, 17) for the MCL-JCV dataset (100 frames).
\vspace{0.2cm} \newline \textbf{E-NeRV.}
We use (166, 498) as the (C1, C2) parameters for the UVG dataset (600 frames) and (24, 48) for the MCL-JCV dataset (100 frames).
\vspace{0.2cm} \newline \textbf{HNeRV.}
We utilize strides of (1,5,3,2,2,2) and kernel sizes of (1,1,3,3,3,3) across six levels in decoder, and set the total parameters to 15M parameters for the UVG dataset (600 frames) and 2.5M for the MCL-JCV dataset (100 frames).
\vspace{0.2cm} \newline \textbf{NVP.}
We use the model configuration of NVP-S provided by the authors without any modifications.

\vspace{0.5cm} 
We also compare a NVTM with parametric encoding methods, such as Instant-NGP \cite{muller2022instant}, 3D ModSIREN and NVP \cite{kim2022scalable} (without compression) on Section~\ref{subsec:param_efficency},~\ref{subsec:downstream}. For a fair comparison of parameter efficiency, we adjust each model size to be comparable. All grids are implemented using \texttt{DenseGrid} in  tiny-cuda-nn \cite{tinycudann}. 
\vspace{0.2cm} \newline \textbf{Instant-NGP.}
This consists of a 3D grid with hash and a base network.
We configure this with a base resolution of (16$\times$16$\times$6), 15 levels, a scale of 1.4, and 4 features per level. The base network is composed of two layers with 64 neurons.
\vspace{0.2cm} \newline \textbf{3D ModSIREN.}
This consists of a 3D grid and a base network.
We configure this as a 3D grid configured with a base resolution of (16$\times$16$\times$6), 5 levels, a scale of 2.225, and 2 features per level. The base network is composed of three layers with 128 neurons.

\section{More Comparison on Downstream Tasks}
\vspace{-0.2cm}
In this section, we describe the evaluation settings for each downstream task and present extra results.
\textbf{More visualization results are in Section~\ref{sec:sup_vis}.}

\vspace{-0.5cm}
\subsection{Video reconstruction}
First, we present sequence-wise results of Table \ref{tb:mainresult_final}.

\begin{table}[H]
\centering
\vspace{-0.5cm}
\caption{\small Video reconstruction results on UVG (Dynamic) with Full HD resolution.} 
\begin{small}
    \begin{tabular}{c|cc|cc|cc|cc}
    \hline
    \multirow{2}{*}{Method} & \multicolumn{2}{c}{Bosphorus} \vline & \multicolumn{2}{c}{Jockey} \vline & \multicolumn{2}{c}{ReadySetGo} \vline & \multicolumn{2}{c}{YachtRide} \\
    & {\scriptsize PSNR$\uparrow$} & {\scriptsize LPIPS$\downarrow$}
    & {\scriptsize PSNR$\uparrow$} & {\scriptsize LPIPS$\downarrow$}
    & {\scriptsize PSNR$\uparrow$} & {\scriptsize LPIPS$\downarrow$}
    & {\scriptsize PSNR$\uparrow$} & {\scriptsize LPIPS$\downarrow$} \\ 
    \hline
    Instant-NGP & 40.98 & 0.064 & 36.63& 0.193 & 33.35 & 0.160& 37.37& 0.085  \\
    3D ModSIREN & 40.92& 	0.051& 	37.80& 	0.187	& 34.02& 	0.084& 	36.23& 	0.057 \\
    NVP & 41.47& 	0.055& 	40.1& 	0.188& 	36.48& 	0.062& 	37.94& 	0.053 \\ 
    % \rowcolor{LightCyan}
    NVTM & \textbf{43.41}& 	\textbf{0.032}&  	\textbf{40.18}& 	\textbf{0.174}& 	\textbf{39.71}& 	\textbf{0.036}& 	\textbf{38.88}& 	\textbf{0.044} \\
    \hline
\end{tabular}
\end{small}
\end{table}

\begin{table}[H]
\centering
\vspace{-1.5cm}
\caption{\small Video reconstruction results on MCL-JCV (Dynamic) with Full HD resolution.} 
\begin{small}
    \begin{tabular}{c|cc|cc|cc|cc|cc}
        \hline
        \multirow{2}{*}{Method} & \multicolumn{2}{c|}{VideoSRC04} & \multicolumn{2}{c|}{VideoSRC05} & \multicolumn{2}{c|}{VideoSRC11} & \multicolumn{2}{c|}{VideoSRC20} & \multicolumn{2}{c}{VideoSRC21} \\
        & {\scriptsize PSNR$\uparrow$} & {\scriptsize LPIPS$\downarrow$} 
        & {\scriptsize PSNR$\uparrow$} & {\scriptsize LPIPS$\downarrow$}
        & {\scriptsize PSNR$\uparrow$} & {\scriptsize LPIPS$\downarrow$}
        & {\scriptsize PSNR$\uparrow$} & {\scriptsize LPIPS$\downarrow$}
        & {\scriptsize PSNR$\uparrow$} & {\scriptsize LPIPS$\downarrow$} \\ 
        \hline
        Instant-NGP & 34.37 & 0.138 & 34.52 & 0.146 & 42.85 & 0.089 & 39.12 & 0.033 & 45.73 & 0.058 \\
        3D ModSIREN & 32.48 & 0.173 & 32.67 & 0.217 & 38.78 & 0.119 & 37.15 & 0.076 & 43.72 & 0.083 \\
        NVP & 33.89 & 0.146 & 34.42 & 0.142 & 41.70 & 0.087 & 41.71 & 0.032 & 45.88 & 0.058 \\
        NVTM & \textbf{34.57} & \textbf{0.134} & \textbf{36.26} & \textbf{0.127} & \textbf{44.07} & \textbf{0.068} & \textbf{45.17} & \textbf{0.020} & \textbf{46.87} & \textbf{0.054} \\
        \hline
    \end{tabular}
\end{small}
\end{table}

We also evaluate NVTM on different resolutions, resulting superior performance on different aspect ratio (1.45dB/0.014 improvements on 1920~$\times$~960) and smaller motion scales (4.62dB/0.037 improvements on 960~$\times$~540). Each dataset is generated through center-cropping and resizing from the original HD video respectively. 
\begin{table}[H]
\vspace{-0.8cm}
\caption{\small Video reconstruction results on UVG (Dynamic) with resolution 1920~$\times$~960.} 
\centering
\begin{small}
\begin{tabular}{c|cc|cc|cc|cc}
    \hline
    \multirow{2}{*}{Method}
     & \multicolumn{2}{c}{Bosphorus} \vline & \multicolumn{2}{c}{Jockey}\vline & \multicolumn{2}{c}{ReadySetGo} \vline & \multicolumn{2}{c}{YachtRide} \\
    & {\scriptsize PSNR$\uparrow$} & {\scriptsize LPIPS$\downarrow$}
    & {\scriptsize PSNR$\uparrow$} & {\scriptsize LPIPS$\downarrow$}
    & {\scriptsize PSNR$\uparrow$} & {\scriptsize LPIPS$\downarrow$}
    & {\scriptsize PSNR$\uparrow$} & {\scriptsize LPIPS$\downarrow$} \\ 
    \hline
    Instant-NGP & 40.93 & 0.067 & 36.61 & 0.187 & 33.62 & 0.156 & 38.69 & 0.073 \\
    3D ModSIREN & 41.00 & 0.048	& 37.96	& 0.182 & 34.45 & 0.079 & 37.63 & 0.049 \\ 
    NVP & 41.64 & 0.052	& 40.20	& 0.179 & 37.22 & 0.052 & 39.26 & 0.048 \\ 
    % \rowcolor{LightCyan}
    % NVTM& 122M$^\star$ & \textbf{43.17} & \textbf{0.030} & \textbf{40.21} & \textbf{0.166} & \textbf{40.13} & \textbf{0.032} & \textbf{39.99} & \textbf{0.043} \\ \hline
    NVTM& \textbf{43.49} & \textbf{0.029} & \textbf{40.26} & \textbf{0.174} & \textbf{40.14} & \textbf{0.033} & \textbf{40.22} & \textbf{0.041} \\ 
    \hline
\end{tabular}%
\end{small}
\end{table}

\begin{table}[H]
\centering
\vspace{-1.5cm}
\caption{\small Video reconstruction results on UVG (Dynamic) with resolution 960~$\times$~540.} 
\label{tb:recon_uvg_960x540}
\begin{small}
\begin{tabular}{c|cc|cc|cc|cc}
    \hline
    \multirow{2}{*}{Method} & \multicolumn{2}{c}{Bosphorus} \vline & \multicolumn{2}{c}{Jockey}\vline & \multicolumn{2}{c}{ReadySetGo} \vline & \multicolumn{2}{c}{YachtRide} \\
    & {\scriptsize PSNR$\uparrow$} & {\scriptsize LPIPS$\downarrow$}
    & {\scriptsize PSNR$\uparrow$} & {\scriptsize LPIPS$\downarrow$}
    & {\scriptsize PSNR$\uparrow$} & {\scriptsize LPIPS$\downarrow$}
    & {\scriptsize PSNR$\uparrow$} & {\scriptsize LPIPS$\downarrow$} \\ 
    \hline
    Instant-NGP & 40.01 & 0.062 & 34.17 & 0.161 & 29.85 & 0.145 & 35.51 & 0.086 \\
    3D ModSIREN & 39.34 & 0.049	& 35.05	& 0.110 & 29.27 & 0.112 & 33.69 & 0.061 \\ 
    NVP & 37.98 & 0.063	& 39.13	& 0.069 & 32.21 & 0.059 & 33.85 & 0.062 \\ 
    % \rowcolor{LightCyan}
    NVTM & \textbf{43.62} & \textbf{0.015} & \textbf{41.40} & \textbf{0.048} & \textbf{38.39} & \textbf{0.014} & \textbf{38.23} & \textbf{0.028} \\ 
    \hline
\end{tabular}%
\end{small}
\end{table}

\subsection{Video super resolution}
To evaluate quantitatively, we perform inference on models which are trained on (600$\times$1080$\times$1920), at an input coordinate of (600$\times$2159~$\times$3839). We then evaluate frames of the generated resolution by using 4K resolution frames. 
% Table~\ref{sup:tb:super_resolution} is a sequence-wise super resolution result of Table~\ref{tb:sr_fruc}.

\begin{table}[H]
\vspace{-1.0cm}
\caption{\small Comparison of PSNR on video super resolution result ($\times$2) in UVG.} 
\setlength{\tabcolsep}{6pt}
\label{sup:tb:super_resolution}
\centering
\begin{small}
    \begin{tabular}{c|cccc}
        \hline
        Method & Bosphorus & Jockey & ReadySetGo & YachtRide \\
        \hline
        3D ModSIREN & 38.76 & 34.10 & 31.87 & 34.13  \\
        NVP & 34.68 & 33.06 & 28.92 & 30.80 \\
        NVTM & \textbf{39.36} & \textbf{34.82} & \textbf{33.99} & \textbf{35.09}\\ \hline
    \end{tabular}%
\end{small}
\end{table}

\vspace{-1.0cm}
\subsection{Video frame interpolation}
We perform inference on models which are trained on (300$\times$1080$\times$1920) consisting of only odd frames, at an input coordinate of (599$\times$1080$\times$1920). 
% Table~\ref{sup:tb:frame_interpolation} is a sequence-wise frame interpolation result of Table~\ref{tb:sr_fruc}.
We also compare frame interpolation performance with NeRV-like Method.
Since DNeRV \cite{zhao2023dnerv} reported with a different resolution and its code is unavailable, we only compared with HNeRV \cite{chen2023hnerv}. NVTM has large margins compared to HNeRV, except Jockey sequence.

\begin{table}[H]
\vspace{-1.0cm}
\caption{\small Comparison of PSNR on frame interpolation result ($\times$2) in UVG.}
\setlength{\tabcolsep}{6pt}
\label{sup:tb:frame_interpolation}
\centering
\begin{small}
    \begin{tabular}{c|cccc}
        \hline Method & Bosphorus & Jockey & ReadySetGo & YachtRide \\ \hline
        HNeRV & 28.61 & \textbf{27.97} & 21.21 & 23.93 \\ 
        3D ModSIREN & 34.97	& 21.32	&	19.27	& 26.06 \\	
        NVP & 31.55 & 	20.66  & 	18.46 & 24.86  \\
        NVTM & \textbf{37.07} &  27.57 & \textbf{28.85} & \textbf{28.47}  \\ \hline       
    \end{tabular}%
\end{small}
\end{table}

\subsection{Video inpainting}
We evaluate video inpainting on blackswan, camel, cows, drift-chicane, soapbox, tennis sequences on DAVIS2017 \cite{DAVIS2017} dataset. 
Evaluation results of more sequences are on Figure~\ref{fig:inpainting}.
As \cite{chen2023hnerv} conducted, we generate 10 random box masks, each with a 100 pixel width for every time step $t$. During training time, the pixels within these masks are excluded from the backpropagation. During inference time, all pixels are predicted, thus testing the model's ability to implicitly infer masked pixels without explicit training on them. 

\begin{figure}[H]
\renewcommand{\mywidth}{1.00}
    \vspace{-0.5cm}
    \centering
    \includegraphics[width=\mywidth\linewidth]{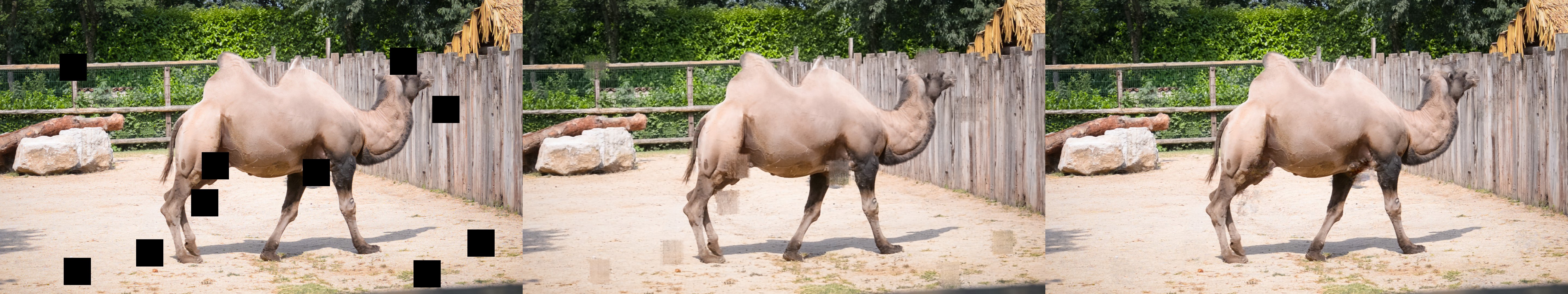}
    \includegraphics[width=\mywidth\linewidth]{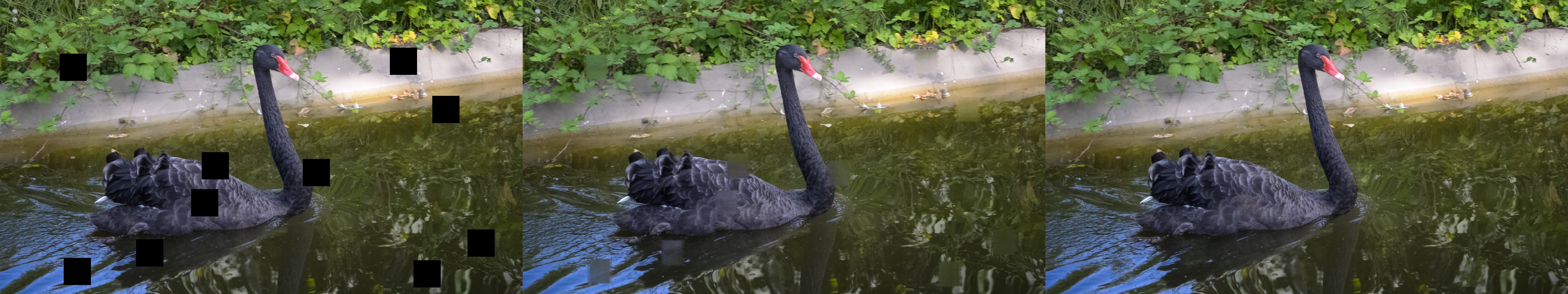}
    \includegraphics[width=\mywidth\linewidth]{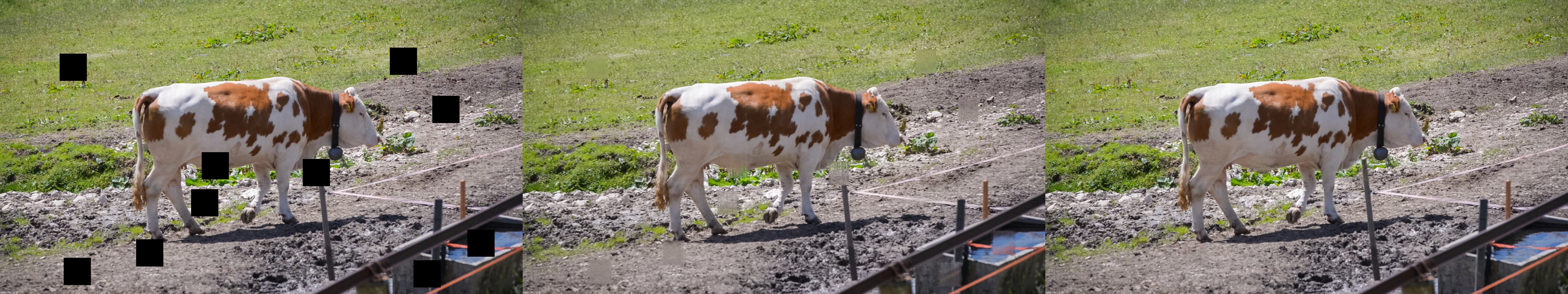}
    \includegraphics[width=\mywidth\linewidth]{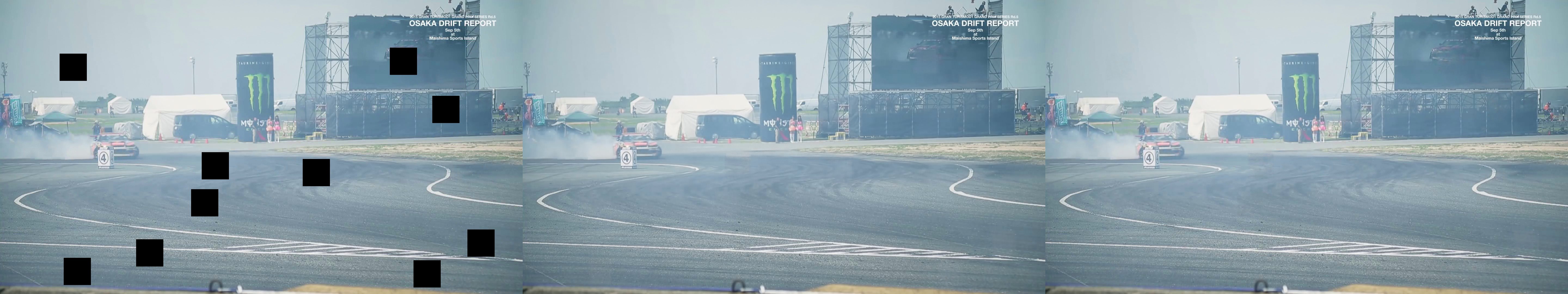}
    \includegraphics[width=\mywidth\linewidth]{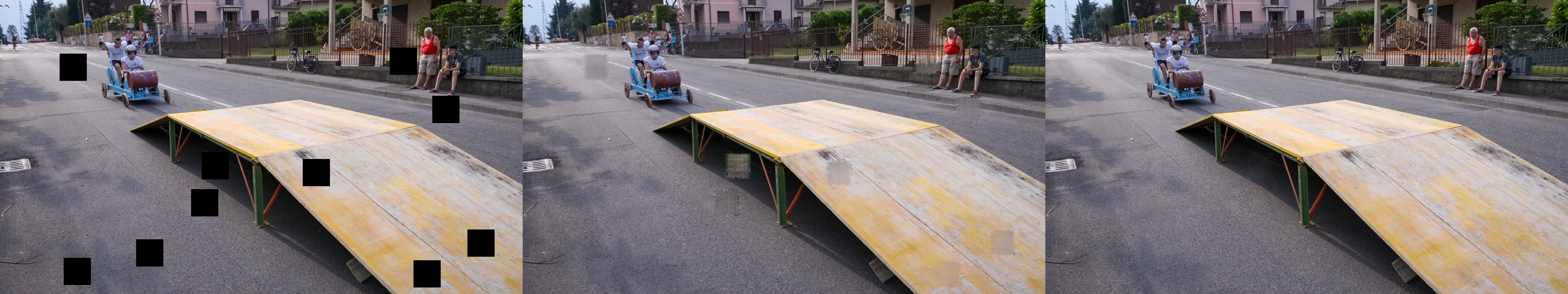}
    \includegraphics[width=\mywidth\linewidth]{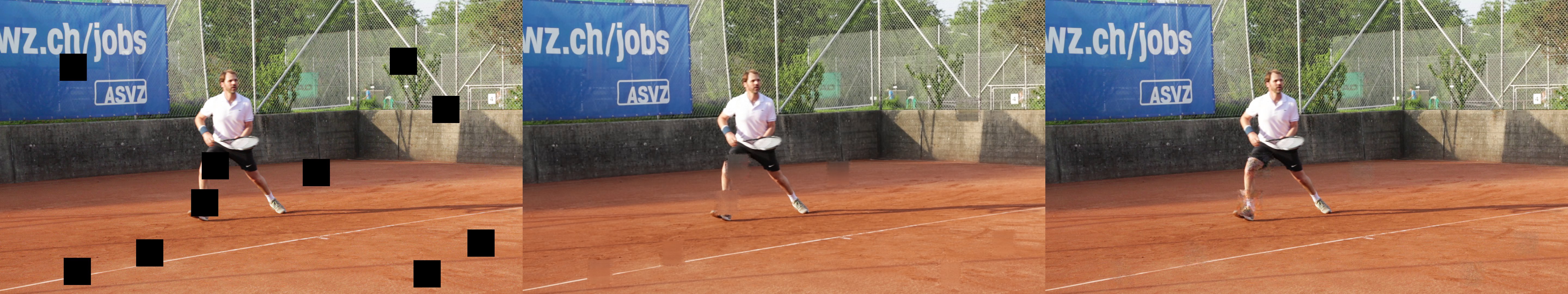}
    \begin{minipage}{0.2833\linewidth}
        \centering
         Masked image
    \end{minipage}%
    \begin{minipage}{0.2833\linewidth}
        \centering
        NVP
    \end{minipage}%
    \begin{minipage}{0.2833\linewidth}
        \centering
        NVTM
    \end{minipage}
    \caption{\small Visualization results of video inpainting. Each are the first frame of Camel, Blackswan, Cows, Drift-Chicane, Soapbox and Tennis sequence on DAVIS2017.}
    \label{fig:inpainting}
    \vspace{-0.5cm}
\end{figure}

\subsection{Video compression}
We evaluate video compression by comparing PSNR against the Bits-Per-Pixel (bpp) for various models.
To compare with our model, we evaluate standard codecs such as H.264 \cite{wiegand2003overview} and HEVC \cite{sullivan2012overview} using FFmpeg \cite{tomar2006converting} as following commands. 
\begin{footnotesize}
\begin{verbatim}
# H.264
ffmpeg -y -i ${seq}/im%04d.png -c:v h264 -preset medium -crf ${crf} \
-bf 0 -framerate 25 ${savefile}.mp4
 
# HEVC
ffmpeg -n -i ${seq}/im%04d.png -c:v hevc -preset medium -crf ${crf} \
-x265-params "bframes=0" -strict experimental -framerate 25 ${savefile}.mp4
\end{verbatim}
\end{footnotesize}

Also we evaluate INR-based video compression methods, such as NeRV \cite{chen2021nerv}, E-NeRV \cite{li2022nerv} and NVP \cite{kim2022scalable} according to the author's configurations. 
Since E-NeRV does not provide configurations for HD resolution, we manually configured this network by adjusting its internal feature sizes to match the parameter sizes of NeRV.
We experiment on five sizes of (C1, C2) as (166, 498), (219, 657), (208, 832), (216, 1080), (222, 1332).

\begin{figure}[H]
    \renewcommand{\mywidth}{0.49}
    \centering
    \setlength{\tabcolsep}{1pt} % Default value: 6pt
    \begin{tabular}{cc}
    \includegraphics[width=\mywidth\linewidth]{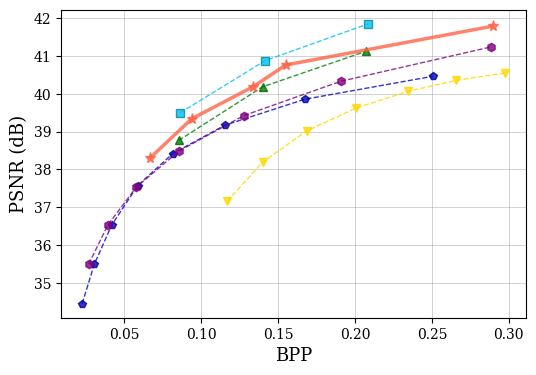} 
    & \includegraphics[width=\mywidth\linewidth]{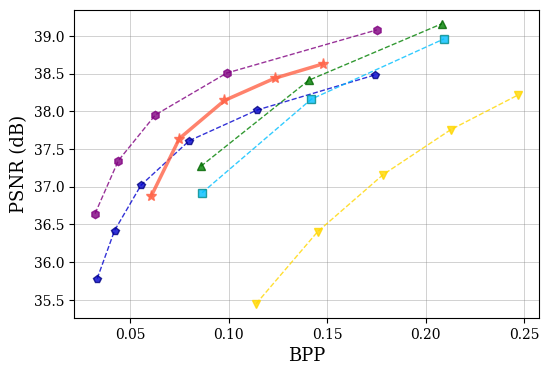} \\
    Bosphorus & ReadySetGo \\
    \includegraphics[width=\mywidth\linewidth]{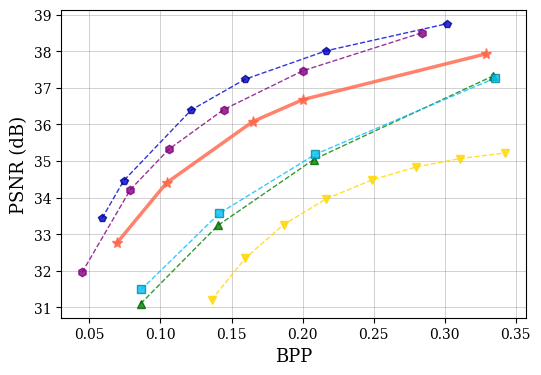} 
    & \includegraphics[width=\mywidth\linewidth]{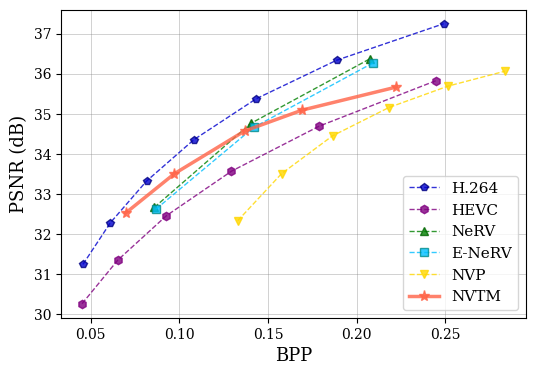} \\
    Bosphorus & ReadySetGo \\
    \end{tabular}%
    \caption{\small Sequence-wise video compression results on UVG (Dynamic).}
    \label{figure:compression_seq}
\end{figure}

\newpage
We also compare ours with D-NeRV \cite{he2023towards}.
We confirmed that ours is still strong in encoding speed, however, when each model is trained longer to converge, ours did not achieve the state-of-the-art performance.
\begin{table}[H]
\vspace{-0.8cm}
    \centering
    \setlength{\tabcolsep}{5pt}
    \begin{small}
    \caption{Encoding speed comparison with D-NeRV in compression performance.} 
    \begin{tabular}{c|ccccc}
        \hline
        Models & 1min. & 5min. & 10min. & 20min. & 60min. \\
        \hline
        NVTM & \textbf{26.52} & \textbf{29.20} & \textbf{29.97} & 30.49 & 31.85 \\ 
        D-NeRV (S)  & 19.12 & 26.12 & 28.64 & \textbf{31.36} & \textbf{34.48} \\ 
        \hline
    \end{tabular}
    \end{small}
\end{table}

\section{More Analysis}

\subsection{Parameter distribution}
Our framework uses two major types of parameters: \texttt{Latent grid (parametric encoding)} and \texttt{base network (MLP)}. We investigate how the proportion of these two types of parameters affects video reconstruction quality and encoding speed. As shown in Figure~\ref{fig:distribution}, decreasing the proportion of \texttt{base network} parameters from A to F results in slower encoding speeds but improved video reconstruction performance. Therefore, the choice of parameter setting depends on the trade-off between training speed and performance.
\begin{figure}[H]
    \centering
    \includegraphics[width=0.60\linewidth]{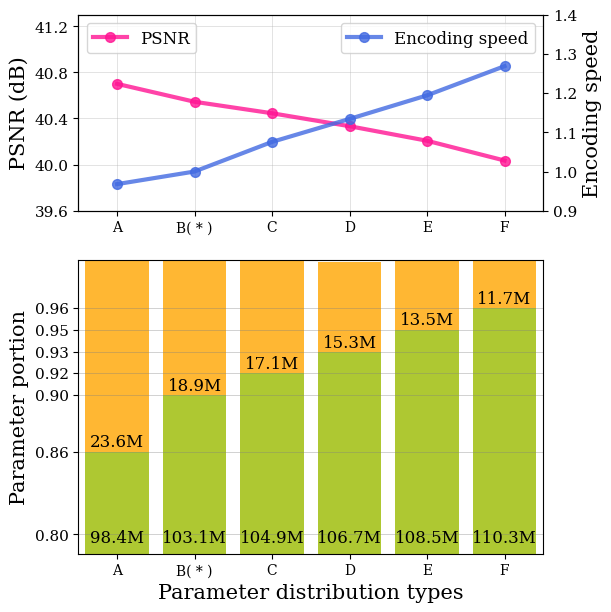}
    \vspace{-0.5cm}
    \caption{\small Trade-off on performance and training speed. A to F represents different types of parameter distribution, and \textbf{\small B(*)} denotes the type we experimented. On upper figure, \textcolor{RubineRed}{\textbf{red line}} represents PSNR (left y-axis), and the \textcolor{RoyalBlue}{\textbf{blue line}} represents encoding speed (right y-axis). The encoding speed is presented as a relative value based on our default configuration set to 1.0. On below figure, each bar graph represents the distribution of parameters: \textcolor{Dandelion}{\textbf{yellow region}} represents parameters for \texttt{base network} and \textcolor{LimeGreen}{\textbf{green region}} for \texttt{Latent grid}. 
    }
    \label{fig:distribution}
\end{figure}

\subsection{Alignment flow for 2D Aligned Coordinate}
On Figure \ref{fig:flow}, the alignment flow network is trained to be similar as optical flow, while having superior mapping results than optical flow (i.e., has less warping error on (c) than (d)). This indicates that the module effectively captures temporal coherent features.
\begin{figure}[H]   % t: top, b: bottom, h: here, p: page of floats
\renewcommand{\mywidth}{0.33}
\centering
\begin{small}
    \begin{tabular}{cc}
        \includegraphics[width=\mywidth\linewidth]{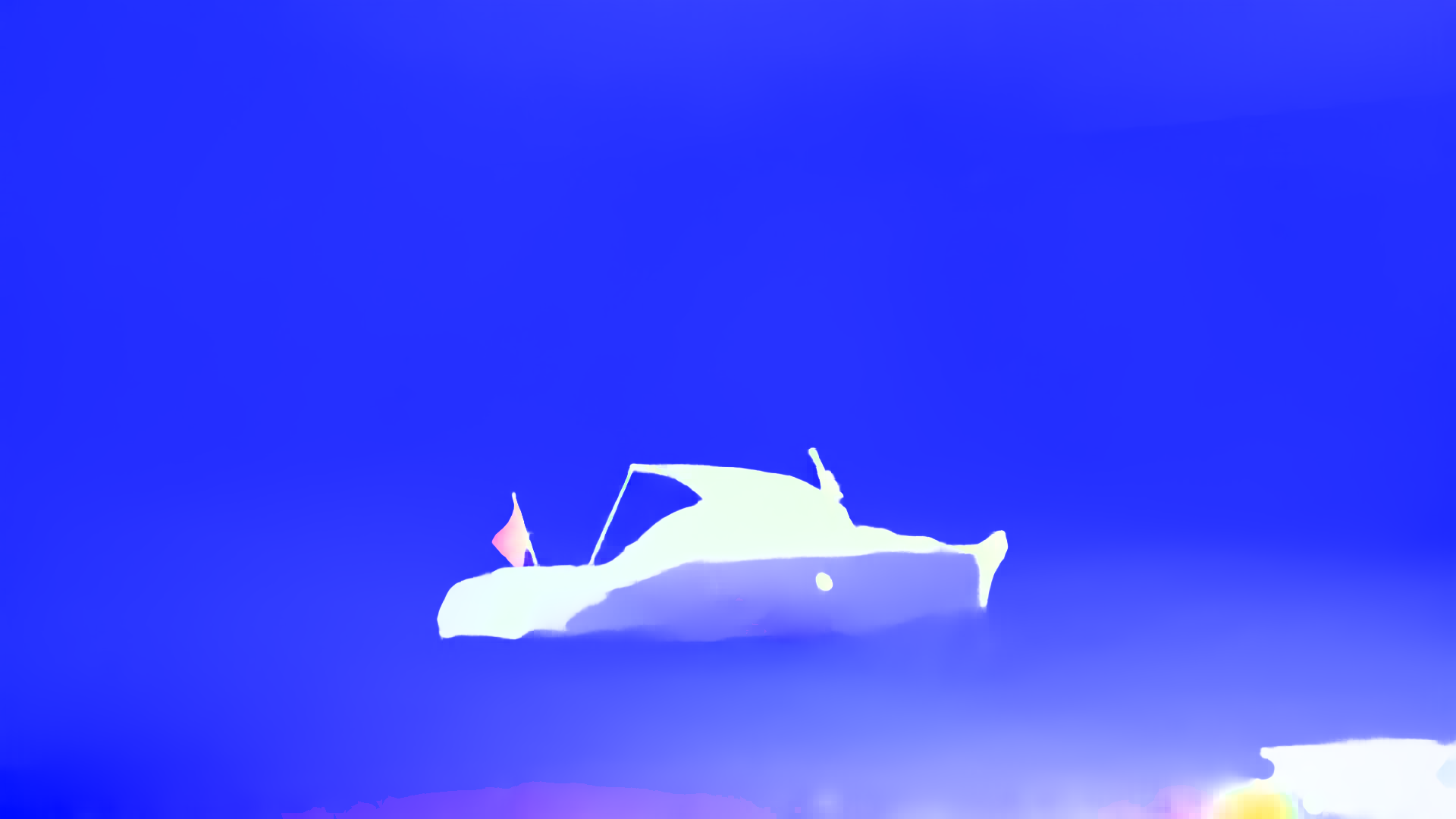} 
        & \includegraphics[width=\mywidth\linewidth]{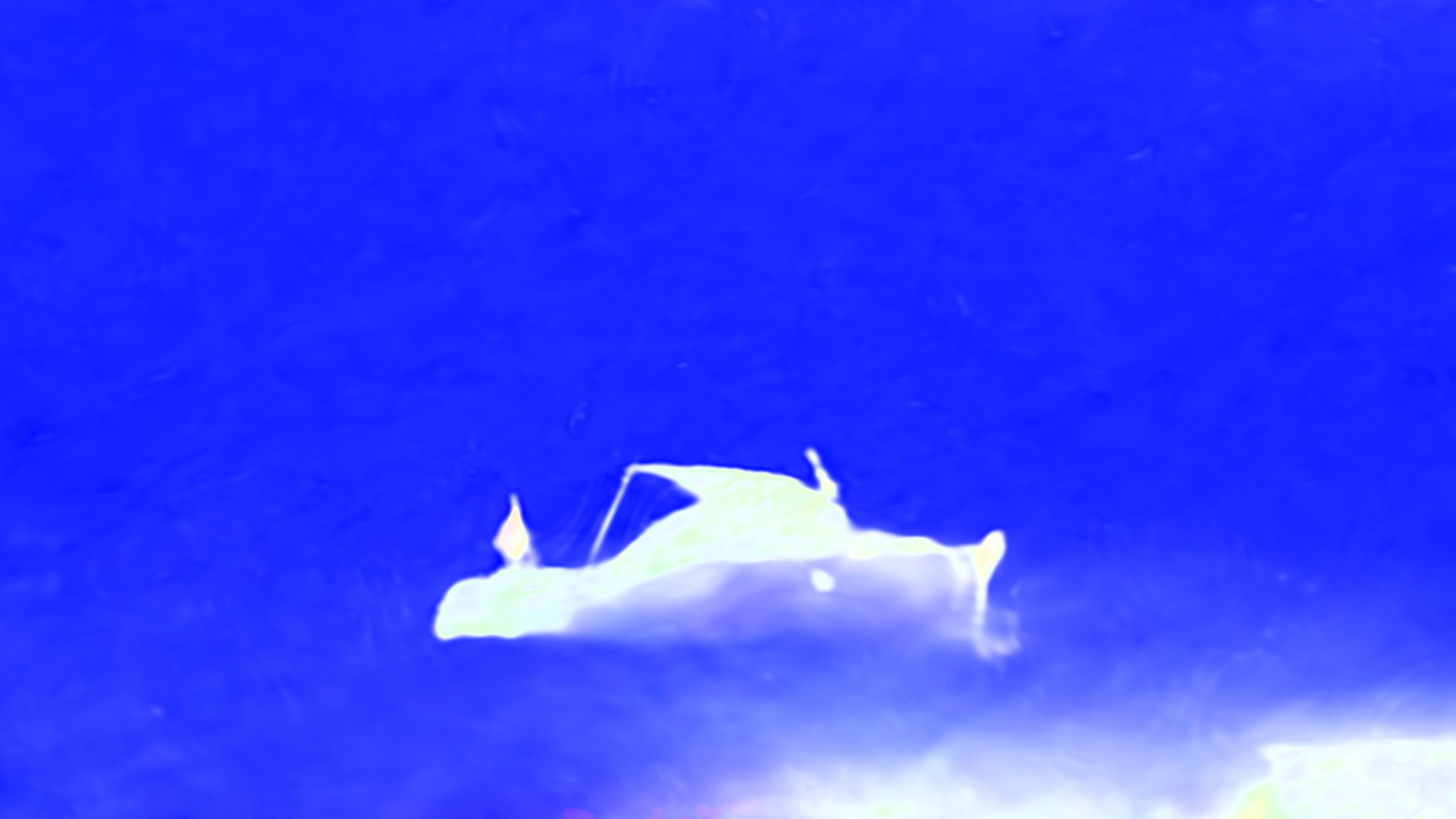}  \\
        \includegraphics[width=\mywidth\linewidth]{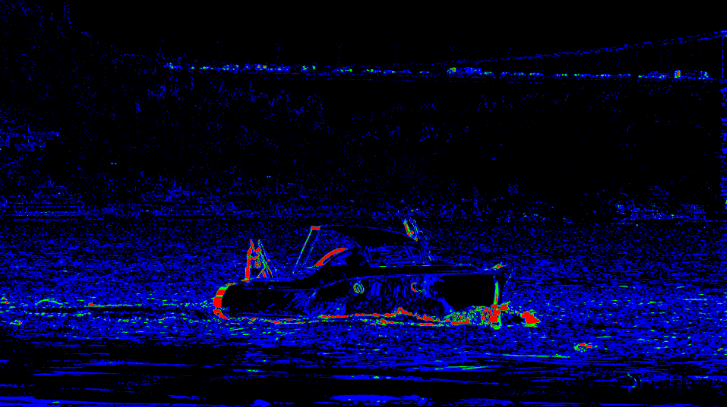}
        & \includegraphics[width=\mywidth\linewidth]{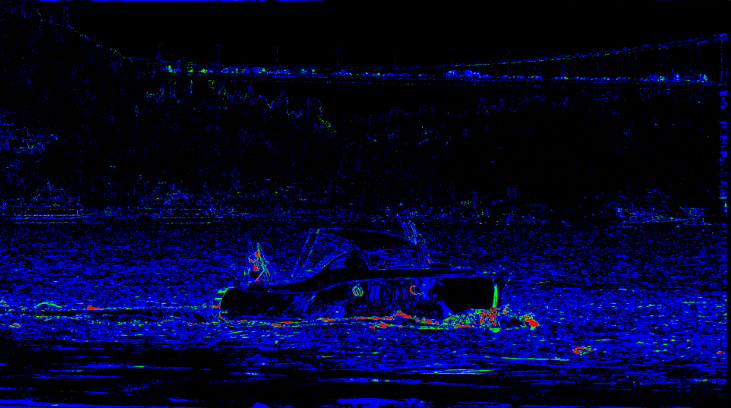} \\ 
        {(a) Optical flow} & {(b) Alignment flow} \\
        % {(c) Warping error with (a)} & {(d) Warping error with (b)} \\
    \end{tabular} 
\end{small}
\caption{\small Comparison on optical flow and our alignment flow. 
Our network generates the alignment flow across different times within the GOP (Eq.~\ref{eq:flow_hyper}), then maps the 2D aligned coordinates (Eq.~\ref{eq:dimension_reduction_naive}). We compare the optical flow, which we used for guidance, with our flow output on Bosphorus sequence.
(a) is visualization of optical flow and RGB-level differences between the original frame and warped results using the flow.
(b) is visualization of our alignment flow and RGB-level differences.}
\label{fig:flow}
\vspace{-0.5cm}
\end{figure}

\subsection{Extremely low-motion video}
As described earlier, our proposed framework is based on processing the corresponding parts in a sequence of video frames, and therefore it is assumed to be effective in videos with a certain degree of motion. Therefore, the UVG sequences Beauty and HoneyBee were not targeted in the previous sequence selection process, as they were judged to have relatively low dynamics.
In addition, we further compared the performance of the models on those sequences.
We evaluate our framework on extremely low-motion sequences in UVG, Beauty and HoneyBee, where it was difficult to demonstrate the strengths of our framework. Table~\ref{tab:motionless} shows that although the performance difference was not significant, NVP performed best on Beauty and NVTM performed best on HoneyBee.

\begin{table}[H]
\centering
\setlength{\tabcolsep}{3pt} % Default value: 6pt
\vspace{-0.5cm}
\caption{\small Video reconstruction performance on extremely low-motion video sequences.} 
\label{tab:motionless}
\begin{small}
    \begin{tabular}{c|cc|cc}
    \hline
    \multirow{2}{*}{Method} & \multicolumn{2}{c|}{Beauty} & \multicolumn{2}{c}{HoneyBee} \\ 
    & {\scriptsize PSNR$\uparrow$} & {\scriptsize LPIPS$\downarrow$}
    & {\scriptsize PSNR$\uparrow$} & {\scriptsize LPIPS$\downarrow$} \\ 
    \hline
    Instant-NGP & 35.20 & 0.350 & 38.24 & 0.112  \\ 
    3D ModSIREN & 35.71 & 0.266 & 38.43 & 0.123  \\ 
    NVP & \textbf{36.24} & \textbf{0.273} & 40.25 & 0.110  \\ 
    NVTM & 35.63 & 0.330 & \textbf{40.43} & \textbf{0.108}  \\
    \hline
    \end{tabular}
\end{small}
\end{table}

\subsection{Decoding speed}
We compare the decoding time with the baselines, as shown in Table~\ref{table:decodingtime}. The decoding speed of NVTM is slightly slower than NVP \cite{kim2022scalable}. Although these pixel-wise INR (NVTM, NVP) are quite slower than frame-wise INR \cite{chen2023hnerv}, they can be further optimized by utilizing C++/CUDA frameworks which enable efficient parallel-computing for the inference of grid parameters as suggested in \cite{tinycudann}.

\begin{table}[H]
\centering
\vspace{-0.5cm}
\caption{\small Decoding speed on a single A100 GPU.} 
\label{table:decodingtime}
\setlength{\tabcolsep}{6pt}
\begin{small}
\begin{tabular}{c|c|c|c}
    \hline
    Model  & Output type & Params. & FPS \\ 
    \hline
    HNeRV & Frame & 122M & 34.2 \\ 
    NVP & Pixel & 136M & 7.76\\
    NVTM & Pxiel & 122M & 6.55\\ 
    \hline
\end{tabular}%
\end{small}
\end{table}

\subsection{Random seed}
We assessed our framework's performance stability across various random seeds. Experiments employed five seeds: 88, 151, 911, 999, and 1004. Figure \ref{fig:randomseed} demonstrates that NVTM has more consistent performance across varying seeds compared to NVP. 

\begin{figure}[H]
\centering
\includegraphics[width=0.5\linewidth]{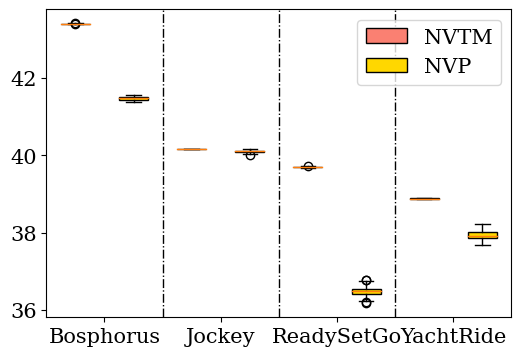}
\caption{\small PSNR variation across random seeds on video reconstruction.} 
\label{fig:randomseed}
\end{figure}

\clearpage
\section{More Visualization Results}
\label{sec:sup_vis}

\subsection{Video reconstruction}
\begin{figure}[H]
\renewcommand{\mywidth}{1.00}
\setlength{\tabcolsep}{1pt} % Default value: 6pt
\centering
    \begin{tabular}{cccc}
        \hline
        \multicolumn{4}{c}{\includegraphics[width=\mywidth\linewidth]{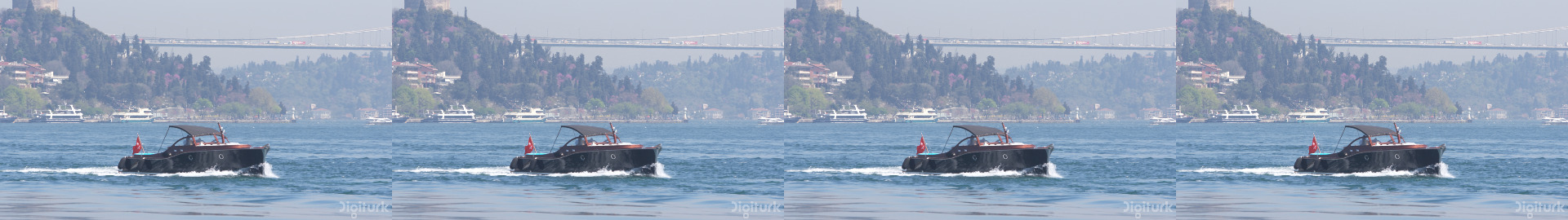}} \\
        \multicolumn{4}{c}{\includegraphics[width=\mywidth\linewidth]{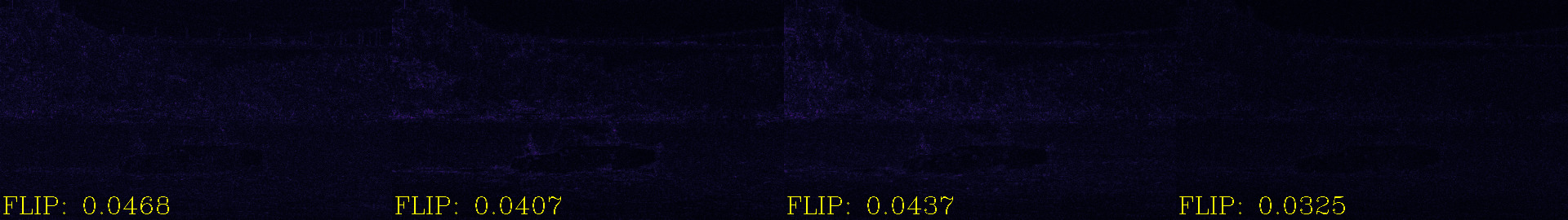}} \\
        \multicolumn{4}{c}{\includegraphics[width=\mywidth\linewidth]{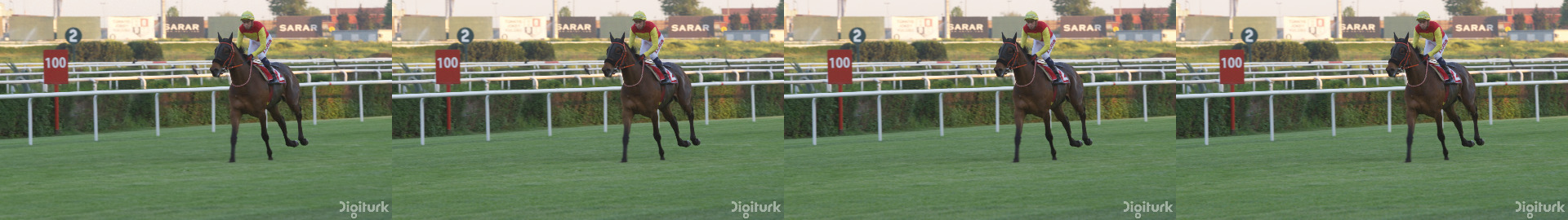}} \\
        \multicolumn{4}{c}{\includegraphics[width=\mywidth\linewidth]{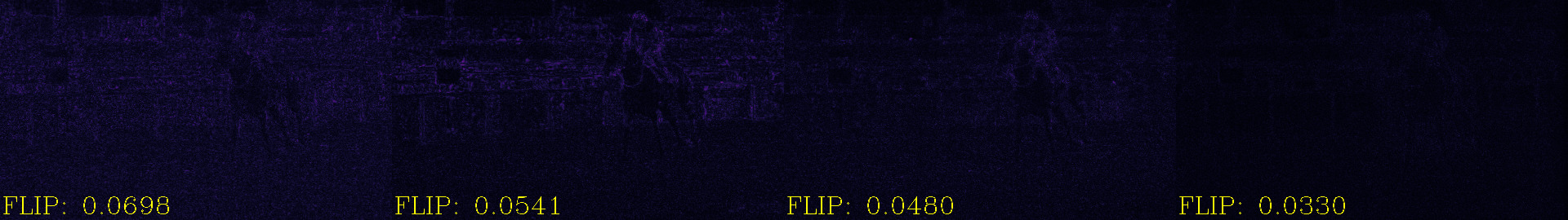}} \\
        \multicolumn{4}{c}{\includegraphics[width=\mywidth\linewidth]{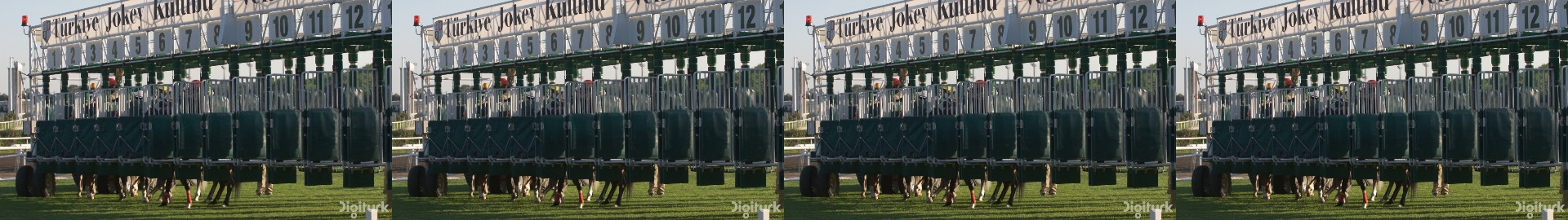}} \\
        \multicolumn{4}{c}{\includegraphics[width=\mywidth\linewidth]{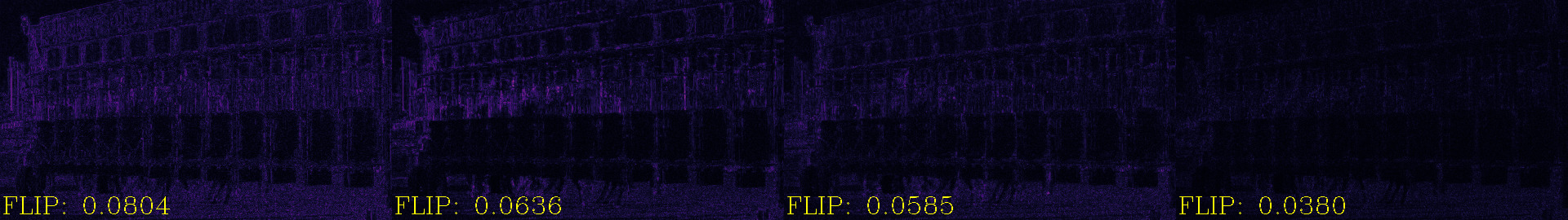}} \\
        \multicolumn{4}{c}{\includegraphics[width=\mywidth\linewidth]{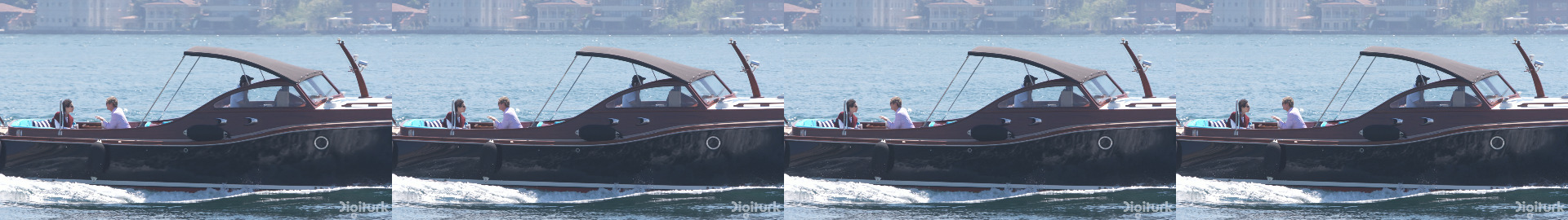}} \\
        \multicolumn{4}{c}{\includegraphics[width=\mywidth\linewidth]{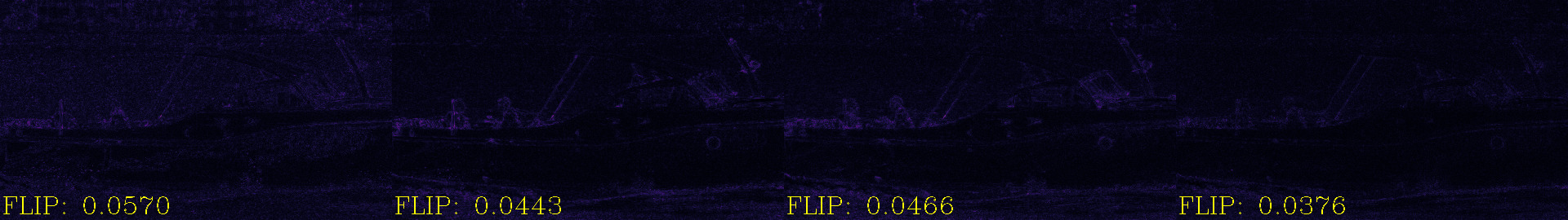}} \\
    \end{tabular}% 
    \\
     \begin{minipage}{0.2375\linewidth}
            \centering
             Instant-NGP
        \end{minipage}%
        \begin{minipage}{0.2375\linewidth}
            \centering
            3D ModSIREN
        \end{minipage}%
        \begin{minipage}{0.2375\linewidth}
            \centering
            NVP
        \end{minipage}%
        \begin{minipage}{0.2375\linewidth}
            \centering
            NVTM
        \end{minipage}
\caption{\small Video reconstruction result of first frame on UVG sequences (Bosphorus, Jockey, ReadySetGo and YachtRide from top to bottom). For each sequence, the first row indicates the decoded results, and the second row stands out the FLIP results, while darker colors indicating better performance.} 
\end{figure}

\begin{figure}[H]
\setlength{\tabcolsep}{1pt} % Default value: 6pt
\renewcommand{\mywidth}{1.00}
\centering
\begin{tabular}{cccc}
    \hline
    \multicolumn{4}{c}{\includegraphics[width=\mywidth\linewidth]{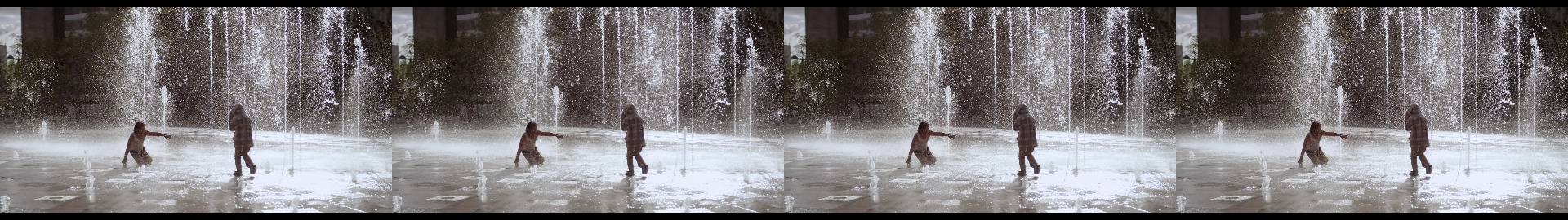}} \\
    \multicolumn{4}{c}{\includegraphics[width=\mywidth\linewidth]{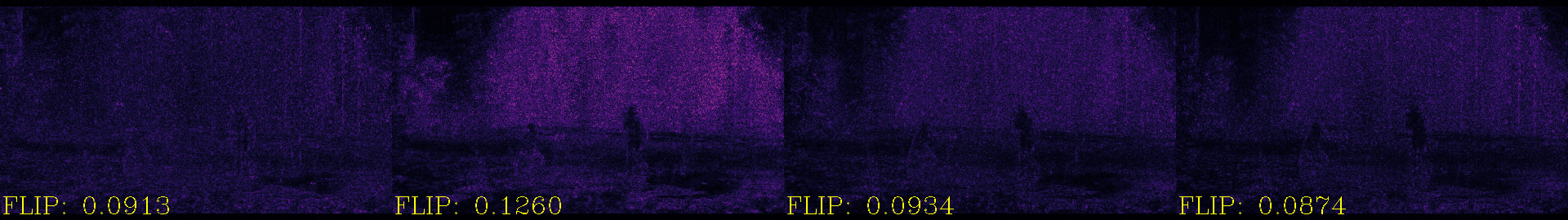}} \\
    \multicolumn{4}{c}{\includegraphics[width=\mywidth\linewidth]{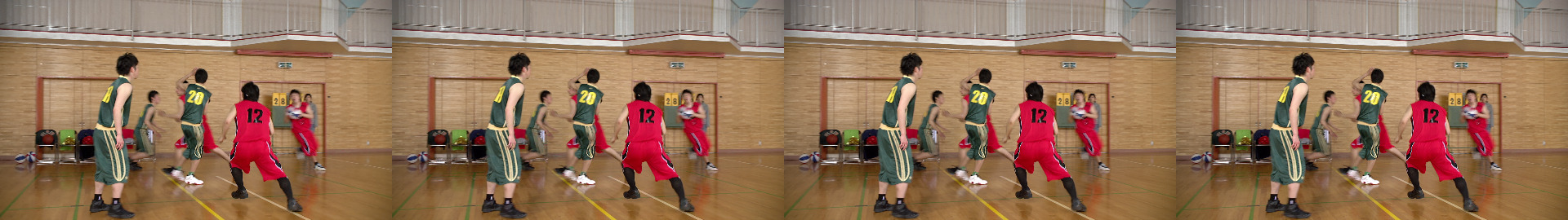}} \\
    \multicolumn{4}{c}{\includegraphics[width=\mywidth\linewidth]{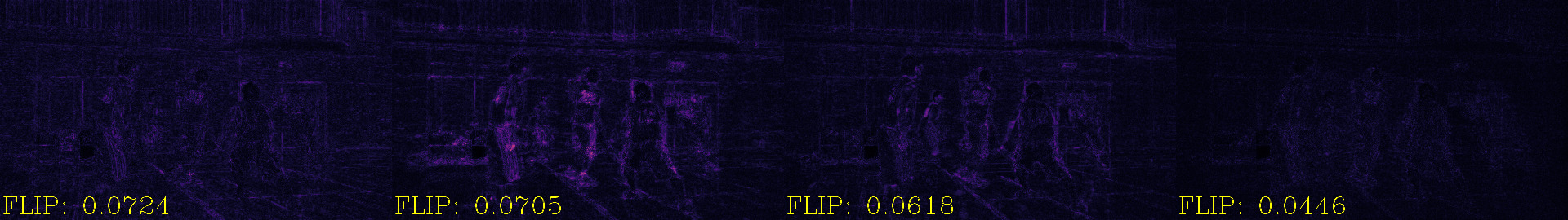}} \\
    \multicolumn{4}{c}{\includegraphics[width=\mywidth\linewidth]{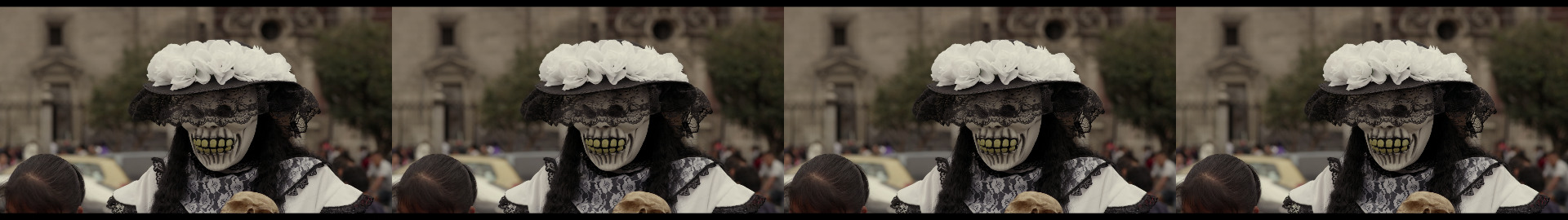}} \\
    \multicolumn{4}{c}{\includegraphics[width=\mywidth\linewidth]{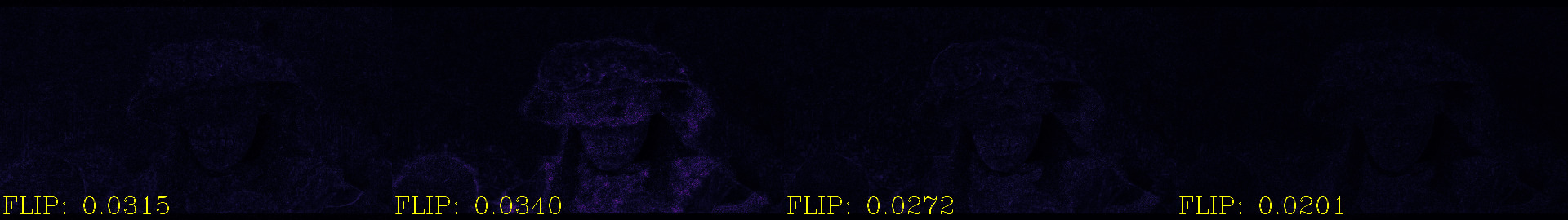}} \\ 
    \multicolumn{4}{c}{\includegraphics[width=\mywidth\linewidth]{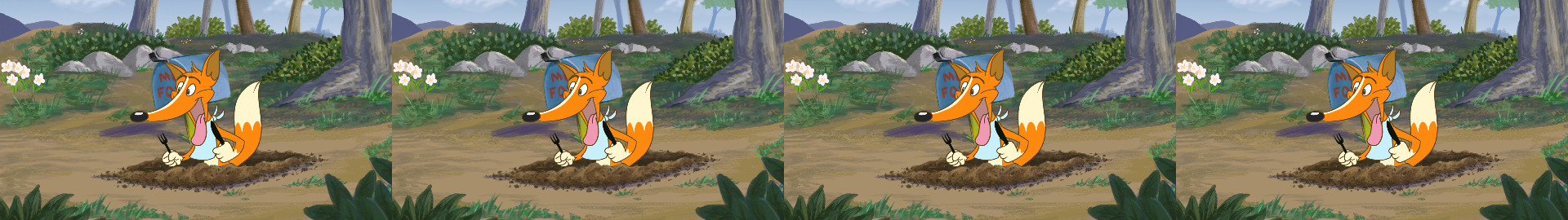}} \\
    \multicolumn{4}{c}{\includegraphics[width=\mywidth\linewidth]{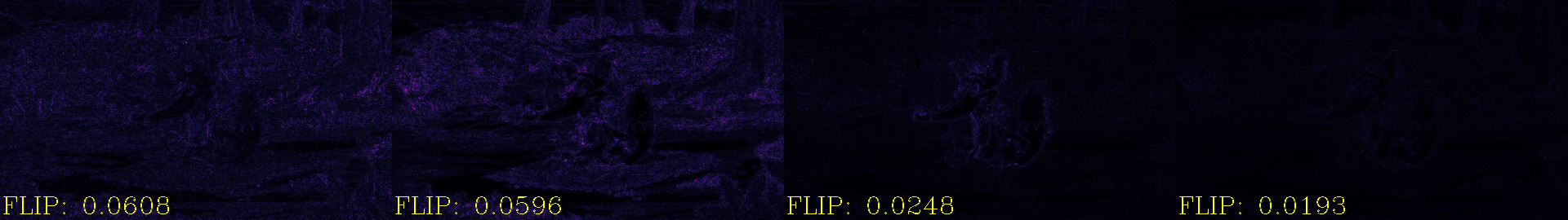}} \\
    \multicolumn{4}{c}{\includegraphics[width=\mywidth\linewidth]{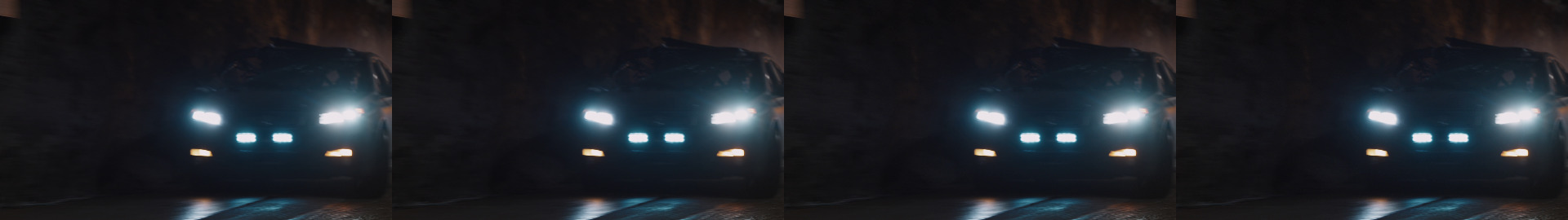}} \\
    \multicolumn{4}{c}{\includegraphics[width=\mywidth\linewidth]{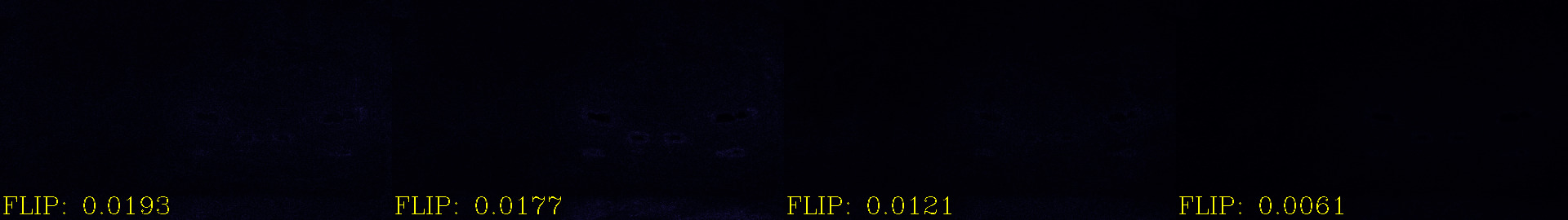}} \\
\end{tabular}%
\\
 \begin{minipage}{0.21\linewidth}
        \centering
         Instant-NGP
    \end{minipage}%
    \begin{minipage}{0.21\linewidth}
        \centering
        3D ModSIREN
    \end{minipage}%
    \begin{minipage}{0.21\linewidth}
        \centering
        NVP
    \end{minipage}%
    \begin{minipage}{0.21\linewidth}
        \centering
        NVTM
    \end{minipage}
\caption{\small Video reconstruction result of first frame on MCL-JCV sequences (04, 05, 11, 20, 21 from top to bottom). For each sequence, the first row indicates the decoded results, and the second row stands out the FLIP results, while darker colors indicating better performance.} 
\end{figure}

\subsection{Video super resolution}
\begin{figure}[H]
\centering
\renewcommand{\mywidth}{0.78}
\setlength{\tabcolsep}{1pt} % Default value: 6pt
\begin{tabular}{ccc}
    \multicolumn{3}{c}{\includegraphics[trim=480 0 0 0,clip,width=\mywidth\linewidth]{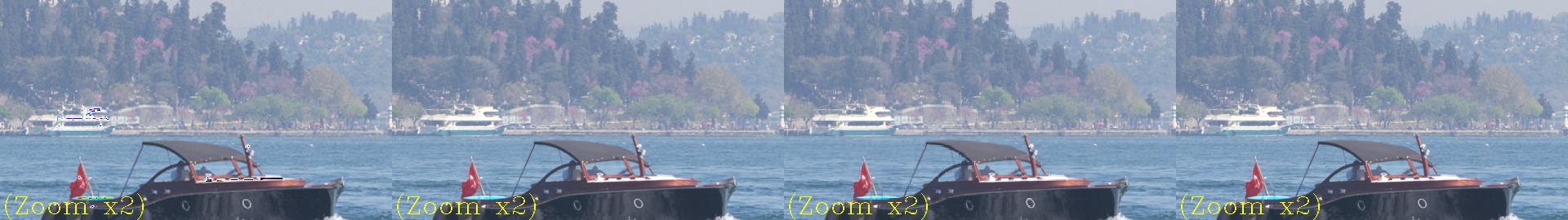}} \\
    \multicolumn{3}{c}{\includegraphics[trim=480 0 0 0,clip,width=\mywidth\linewidth]{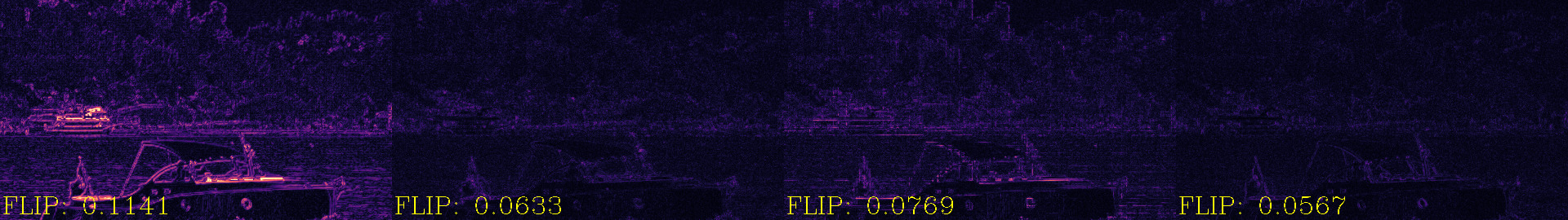}} \\
    \multicolumn{3}{c}{\includegraphics[trim=480 0 0 0,clip,width=\mywidth\linewidth]{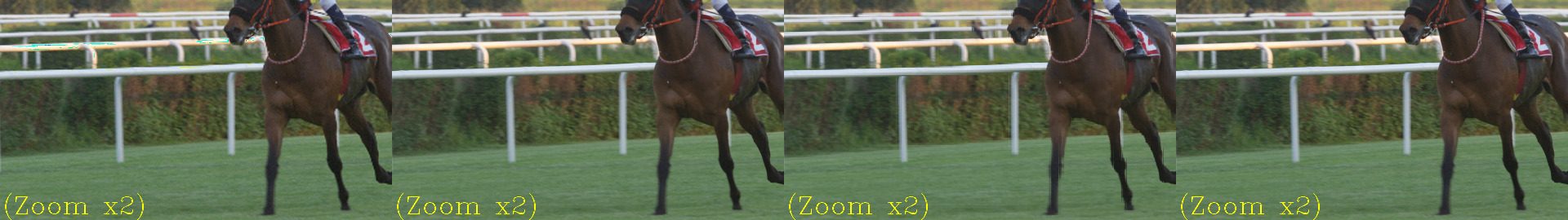}} \\
    \multicolumn{3}{c}{\includegraphics[trim=480 0 0 0,clip,width=\mywidth\linewidth]{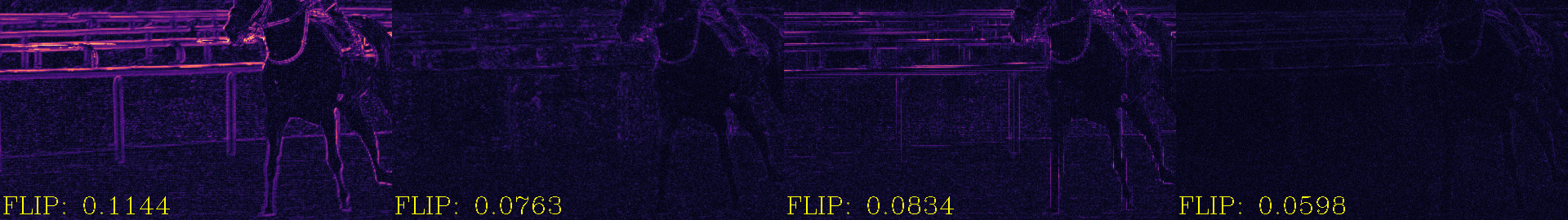}} \\
    \multicolumn{3}{c}{\includegraphics[trim=480 0 0 0,clip,width=\mywidth\linewidth]{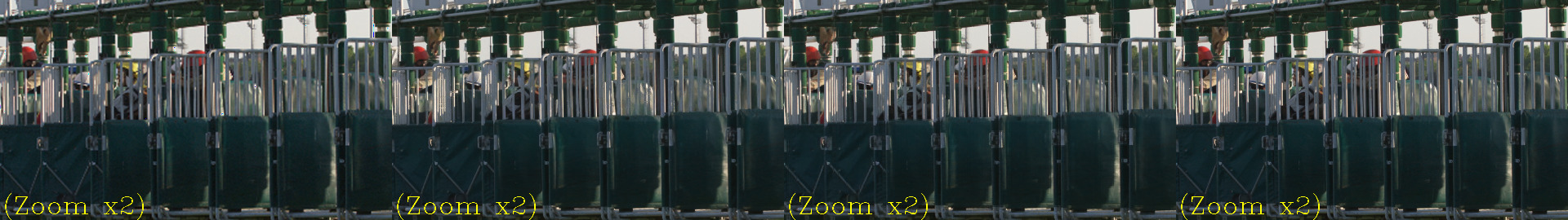}} \\
    \multicolumn{3}{c}{\includegraphics[trim=480 0 0 0,clip,width=\mywidth\linewidth]{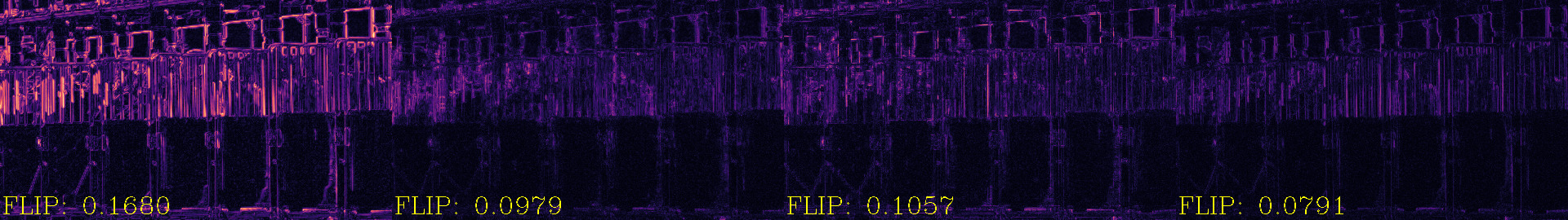}} \\ 
    \multicolumn{3}{c}{\includegraphics[trim=480 0 0 0,clip,width=\mywidth\linewidth]{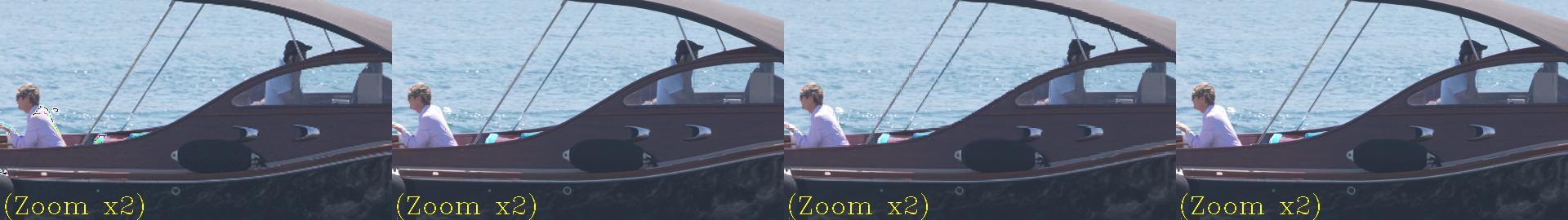}} \\
    \multicolumn{3}{c}{\includegraphics[trim=480 0 0 0,clip,width=\mywidth\linewidth]{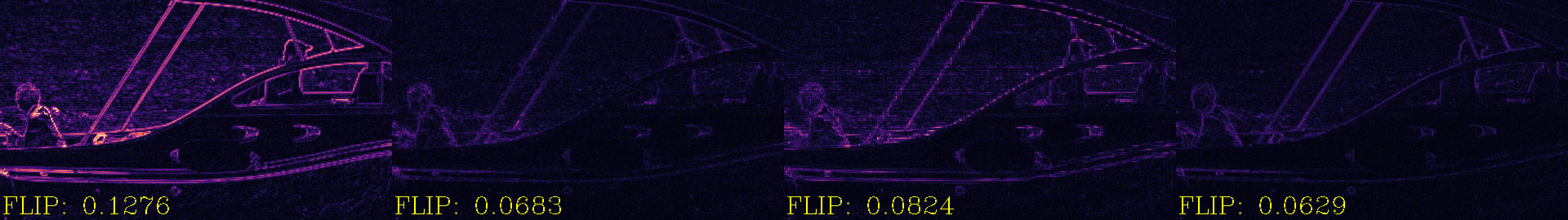}}
\end{tabular}%
\\
\begin{minipage}{0.2375\linewidth}
    \centering
    3D ModSIREN
\end{minipage}%
\begin{minipage}{0.2375\linewidth}
    \centering
    NVP
\end{minipage}%
\begin{minipage}{0.2375\linewidth}
    \centering
    NVTM
\end{minipage}
\caption{\small Visualization of video super resolution results for the first frame on the UVG sequences (Bosphorus, Jockey, ReadySetGo and YachtRide from top to bottom). For each sequence, the first row indicates the \textit{doubled spatial} decoded results and we zoomed them for better clarity.
The second row stands out the FLIP results, while darker colors indicating better performance.} 
\label{fig:sr_uvg_hd}
\end{figure}

\subsection{Video frame interpolation}
\begin{figure}[H]
\renewcommand{\mywidth}{0.78}
\setlength{\tabcolsep}{1pt} % Default value: 6pt
\centering
\begin{tabular}{cccc}
    \hline
    \multicolumn{4}{c}{\includegraphics[trim=480 0 0 0,clip,width=\mywidth\linewidth]{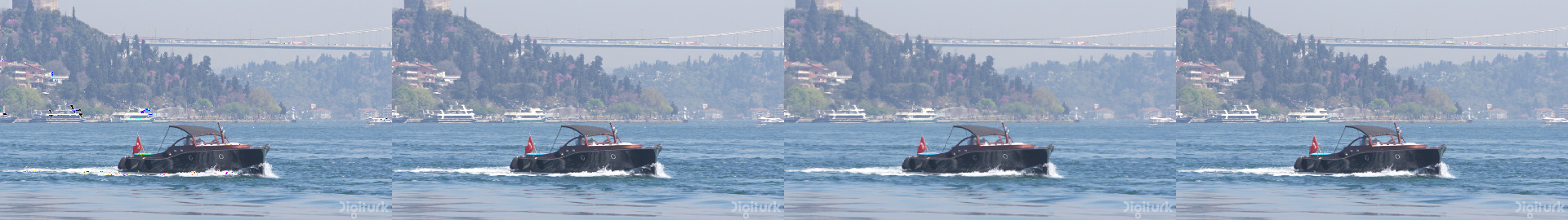}} \\
    \multicolumn{4}{c}{\includegraphics[trim=480 0 0 0,clip,width=\mywidth\linewidth]{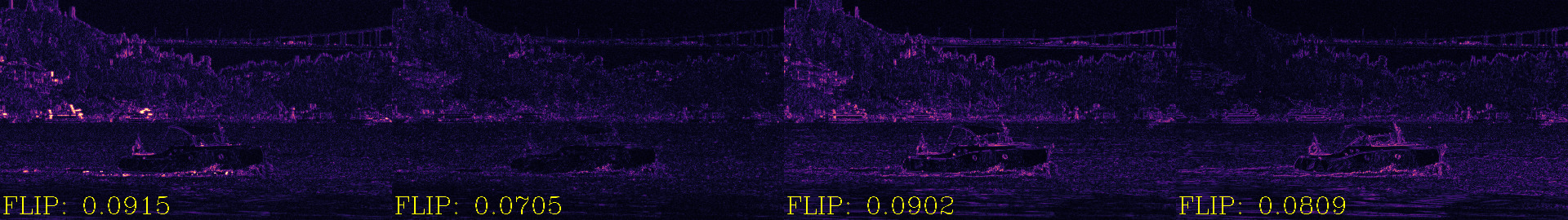}} \\ \hline
    \multicolumn{4}{c}{\includegraphics[trim=480 0 0 0,clip,width=\mywidth\linewidth]{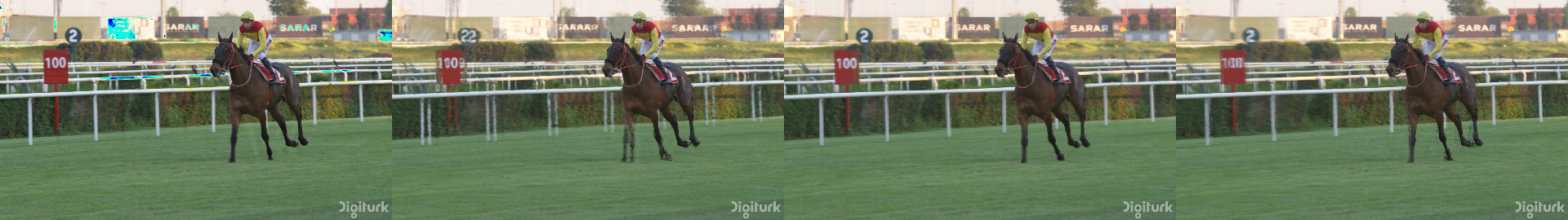}} \\
    \multicolumn{4}{c}{\includegraphics[trim=480 0 0 0,clip,width=\mywidth\linewidth]{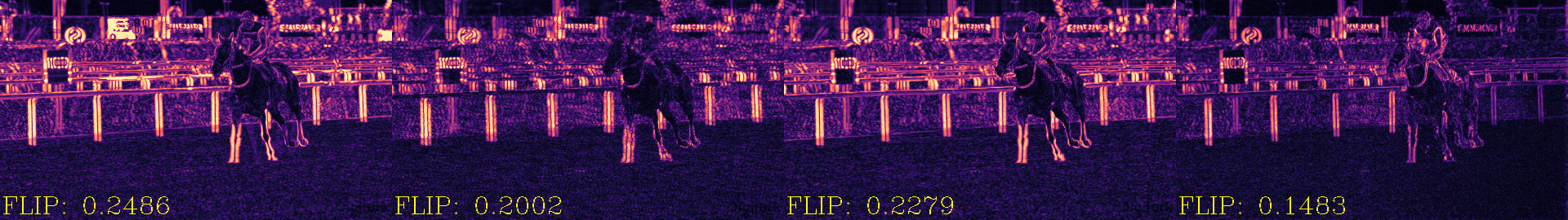}} \\ \hline
    \multicolumn{4}{c}{\includegraphics[trim=480 0 0 0,clip,width=\mywidth\linewidth]{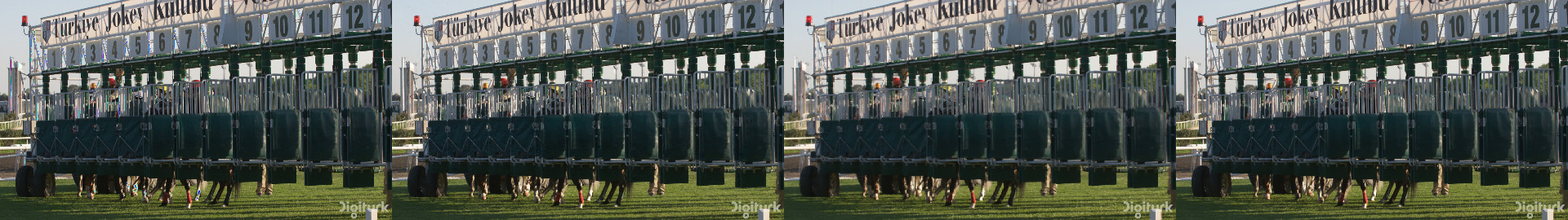}} \\
    \multicolumn{4}{c}{\includegraphics[trim=480 0 0 0,clip,width=\mywidth\linewidth]{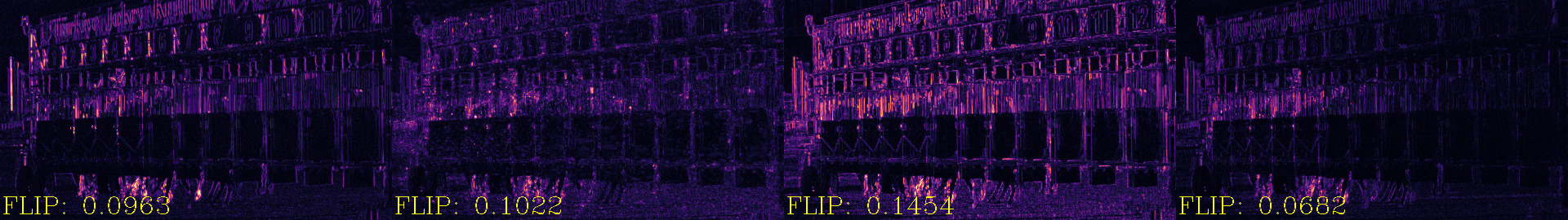}} \\ \hline
    \multicolumn{4}{c}{\includegraphics[trim=480 0 0 0,clip,width=\mywidth\linewidth]{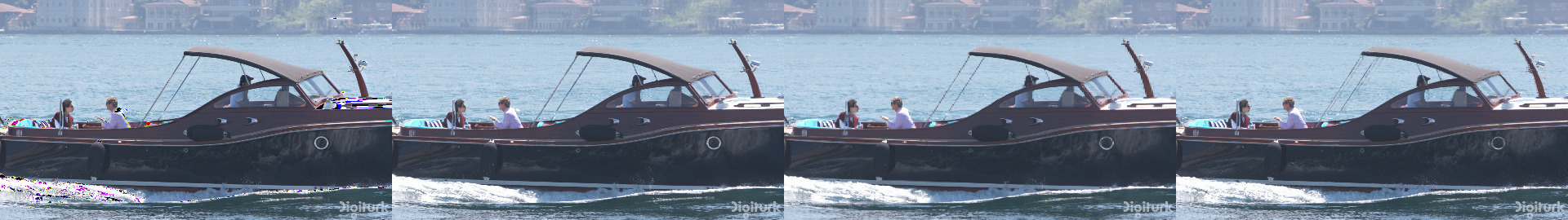}} \\
    \multicolumn{4}{c}{\includegraphics[trim=480 0 0 0,clip,width=\mywidth\linewidth]{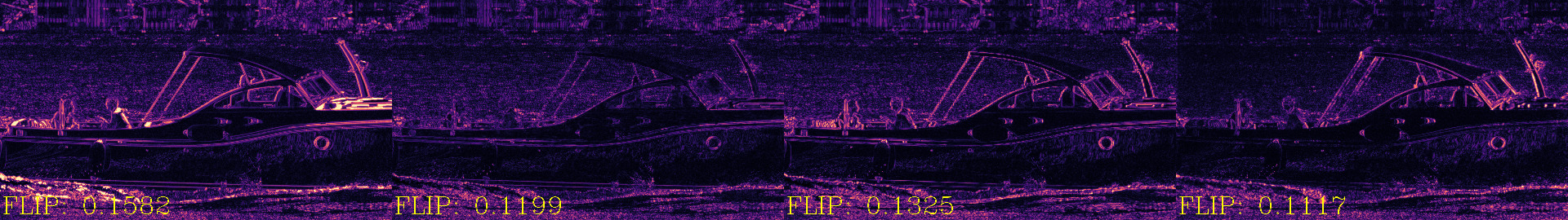}}
\end{tabular}%
\\ 
    \begin{minipage}{0.2375\linewidth}
        \centering
        3D ModSIREN
    \end{minipage}%
    \begin{minipage}{0.2375\linewidth}
        \centering
        NVP
    \end{minipage}%
    \begin{minipage}{0.2375\linewidth}
        \centering
        NVTM
    \end{minipage}
\caption{\small Visualization of Video frame interpolation results for the second frame (i.e. first interpolated frame) on the UVG sequences (Bosphorus, Jockey, ReadySetGo and YachtRide from top to bottom). For each sequence, the first row indicates the \textit{doubled temporal} decoded results, and the second row stands out the FLIP results, while darker colors indicating better performance.} 
\label{fig:fi_uvg_hd}
\end{figure}

\end{document}